\documentclass[10pt,twocolumn,letterpaper]{article}

\usepackage[pagenumbers]{cvpr} %

\usepackage{siunitx}
\usepackage{multirow}
\usepackage{makecell}
\usepackage{lipsum}
\usepackage{comment}
\usepackage{array}
\usepackage{subcaption}
\usepackage{caption}
\usepackage[T1]{fontenc}
\usepackage[utf8]{inputenc} 
\usepackage[normalem]{ulem} %
\usepackage{relsize}
\usepackage{xcolor}
\usepackage{soul}
\usepackage{cuted}

\usepackage[accsupp]{axessibility}  %

\newcommand{\ci}[1]{} %
\newcommand{\maskfn}{\text{Mask}}
\newcommand{\enct}{$\overline{E}$ }
\newcommand{\enc}{\overline{E}}

\newcommand{\Real}{\mathbb{R}}
\newcommand{\rando}{R\&O}
\newcommand{\prmt}{{PriVi}}
\newcommand{\dalp}{{DaLP}}

\newcommand{\blind}[1]{#1} %

\newcommand{\crossdomain}[1]{{\color{gray} #1}}
\newcommand{\csquare}[1]{\raisebox{0.25 em}{{\fcolorbox{#1}{#1}{\null}}}}

\newcommand{\nicetable}{
    \centering
    \small
    \setlength{\tabcolsep}{4pt}
\renewcommand{\arraystretch}{1.2}
    }

\usepackage{microtype}

\definecolor{vjepa_blue}{RGB}{204, 204, 255}

\definecolor{mplaliceblue}{rgb}{0.941176470588,0.972549019608,1.0}
\definecolor{mplantiquewhite}{rgb}{0.980392156863,0.921568627451,0.843137254902}
\definecolor{mplaqua}{rgb}{0.0,1.0,1.0}
\definecolor{mplaquamarine}{rgb}{0.498039215686,1.0,0.83137254902}
\definecolor{mplazure}{rgb}{0.941176470588,1.0,1.0}
\definecolor{mplbeige}{rgb}{0.960784313725,0.960784313725,0.862745098039}
\definecolor{mplbisque}{rgb}{1.0,0.894117647059,0.76862745098}
\definecolor{mplblack}{rgb}{0.0,0.0,0.0}
\definecolor{mplblanchedalmond}{rgb}{1.0,0.921568627451,0.803921568627}
\definecolor{mplblue}{rgb}{0.0,0.0,1.0}
\definecolor{mplblueviolet}{rgb}{0.541176470588,0.16862745098,0.886274509804}
\definecolor{mplbrown}{rgb}{0.647058823529,0.164705882353,0.164705882353}
\definecolor{mplburlywood}{rgb}{0.870588235294,0.721568627451,0.529411764706}
\definecolor{mplcadetblue}{rgb}{0.372549019608,0.619607843137,0.627450980392}
\definecolor{mplchartreuse}{rgb}{0.498039215686,1.0,0.0}
\definecolor{mplchocolate}{rgb}{0.823529411765,0.411764705882,0.117647058824}
\definecolor{mplcoral}{rgb}{1.0,0.498039215686,0.313725490196}
\definecolor{mplcornflowerblue}{rgb}{0.392156862745,0.58431372549,0.929411764706}
\definecolor{mplcornsilk}{rgb}{1.0,0.972549019608,0.862745098039}
\definecolor{mplcrimson}{rgb}{0.862745098039,0.078431372549,0.235294117647}
\definecolor{mplcyan}{rgb}{0.0,1.0,1.0}
\definecolor{mpldarkblue}{rgb}{0.0,0.0,0.545098039216}
\definecolor{mpldarkcyan}{rgb}{0.0,0.545098039216,0.545098039216}
\definecolor{mpldarkgoldenrod}{rgb}{0.721568627451,0.525490196078,0.043137254902}
\definecolor{mpldarkgray}{rgb}{0.662745098039,0.662745098039,0.662745098039}
\definecolor{mpldarkgreen}{rgb}{0.0,0.392156862745,0.0}
\definecolor{mpldarkgrey}{rgb}{0.662745098039,0.662745098039,0.662745098039}
\definecolor{mpldarkkhaki}{rgb}{0.741176470588,0.717647058824,0.419607843137}
\definecolor{mpldarkmagenta}{rgb}{0.545098039216,0.0,0.545098039216}
\definecolor{mpldarkolivegreen}{rgb}{0.333333333333,0.419607843137,0.18431372549}
\definecolor{mpldarkorange}{rgb}{1.0,0.549019607843,0.0}
\definecolor{mpldarkorchid}{rgb}{0.6,0.196078431373,0.8}
\definecolor{mpldarkred}{rgb}{0.545098039216,0.0,0.0}
\definecolor{mpldarksalmon}{rgb}{0.913725490196,0.588235294118,0.478431372549}
\definecolor{mpldarkseagreen}{rgb}{0.560784313725,0.737254901961,0.560784313725}
\definecolor{mpldarkslateblue}{rgb}{0.282352941176,0.239215686275,0.545098039216}
\definecolor{mpldarkslategray}{rgb}{0.18431372549,0.309803921569,0.309803921569}
\definecolor{mpldarkslategrey}{rgb}{0.18431372549,0.309803921569,0.309803921569}
\definecolor{mpldarkturquoise}{rgb}{0.0,0.807843137255,0.819607843137}
\definecolor{mpldarkviolet}{rgb}{0.580392156863,0.0,0.827450980392}
\definecolor{mpldeeppink}{rgb}{1.0,0.078431372549,0.576470588235}
\definecolor{mpldeepskyblue}{rgb}{0.0,0.749019607843,1.0}
\definecolor{mpldimgray}{rgb}{0.411764705882,0.411764705882,0.411764705882}
\definecolor{mpldimgrey}{rgb}{0.411764705882,0.411764705882,0.411764705882}
\definecolor{mpldodgerblue}{rgb}{0.117647058824,0.564705882353,1.0}
\definecolor{mplfirebrick}{rgb}{0.698039215686,0.133333333333,0.133333333333}
\definecolor{mplfloralwhite}{rgb}{1.0,0.980392156863,0.941176470588}
\definecolor{mplforestgreen}{rgb}{0.133333333333,0.545098039216,0.133333333333}
\definecolor{mplfuchsia}{rgb}{1.0,0.0,1.0}
\definecolor{mplgainsboro}{rgb}{0.862745098039,0.862745098039,0.862745098039}
\definecolor{mplghostwhite}{rgb}{0.972549019608,0.972549019608,1.0}
\definecolor{mplgold}{rgb}{1.0,0.843137254902,0.0}
\definecolor{mplgoldenrod}{rgb}{0.854901960784,0.647058823529,0.125490196078}
\definecolor{mplgray}{rgb}{0.501960784314,0.501960784314,0.501960784314}
\definecolor{mplgreen}{rgb}{0.0,0.501960784314,0.0}
\definecolor{mplgreenyellow}{rgb}{0.678431372549,1.0,0.18431372549}
\definecolor{mplgrey}{rgb}{0.501960784314,0.501960784314,0.501960784314}
\definecolor{mplhoneydew}{rgb}{0.941176470588,1.0,0.941176470588}
\definecolor{mplhotpink}{rgb}{1.0,0.411764705882,0.705882352941}
\definecolor{mplindianred}{rgb}{0.803921568627,0.360784313725,0.360784313725}
\definecolor{mplindigo}{rgb}{0.294117647059,0.0,0.509803921569}
\definecolor{mplivory}{rgb}{1.0,1.0,0.941176470588}
\definecolor{mplkhaki}{rgb}{0.941176470588,0.901960784314,0.549019607843}
\definecolor{mpllavender}{rgb}{0.901960784314,0.901960784314,0.980392156863}
\definecolor{mpllavenderblush}{rgb}{1.0,0.941176470588,0.960784313725}
\definecolor{mpllawngreen}{rgb}{0.486274509804,0.988235294118,0.0}
\definecolor{mpllemonchiffon}{rgb}{1.0,0.980392156863,0.803921568627}
\definecolor{mpllightblue}{rgb}{0.678431372549,0.847058823529,0.901960784314}
\definecolor{mpllightcoral}{rgb}{0.941176470588,0.501960784314,0.501960784314}
\definecolor{mpllightcyan}{rgb}{0.878431372549,1.0,1.0}
\definecolor{mpllightgoldenrodyellow}{rgb}{0.980392156863,0.980392156863,0.823529411765}
\definecolor{mpllightgreen}{rgb}{0.564705882353,0.933333333333,0.564705882353}
\definecolor{mpllightgrey}{rgb}{0.827450980392,0.827450980392,0.827450980392}
\definecolor{mpllightpink}{rgb}{1.0,0.713725490196,0.756862745098}
\definecolor{mpllightsalmon}{rgb}{1.0,0.627450980392,0.478431372549}
\definecolor{mpllightseagreen}{rgb}{0.125490196078,0.698039215686,0.666666666667}
\definecolor{mpllightskyblue}{rgb}{0.529411764706,0.807843137255,0.980392156863}
\definecolor{mpllightslategray}{rgb}{0.466666666667,0.533333333333,0.6}
\definecolor{mpllightslategrey}{rgb}{0.466666666667,0.533333333333,0.6}
\definecolor{mpllightsteelblue}{rgb}{0.690196078431,0.76862745098,0.870588235294}
\definecolor{mpllightyellow}{rgb}{1.0,1.0,0.878431372549}
\definecolor{mpllime}{rgb}{0.0,1.0,0.0}
\definecolor{mpllimegreen}{rgb}{0.196078431373,0.803921568627,0.196078431373}
\definecolor{mpllinen}{rgb}{0.980392156863,0.941176470588,0.901960784314}
\definecolor{mplmagenta}{rgb}{1.0,0.0,1.0}
\definecolor{mplmaroon}{rgb}{0.501960784314,0.0,0.0}
\definecolor{mplmediumaquamarine}{rgb}{0.4,0.803921568627,0.666666666667}
\definecolor{mplmediumblue}{rgb}{0.0,0.0,0.803921568627}
\definecolor{mplmediumorchid}{rgb}{0.729411764706,0.333333333333,0.827450980392}
\definecolor{mplmediumpurple}{rgb}{0.576470588235,0.439215686275,0.858823529412}
\definecolor{mplmediumseagreen}{rgb}{0.235294117647,0.701960784314,0.443137254902}
\definecolor{mplmediumslateblue}{rgb}{0.482352941176,0.407843137255,0.933333333333}
\definecolor{mplmediumspringgreen}{rgb}{0.0,0.980392156863,0.603921568627}
\definecolor{mplmediumturquoise}{rgb}{0.282352941176,0.819607843137,0.8}
\definecolor{mplmediumvioletred}{rgb}{0.780392156863,0.0823529411765,0.521568627451}
\definecolor{mplmidnightblue}{rgb}{0.0980392156863,0.0980392156863,0.439215686275}
\definecolor{mplmintcream}{rgb}{0.960784313725,1.0,0.980392156863}
\definecolor{mplmistyrose}{rgb}{1.0,0.894117647059,0.882352941176}
\definecolor{mplmoccasin}{rgb}{1.0,0.894117647059,0.709803921569}
\definecolor{mplnavajowhite}{rgb}{1.0,0.870588235294,0.678431372549}
\definecolor{mplnavy}{rgb}{0.0,0.0,0.501960784314}
\definecolor{mploldlace}{rgb}{0.992156862745,0.960784313725,0.901960784314}
\definecolor{mplolive}{rgb}{0.501960784314,0.501960784314,0.0}
\definecolor{mplolivedrab}{rgb}{0.419607843137,0.556862745098,0.137254901961}
\definecolor{mplorange}{rgb}{1.0,0.647058823529,0.0}
\definecolor{mplorangered}{rgb}{1.0,0.270588235294,0.0}
\definecolor{mplorchid}{rgb}{0.854901960784,0.439215686275,0.839215686275}
\definecolor{mplpalegoldenrod}{rgb}{0.933333333333,0.909803921569,0.666666666667}
\definecolor{mplpalegreen}{rgb}{0.596078431373,0.98431372549,0.596078431373}
\definecolor{mplpalevioletred}{rgb}{0.686274509804,0.933333333333,0.933333333333}
\definecolor{mplpapayawhip}{rgb}{1.0,0.937254901961,0.835294117647}
\definecolor{mplpeachpuff}{rgb}{1.0,0.854901960784,0.725490196078}
\definecolor{mplperu}{rgb}{0.803921568627,0.521568627451,0.247058823529}
\definecolor{mplpink}{rgb}{1.0,0.752941176471,0.796078431373}
\definecolor{mplplum}{rgb}{0.866666666667,0.627450980392,0.866666666667}
\definecolor{mplpowderblue}{rgb}{0.690196078431,0.878431372549,0.901960784314}
\definecolor{mplpurple}{rgb}{0.501960784314,0.0,0.501960784314}
\definecolor{mplred}{rgb}{1.0,0.0,0.0}
\definecolor{mplrosybrown}{rgb}{0.737254901961,0.560784313725,0.560784313725}
\definecolor{mplroyalblue}{rgb}{0.254901960784,0.411764705882,0.882352941176}
\definecolor{mplsaddlebrown}{rgb}{0.545098039216,0.270588235294,0.0745098039216}
\definecolor{mplsalmon}{rgb}{0.980392156863,0.501960784314,0.447058823529}
\definecolor{mplsandybrown}{rgb}{0.980392156863,0.643137254902,0.376470588235}
\definecolor{mplseagreen}{rgb}{0.180392156863,0.545098039216,0.341176470588}
\definecolor{mplseashell}{rgb}{1.0,0.960784313725,0.933333333333}
\definecolor{mplsienna}{rgb}{0.627450980392,0.321568627451,0.176470588235}
\definecolor{mplsilver}{rgb}{0.752941176471,0.752941176471,0.752941176471}
\definecolor{mplskyblue}{rgb}{0.529411764706,0.807843137255,0.921568627451}
\definecolor{mplslateblue}{rgb}{0.41568627451,0.352941176471,0.803921568627}
\definecolor{mplslategray}{rgb}{0.439215686275,0.501960784314,0.564705882353}
\definecolor{mplslategrey}{rgb}{0.439215686275,0.501960784314,0.564705882353}
\definecolor{mplsnow}{rgb}{1.0,0.980392156863,0.980392156863}
\definecolor{mplspringgreen}{rgb}{0.0,1.0,0.498039215686}
\definecolor{mplsteelblue}{rgb}{0.274509803922,0.509803921569,0.705882352941}
\definecolor{mpltan}{rgb}{0.823529411765,0.705882352941,0.549019607843}
\definecolor{mplteal}{rgb}{0.0,0.501960784314,0.501960784314}
\definecolor{mplthistle}{rgb}{0.847058823529,0.749019607843,0.847058823529}
\definecolor{mpltomato}{rgb}{1.0,0.388235294118,0.278431372549}
\definecolor{mplturquoise}{rgb}{0.250980392157,0.878431372549,0.81568627451}
\definecolor{mplviolet}{rgb}{0.933333333333,0.509803921569,0.933333333333}
\definecolor{mplwheat}{rgb}{0.960784313725,0.870588235294,0.701960784314}
\definecolor{mplwhite}{rgb}{1.0,1.0,1.0}
\definecolor{mplwhitesmoke}{rgb}{0.960784313725,0.960784313725,0.960784313725}
\definecolor{mplyellow}{rgb}{1.0,1.0,0.0}
\definecolor{mplyellowgreen}{rgb}{0.603921568627,0.803921568627,0.196078431373}

\definecolor{cvprblue}{rgb}{0.21,0.49,0.74}
\usepackage[pagebackref,breaklinks,colorlinks,allcolors=cvprblue]{hyperref}
\usepackage{printlen} %
\renewcommand{\paragraph}[1]{%
  \vspace{4pt}%
  \par\noindent\textbf{#1.}\ %
}
\newcommand{\para}[1]{%
  \vspace{4pt}%
  \par\noindent\textbf{#1}%
}

\def\paperID{****} %
\def\confName{CVPR}
\def\confYear{2026}

\title{PriVi: Towards a General-Purpose Video Model for Primate Behavior in the Wild}

\begin{document}
\maketitle

\setlength{\stripsep}{-44pt}
\begin{strip}
{\bf
Felix B. Mueller\textsuperscript{1},
Jan F. Meier\textsuperscript{1},
Timo Lueddecke\textsuperscript{1},
Richard Vogg\textsuperscript{1,3},
Roger L. Freixanet\textsuperscript{1},
Valentin Hassler\textsuperscript{1},
Tiffany Bosshard\textsuperscript{2},
Elif Karakoc\textsuperscript{3},
William J. O'Hearn\textsuperscript{1,2,*},
Sofia M. Pereira\textsuperscript{1,4},
Sandro Sehner\textsuperscript{3,\dag},
Kaja Wierucka\textsuperscript{3},
Judith Burkart\textsuperscript{7},
Claudia Fichtel\textsuperscript{3},
Julia Fischer\textsuperscript{1,2},
Alexander Gail\textsuperscript{1,5},
Catherine Hobaiter\textsuperscript{8},
Julia Ostner\textsuperscript{1,4},
Liran Samuni\textsuperscript{6},
Oliver Sch\"ulke\textsuperscript{1,4},
Neda Shahidi\textsuperscript{1,5},
Erin G. Wessling\textsuperscript{6},
Alexander S. Ecker\textsuperscript{1,9}
}\\
{\footnotesize
\textsuperscript{\bf 1}\,University of G\"ottingen, Germany \;\;
\textbf{\{}\textsuperscript{\bf 2}\,Cognitive Ethology Lab \;\;
\textsuperscript{\bf 3}\,Behavioral Ecology \& Sociobiology \;\;
\textsuperscript{\bf 4}\,Social Evolution in Primates \;\;
\textsuperscript{\bf 5}\,Cognitive Neuroscience SMG \;\;
\textsuperscript{\bf 6}\,Cooperative Evolution\textbf{\}}, German Primate Center (DPZ) -- Leibniz Institute for Primate Research, G\"ottingen, Germany \;\;
\textsuperscript{\bf 7}\,University of Zurich, Switzerland \;\;
\textsuperscript{\bf 8}\,University of St. Andrews, United Kingdom \;\;
\textsuperscript{\bf 9}\,Max Planck Institute for Dynamics and Self-Organization, G\"ottingen, Germany.\\
\textsuperscript{\bf *}\,Current address: University of Exeter, United Kingdom \;\;
\textsuperscript{\bf \dag}\,Current address: Domestication Lab, Konrad Lorenz Institute of Ethology, University of Veterinary Medicine Vienna, Austria. 
 \textbf{Contact:}
\href{mailto:felix.mueller@cs.uni-goettingen.de}{felix.mueller@cs.uni-goettingen.de}, \href{mailto:ecker@cs.uni-goettingen.de}{ecker@cs.uni-goettingen.de}
}
\vspace{44pt}
\vspace{20pt}
\end{strip}

\begin{abstract}

Non-human primates are our closest living relatives, and analyzing their behavior is central to research in cognition, evolution, and conservation. Computer vision could greatly aid this research, but existing methods often rely on human-centric pretrained models and focus on single datasets, which limits generalization. We address this limitation by shifting from a model-centric to a data-centric approach and introduce \prmt{}, a large-scale primate-centric video pretraining dataset. \prmt{} contains 424 hours of curated video, combining 174 hours from behavioral research across 11 settings with 250 hours of diverse web-sourced footage, assembled through a scalable data curation pipeline. We continue pretraining V-JEPA, a large-scale video model,  on \prmt{} to learn primate-specific representations and evaluate it using a lightweight frozen classifier. Across four benchmark datasets --~ChimpACT, PanAf500, BaboonLand, and ChimpBehave~-- our approach consistently outperforms prior work, including fully finetuned baselines, and scales favorably with fewer labels. These results demonstrate for the first time that domain-level pretraining, where pretraining is conducted on similar data but not the target dataset itself, works for video models. Our primate-centric pretraining substantially improves data efficiency and generalization, making it a promising approach for low-label applications. Dataset, code, and models are available at \url{https://privi.eckerlab.org}.

\end{abstract}

\begin{figure}
    \centering
    \includegraphics[width=0.95\linewidth]{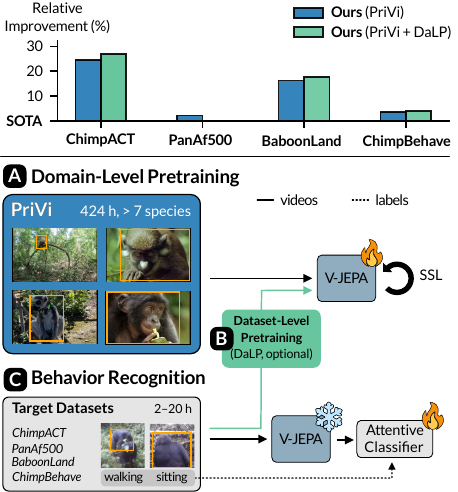}
    \caption{\textbf{Domain-level pretraining \textsf{(A)} on our diverse, large-scale dataset \prmt{} surpasses state-of-the-art (SOTA) models across four behavior recognition datasets.} Additional dataset-level pretraining \textbf{\textsf{(B)}} using self-supervised learning (SSL) further improves performance on most datasets. We recognize behaviors using frozen evaluation \textbf{\textsf{(C)}}.  Images: ours, \cite{brookes_panaf20k_2024}.}
    \label{fig:teaser}
\end{figure}

\section{Introduction}
\label{sec:intro}

\begin{figure*}
    \centering
    \includegraphics[width=1\linewidth]{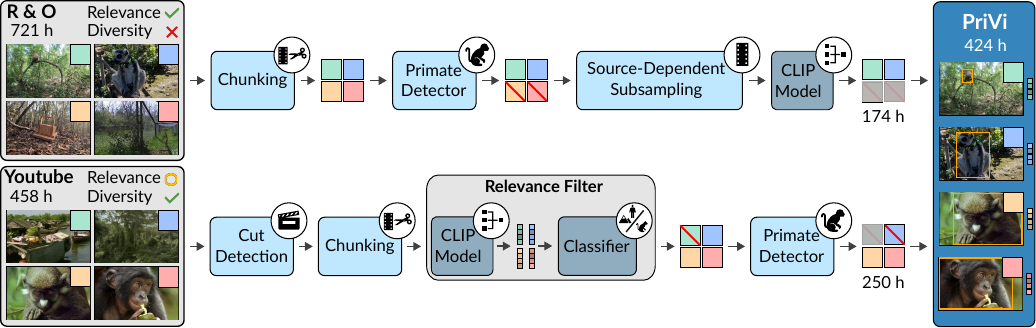}
    \caption{\textbf{Our preprocessing pipeline for \prmt{}.} YouTube data is filtered using a learned relevance classifier, while \rando{} is subsampled based on source dataset metadata. We apply zero-shot primate detection on both to generate bounding boxes and discard empty frames. \prmt{} consists of 424\,h of unique video complete with bounding boxes and CLIP embeddings for keyframes.}
    \label{fig:data_pipeline}
\end{figure*}

Understanding the behavior of wild animals is crucial for  fields like cognition, ecology, and animal conservation, with the behavior of non-human primates being of especially high interest, due to its complexity and relation to human cognition. Advances in video recording technology could revolutionize the study of primate behavior by allowing to capture and analyze vast amounts of data. Current research protocols rely mostly on manual annotation by experts. This approach limits the amount of video material that can be used to draw scientific conclusions and is subject to human bias. Computer vision methods have the potential to provide complementary tools and objective means to assess video material, opening new avenues for behavioral research. 

Consequently, there has been a growing interest and impressive progress in computer vision for both primate behavior \cite{bain_automated_2021, brookes_panaf20k_2024, brookes_panaf-fgbg_2025, ma_chimpact_2023, fuchs_forest_2025, duporge_baboonland_2025} and  animal behavior in general  \cite{rodriguez-juan_visual_2025, ng_animal_2022, gabeff_mammalps_2025}. Most studies introduce specialized models optimized for specific datasets which require considerable amounts of training data \cite{bain_automated_2021, ma_alphachimp_2024, brookes_panaf20k_2024}. However, for widespread adoption, the community needs models that can be shared between tasks and datasets and can be adapted to specific tasks with few labeled samples. 

Recent foundation models generalize impressively across many computer vision tasks, including -- to some extent -- animal behavior \cite{sun_video_2024, zhao_videoprism_2024}.
Yet, pretraining data for those models is still very human-centric \cite{bardes_revisiting_2024} and thus out-of-domain.  We know that continuing pretraining in-domain on the target dataset, e.g. a specific action recognition dataset, (\textit{dataset-level pretraining}) improves video model performance~\cite{tong_videomae_2022, mueller_domain-adaptive_2025}. However, this requires retraining the model for each target dataset. Our goal is to do \textit{domain-level pretraining}, i.e. pretraining on diverse videos from one domain to improve performance across all target datasets in this domain, without having to include the target dataset in the pretraining data. This has been shown to work for language models~\cite{DBLP:conf/acl/GururanganMSLBD20}, but results for video models are still lacking.

Domain-level pretraining requires large-scale diverse datasets and while there have been recent efforts to curate such datasets of animals \cite{zhao_web-scale_2025, yang_apt-36k_2022, ng_animal_2022}, these datasets contain few to no primates. Also, state-of-the-art data curation pipelines typically require high-quality seed datasets or text captions~\cite{oquab_dinov2_2024, assran_v-jepa_2025, zhao_videoprism_2024, gadre_datacomp_2023, bardes_revisiting_2024}, which is infeasible for many application domains.

In this paper, we work towards filling these gaps. Our contributions are: 
\begin{itemize}
    \item A data curation pipeline, which uniquely requires neither a seed dataset nor text captions.
    \item \prmt{}, a 424-hour large-scale diverse \textbf{Pri}mate \textbf{Vi}deo dataset, assembled using the data curation pipeline.
    \item A frozen evaluation setup for behavior recognition that scales well to small datasets and is further improved by optional dataset-level pretraining.
    \item Showing for the first time that domain-level pretraining improves performance for video models.
    \item Our approach outperforms prior work across four primate behavior datasets. (\cref{fig:teaser}).
\end{itemize}

\section{Related Work}
\label{sec:related_work}

\paragraph{Self-supervised video learning} Unsupervised representation learning has been a cornerstone of machine learning research \cite{tenenbaum_global_2000, hinton_fast_2006, hinton_reducing_2006, vincent_extracting_2008, kingma_auto-encoding_2014, rezende_stochastic_2014}. The recent success of large language models using masked language modeling popularized self-supervised learning (SSL) as we know it today \cite{devlin_bert_2019, radford_improving_2018, brown_language_2020}. A plethora of SSL techniques exist, ranging from contrastive and deep metric learning \cite{radford_learning_2021, chen_simple_2020} over self-distillation \cite{caron_emerging_2021} to masked reconstruction in pixel~\cite{tong_videomae_2022, he_masked_2021} or latent space~\cite{bardes_revisiting_2024, salehi_sigma_2024, assran_self-supervised_2023, assran_v-jepa_2025}, with many modern approaches combining several of those paradigms \cite{oquab_dinov2_2024, zhou_image_2022}.

The key ingredient for today's foundation models is combining SSL with massive Web-scale datasets~\cite{schuhmann_laion-5b_2022, miech_howto100m_2019, kuznetsova_open_2020}. These image foundation models, most notably CLIP~\cite{radford_learning_2021} and DINOv2~\cite{oquab_dinov2_2024},  caused a paradigm shift from handcrafted and task-specific architectures towards models using foundation models as frozen feature extractors for many computer vision tasks~\cite{luddecke_image_2022, ryan_gaze-lle_2025}. Recently, there has also been considerable advancement in foundation models for video understanding with several architectures proposed based on video-caption pairs like PerceptionEncoder~\cite{bolya_perception_2025} and VideoCLIP~\cite{xu_videoclip_2021}, masked video autoencoding \cite{tong_videomae_2022, feichtenhofer_masked_2022}, masked reconstruction in latent space like V-JEPA~\cite{bardes_revisiting_2024,assran_v-jepa_2025}, and combined approaches like VideoPrism~\cite{zhao_videoprism_2024}.

Recent work on data-centric learning \cite{gadre_datacomp_2023} shows that neither model nor data scale is sufficient: Careful curation of training data considerably improves performance. Many foundation models thus employ data engines for automated data curation \cite{oquab_dinov2_2024, bolya_perception_2025, ravi_sam_2025, kirillov_segment_2023} .

\paragraph{Animal behavior recognition} 
Recent years have seen a rapidly growing number of annotated datasets for animal behavior recognition, either for single species~\cite{rodriguez-juan_visual_2025, gabeff_mammalps_2025} or across species~\cite{ng_animal_2022, chen_mammalnet_2023}. Primate specific datasets cover diverse settings, ranging from zoo recordings in \mbox{ChimpACT}~\cite{ma_chimpact_2023} and ChimpBehave~\cite{fuchs_forest_2025} over in-the-wild recordings using camera traps in the PanAf-family of datasets~\cite{brookes_panaf20k_2024, brookes_panaf-fgbg_2025} to drone footage in BaboonLand~\cite{duporge_baboonland_2025}. The size of these datasets ranges between 2\,h of frame-wise  annotations (\mbox{ChimpACT}, PanAf500) to 80\,h of clip-wise annotations (PanAf20k). Behavior recognition is usually operationalized as an action classification task, e.\,g. classifying which actions are performed in a specific miniclip \cite{duporge_baboonland_2025, fuchs_forest_2025, brookes_panaf20k_2024}, or spatio-temporal action recognition, where actions need to be localized in videos with \cite{ma_chimpact_2023} or without \cite{ma_alphachimp_2024} ground truth bounding boxes of animals given.

There has been considerable prior work for automated behavior recognition: \citet{bain_automated_2021} showed that they can discriminate two distinctive actions in wild chimpanzees using audiovisual input. Several methods for more challenging behavior recognition, like distinguishing between all classes in PanAf \cite{brookes_panaf20k_2024} or ChimpACT \cite{ma_chimpact_2023}, have been proposed  \cite{ma_alphachimp_2024, brookes_chimpvlm_2024, brookes_triple-stream_2023, sakib_visual_2020, brookes_panaf-fgbg_2025, fuchs_asbar_2023}. However, each of these works focuses on a single dataset and most build on models pretrained on human-centric datasets, like Kinetics~\cite{kay_kinetics_2017}.

\paragraph{Self-supervised learning for behavior recognition}
Foundation models are promising feature extractors for behavior recognition. Frozen evaluation of VideoPRISM models is state-of-the-art or competitive with specialized methods on ChimpACT~\cite{ma_chimpact_2023}, KABR~\cite{kholiavchenko_kabr_2024} -- a dataset of Kenyan wildlife --, as well as various datasets of lab rodents~\cite{sun_video_2024, zhao_videoprism_2024}. Jointly finetuning a large vision-language model on multiple animal behavior datasets improves performance across all \cite{mamooler_fine-tuning_2025}, suggesting a promising step towards unified, general-purpose models without the need for specialized architectures.

For animal behavior, unlabeled data is usually easy to acquire, while labeled data is scarce. Thus, several works explored utilizing in-domain unlabeled data and found it beneficial for animal identification~\cite{iashin_self-supervised_2025}, animal behavior analysis in the lab~\cite{wang_self-supervised_2025}, or weakly-supervised training for behavior retrieval~\cite{santo_fine-tuning_2025}. However, all of these methods again develop models specialized for individual datasets.

\begin{table}[]
    \nicetable
    \setlength{\tabcolsep}{6pt}
    \caption{\textbf{Overview of the composition of our dataset.} The species distribution of YT-Filtered was estimated by labeling a random subset of 900 frames, for \rando{} the species distribution is known.}
    \begin{tabular}{
    l
    S[table-format=2.1]
    S[table-format=2.1]
    S[table-format=2.1]}
    \toprule
    & \multicolumn{2}{c}{\textbf{Subsets}} & \multicolumn{1}{c}{\multirow{2}{*}{\textbf{\prmt}}}\\
    \cmidrule(lr){2-3}
         &  {YT-Filt.}& {\rando}&  \\
         \midrule
  Unique Hours & {250\,h} & {174\,h} & {424\,h}\\
 \midrule
         \multicolumn{1}{l}{\textbf{Genus/Family} [\%]} &&&\\
 Macaques& 63.1 & 14.1&43.0\\
 Chimpanzees& 7.8 & 35.7&19.3\\
 Orangutans & 4.1 & 0& 2.4\\
 Baboons & 1.3 & 16.2& 7.4\\
 True Lemurs& {$<1$} & 22.7&9.8\\
 Marmosets & {$<1$} & 5.7& 2.1\\
 Squirrel monkeys & {$<1$} & 5.4& 2.0\\
 Others & 8.1 & 0&4.8\\
 Not identifiable & 9.4 & 0& 5.5\\
 No primate & 6.0 & 0& 3.5\\
 \midrule
 \textbf{Setting} [\%]& & &\\
 In the wild & 59.6 & 62.7&60.9\\ 
 Wild-like & 27.8 & 22.4&25.6\\
 Indoors& 4.1 & 14.6&8.4\\
 Not identifiable & 8.6 & 0& 5.1\\
 \bottomrule 
 \end{tabular}

        \label{tab:dataset_composition}
\end{table}

\section{Methodology}
\label{sec:methodology}

Our approach utilizes diverse unlabeled primate videos for primate behavior recognition. Instead of training only on a small target dataset to learn representations for this specific task, we pretrain on a broad, large-scale primate dataset to improve representations useful for varying primate-related target datasets.

In this section, we first describe our pretraining dataset \prmt{} and our data curation pipeline. Then we describe our framework of using domain-level pretraining for primate behavior recognition, which consists of self-supervised pretraining and a frozen evaluation protocol.

\subsection{\prmt{}: Primate-Specific Pretraining Dataset}
\label{sec:dataset-description}

\paragraph{Research and Observational Data (\rando{})} To overcome the data-scarce regime, where it is best for each method to train only on their own data, we assemble a diverse pretraining dataset that spans various applications of primate behavior analysis in the wild. To do so, we pool large amounts of data from various past and ongoing behavioral ecology and animal behavior research projects 
\cite{pereira_complex_2025, shahidi_freely_2026, karakoc_foraging_2025, ohearn_increased_2025, ohearn_lessons_2024, bosshard_ecological_2025, sehner_sensitivity_2025}. In total, we pool 721\,h of raw video material from 11 source datasets. A source dataset is a collection of videos that are  homogeneous in their recording location, setting, and the species visible. 

Five of the source datasets contain primates in their natural habitat, two in semi-free-ranging settings, and four are of captive primates. Source datasets contain both purely observational studies as well as experimental interventions, e.\,g. feeding boxes, matching the diversity of real-world animal behavior recognition tasks. Recording locations in the natural habitat were in \blind{Kirindy, Madagascar; Simenti, Senegal; Phu Khieo, Thailand; and Moyen-Bafing, Guinea}. Semi-free ranging locations were \blind{Strau\ss berg, Germany and Rocamadour, France}, and captive settings were at the \blind{German Primate Center, G\"ottingen, Germany and the University of Zurich, Switzerland}. 

\paragraph{YouTube Data} The \rando{} dataset features realistic settings close to the setup of existing and future applications of primate behavior analysis. However, it still has low diversity compared to web-scale pretraining datasets. To overcome this limitation, we scrape a large dataset of YouTube videos to increase diversity. We search YouTube for playlists of videos of primates in the wild and manually select 19 playlists with long, high-quality videos. Downloading these playlists yielded a corpus of 458\,h of raw video material. We find that YouTube videos are diverse but on average low quality, with many frames containing e.\,g. interviews with human researchers, primates in cities or otherwise undesired content. We thus apply a data filtering and curation pipeline (see \cref{sec:data-curation-pipeline} below). 

\paragraph{Dataset versions: \prmt{}, YT-Filtered, \rando, and YT-Random} After filtering, we obtain 250 unique hours of curated YouTube videos (which we refer to as {YT-Filtered}) and 174 unique hours of curated research and observational data (which we refer to as {\rando}). Both together comprise our pretraining dataset {\prmt}. It consists of 720,000 three-second video snippets, which are partially overlapping, depending on the source dataset. 63\,\% of the snippets are from YT-Filtered and 37\,\% from \rando{}. For each video snippet, the dataset contains bounding boxes and clip embeddings for the center frame. For \rando{}, we provide species labels inferred from the source-dataset-specific species information and a zero-shot object detector (see \cref{sec:data-curation-pipeline}).

We release \prmt{} except for one source dataset comprising 14\,\% of our pretraining data. This source dataset contains camera trap footage from ongoing research projects and cannot yet be shared publicly. 
Consult \cref{tab:dataset_composition} for a breakdown of species and settings in \prmt{} and \cref{sec:supp_dataset} for more details and example images. For ablation purposes, we create {YT-Random}, a YouTube dataset of the same size as YT-Filtered, but with a random selection of videos instead of the curated set.

\subsection{Data Curation Pipeline}
\label{sec:data-curation-pipeline}

Previous work~\cite{gadre_datacomp_2023, oquab_dinov2_2024} has repeatedly shown that even for web-scale datasets, good data curation is crucial for model performance.  On the scale of hundreds of hours of video material, manual curation is infeasible. We thus automatically filter the dataset using a relevance scoring model on latent representations and a zero-shot object detector. We also screen for data contamination, see \cref{fig:data_pipeline} for an overview and \cref{sec:supp_data_pipeline} for more details.

\paragraph{Relevance Filtering}
Unlike naturalistic videos, YouTube videos feature frequent cuts. Having cuts in our training videos is undesirable, as the scene before and after the cut might be completely different, making self-supervised modeling harder and less aligned with cut-free target datasets. Following~\cite{zhao_web-scale_2025}, we detect cuts in the video by using an off-the-shelf cut detector~\cite{castellano_pyscenedetect_nodate}.
Even after cut detection, a longer scene might contain both relevant and irrelevant parts, e.\,g. a primate leaving the frame after half the clip. To select only relevant video snippets, we chunk videos in 3-s snippets with a stride of 2\,s, allowing for an 0.5\,s overlap with the previous and next snippet each, and determine for each snippet whether it is relevant for training. We specify a list of inclusion criteria to filter only real-world videos prominently featuring primates. To allow for scalable relevance filtering, one annotator manually labeled 2,500 images based on these criteria. Automated decision of these inclusion criteria requires image-level summary information, so we train a classifier on top of CLIP embeddings of center frames for relevance filtering. Our 2-layer MLP classifier achieves 82.8\,\% recall and 90.3\,\% precision on a held-out validation set with the threshold being optimized for higher precision than recall (ROC-AUC 95.9\,\%).

\paragraph{Detection Filtering}
Another considerable problem is that many collected videos contain either mostly background or are completely empty, especially for the naturalistic recordings in \rando{}. This is inefficient as self-supervised training would spend most of the training compute on learning representations of background information instead of capturing fine-grained details of primate behavior. We use a GroundingDINO~\cite{liu_grounding_2024} as zero-shot object detector and prompt it for primate bounding boxes on the chunk center frames on both \rando{} and YT-Filtered. Afterwards, we perform thresholding, non-maximum suppression, and discard chunks without any detections. We find zero-shot primate detection to work well enough for data preprocessing, with 82.7\,mAP\textsubscript{All} on the PanAf500 primate detection task.

\paragraph{Subsampling}
The YT dataset is already of suitable size after the relevance classifier. In \rando{}, however, source datasets differ vastly in size and diversity. We thus subsample based on source dataset metadata. In the \rando{} dataset, we aim for a proportion of 30\,\% for the diverse camera trap dataset, 10\,\% for other less diverse wild and semi-free ranging datasets, and 3 to 6\,\% for datasets of captive animals with low and high scene diversity, respectively. We initially extract 3-s snippets with a chunking stride between 1\,s and 3\,s for \rando{} to ensure that we capture enough snippets per dataset. After detection filtering, we subsample based on target proportion and remaining number of chunks to achieve the desired dataset composition. Note that the final proportions deviate from the target proportions, as some source datasets did not contain enough suitable samples.

\subsection{Self-Supervised Pretraining}
\label{subsec:ssl}

\begin{figure}
    \centering
    \includegraphics[width=1\columnwidth]{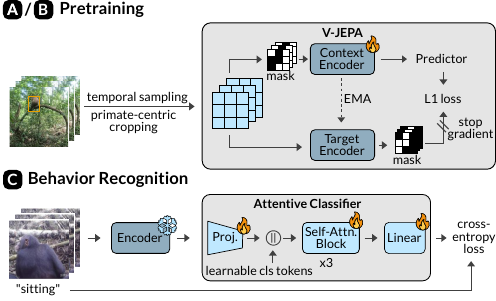} %
    \caption{\textbf{Our pretraining and evaluation architecture.} EMA: exponential moving average. $\|$: sequence concatenation. \textbf{A/B.} We continue self-supervised pretraining of a V-JEPA model on our primate dataset \prmt{} and optionally on the target dataset, doing masked prediction in latent space. \textbf{C.} We train a classifier for each target dataset. Images are ours or \cite{brookes_panaf20k_2024}.
    }
    \label{fig:model}
\end{figure}

We utilize V-JEPA~\cite{bardes_revisiting_2024}, a self-supervised learning approach that does not require video--text pairs but does masked prediction in latent space. We use its default settings.

The V-JEPA architecture consists of a context encoder $E$, a target encoder $\overline{E}$, and a predictor $P$ (see \cref{fig:model}). The input is a video tensor $x \in \Real^{16\times 224 \times 224}$ consisting of 16 frames and with a resolution of $224 \times 224$ pixels. The input tensor is obtained by performing random cropping around the bounding boxes predicted by the zero-shot primate detector (\cref{sec:data-curation-pipeline}). The input is tokenized by using non-overlapping $2 \times 16 \times 16$ patches and combined with sine-cosine positional embeddings to produce $N=1568$ tokens $X \in \Real^{N \times D}$, with token embedding dimension $D=1024$.

During training, part of the input sequence is masked and the training objective is to predict the latent representation of masked tokens given the unmasked tokens (see \cref{fig:model}). Let $\maskfn$ and $\maskfn^C$ be functions that split a set of tokens into the set of masked and unmasked tokens, respectively. The loss for an unlabeled data sample $X$ is 
\begin{equation}
\label{eq:LMAE}
    L_{\text{JEPA}} =\left\Vert P\left(E\left(\maskfn(X)\right)  \right) - \maskfn^C\left(\overline{E}(X)\right)\right\Vert_1.
\end{equation}
The parameters of context encoder $E$ and predictor $P$ are trained. To avoid collapse to trivial solutions, the parameters of target encoder $\overline{E}$ are not trained but computed as an exponential moving average over the past weights of the context encoder $E$~\cite{bardes_revisiting_2024}.

\subsection{Supervised Attentive Classifier}%
\label{subsec:behavior-recognition}

Frozen evaluation of large vision models is not only a means to demonstrate feature quality, it is also competitive for action recognition, particularly on small datasets \cite{sun_video_2024, zhao_videoprism_2024}. We aim to design an attentive classifier for primate behavior recognition on a frozen pretrained model. Existing designs are surprisingly large: V-JEPA~\cite{bardes_revisiting_2024} uses one self-attention block (12\,M parameters) and V-JEPA2~\cite{assran_v-jepa_2025} three self-attention plus one cross-attention block (49\,M). We hypothesize that this is overparameterized for typical small-scale animal behavior datasets, making models prone to overfitting and unstable training.

To reduce parameter count while keeping the classifier sufficiently deep to prevent underfitting, we downproject from $D=1024$ to $D'=64$ dimensions before the first self-attention block in the classifier (\cref{fig:model}). The small latent dimension might make it difficult for one CLS token to aggregate information about all classes (e.\,g. 23 classes in \mbox{ChimpACT}). To avoid information bottlenecks, we concatenate $C$ learned CLS tokens (one per class, initialized randomly) to the patch tokens. %

In summary, given $N$ patch tokens $x_i$ produced by the encoder \enct and $C$ trainable CLS tokens $q_j$, predicted class probabilities $\hat{y} \in [0, 1]^C$ are calculated as
\begin{align}
    \tilde x_i &= U \cdot \text{LayerNorm}(x_i) + b & i \in \{1, \dots, N\}\\
    X &= [q_1, \dots, q_C, \tilde x_1, \dots, \tilde x_N] & \\
    X' &= \text{SelfAttentionBlock}^3(X) & \\
    \hat{y}_j &= \sigma(v_j^\text{T} x'_j + c_j) & j \in \{1, \dots, C\}
\end{align}
with $\sigma$ being the \textit{softmax} function for single-label classification and the \textit{sigmoid} function for multi-label classification. Trainable parameters include all parameters of the self attention blocks as well as $U \in \Real^{D' \times D}$, $b \in \Real^{D'}$, $q_j \in \Real^{D'}$, $v_j \in \Real^{D'}$, and $c_j \in \Real$.

\section{Experiments}
\label{sec:experiments}

\subsection{Datasets and Evaluation Setup}
\label{subsec:eval-setup}
\label{subsec:dap-data}

\begin{figure*}[t]
    \centering
    \includegraphics[width=\linewidth]{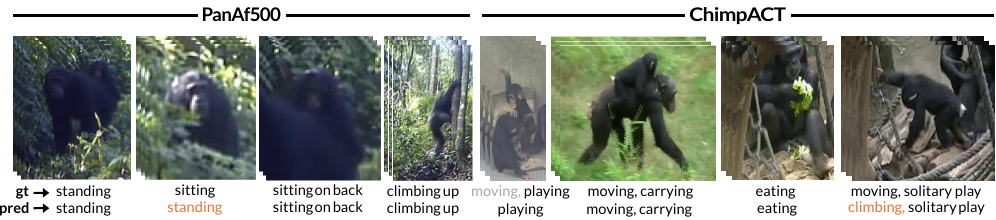}
    \caption{\textbf{Example predictions of our model on PanAf500 and ChimpACT.} We show  ground truth (gt) and predictions (pred) of the model pretrained on \prmt{} only.  Images: \cite{ma_chimpact_2023, brookes_panaf20k_2024} (\emph{val} sets).}
    \label{fig:eval_examples}
\end{figure*}

We evaluate on four primate video datasets: {PanAf500}~\cite{brookes_panaf20k_2024} (camera trap footage of chimpanzees and gorillas from tropical Africa), ChimpBehave~\cite{fuchs_forest_2025} (chimpanzees in an indoor enclosure at the Basel zoo), BaboonLand~\cite{duporge_baboonland_2025} (drone recording of olive baboons in Kenya), ChimpACT~\cite{ma_chimpact_2023} (chimpanzees at Leipzig zoo). Datasets feature between 7 and 23 action classes. ChimpACT has two behavior recognition protocols: One with and one without access to ground truth bounding boxes. PanAf500, ChimpBehave, and BaboonLand all define behavior recognition with access to ground truth bounding boxes.  See \cref{sec:supp_evaluation_datasets} for more details.

\paragraph{Evaluation Metrics}
Following the established evaluation protocols~\cite{duporge_baboonland_2025, brookes_panaf20k_2024, fuchs_forest_2025}, we report top-1 accuracy (Acc) and average per-class accuracy (balanced accuracy, B-Acc) for  PanAf500, BaboonLand, ChimpBehave. As \mbox{ChimpACT} is multi-label, it uses mAP instead, following the AVA protocol~\cite{gu_ava_2017}. In addition to mAP, which weights all classes equally, we also report weighted mean average  precision (mAP$_w$), where each class average precision (AP) is weighted by its support. We report mAP and mAP$_w$ over 23 classes. Thus, we report one metric weighting all classes equally (B-Acc, mAP) and one metric weighting classes by their support (Acc, mAP$_w$) for each dataset. BaboonLand and ChimpBehave do not contain a validation set. We therefore report ablations on \mbox{ChimpACT} and PanAf500 only.

\begin{table}[t]
    \nicetable
    \footnotesize
    \caption{\textbf{All parts of the \prmt{} dataset improve behavior recognition.} We ablate the pretraining data composition. We compare \textbf{(a)} human-centric data (\emph{V-JEPA}) and dataset-level pretraining (\dalp{}) with \textbf{(b)} uncurated YouTube videos ({YT-Random}), our datasets (\prmt{}, YT-Filtered, \rando{}) and additional \dalp{}. All experiments on \textit{val} sets using our attentive classifier, see \cref{fig:model}. Cross-domain comparisons in \crossdomain{gray}.}
    \begin{tabular}{l
    S[table-format=3.1]
    S[table-format=2.2]
    S[table-format=2.2]
    S[table-format=2.2]
    S[table-format=2.2]}
    \toprule
        \multirow{2}{*}{\thead{\textbf{Pretrain Data}}} & 
\multicolumn{1}{c}{\multirow{2}{*}{\thead{\textbf{Unique} \\ \textbf{Hours}}}} &  \multicolumn{2}{c}{\textbf{ChimpACT}}&  \multicolumn{2}{c}{\textbf{PanAf500}}\\
        \cmidrule(lr){3-4} \cmidrule(lr){5-6}
         &  &  {mAP}&  {mAP$_w$}&  {Acc}& {B-Acc}\\
         \midrule
         (\textit{V-JEPA}) &  {-} &  32.00& 47.88	 & 84.70 & 71.95 \\ %
        +\,\dalp{}:\,ChimpACT & 1.4& 35.86 & 51.12 &    \crossdomain{82.48}& \crossdomain{70.82}\\ %
        +\,\dalp{}:\,PanAf500  & 1.5& \crossdomain{28.94}& \crossdomain{45.22}& 88.50 & 78.57\\ %
         \midrule
         YT-Random& 280.0& 33.87 & 50.55 & 86.13 &	71.61 \\ %
         
         YT-Filtered  &250.0& 37.88 & 53.70& {89.17}	& 76.33 \\ %
         
         R\&O  & 174.0& 33.01 & 49.58 & 89.35 & 73.85 \\
         
         \textbf{\prmt{}} (\footnotesize{YT-F+\rando{}}) &424.0 & \underline{38.75} & \underline{54.32} & \underline{89.65} & \underline{79.95}\\
         
         +\,\dalp{}:\,ChimpACT& 1.4 & \textbf{41.43}  & \textbf{57.22} & \crossdomain{84.93} & \crossdomain{69.05} \\ 
         +\,\dalp{}:\,PanAf500&1.5& \crossdomain{32.90}	& \crossdomain{48.91}	 & \textbf{90.53} & \textbf{87.29}  \\ 
         
         \bottomrule
    \end{tabular}
    
    \label{tab:pretrain_data}
\end{table}

\subsection{Architecture and Pretraining Details}
\label{subsec:hyperparameters}

\paragraph{Domain-Level Pretraining}
Following V-JEPA~\cite{bardes_revisiting_2024}, we choose a ViT-L model with video input for $E$  and a narrow 12-layer transformer for the V-JEPA predictor $P$. $E$ has 304\,M parameters, $P$ has 22\,M parameters. We initialize the weights from a checkpoint pretrained on VideoMix2M, consisting of Kinetics, SomethingSomethingv2, and HowTo100M, and perform pretraining for 75\,k steps with a batch size of 80. This corresponds to approximately eight epochs on the \prmt{} dataset. As the initialized weights have already been cosine annealed, we only perform warmup and train with a constant small learning rate of \num{1.5e-5}, following \citet{singh_beyond_2025}. Training takes 11\,h on a single node with 4 A100 GPUs.

\begin{figure*}[t]
   \begin{minipage}[t]{0.65\textwidth}
        \nicetable
        \vspace{0pt}
        \captionof{table}{\textbf{Our method outperforms prior methods.} We compare our attentive classifier with V-JEPA pretrained on only human-centric data (\emph{V-JEPA}), pretrained on \prmt{}, and with additional dataset-level pretraining (\dalp{}) to various baseline and state-of-the-art methods. Results on \emph{test} sets.} 
        \begin{tabular}{l
            S[table-format=2.2]
            S[table-format=2.2]
            S[table-format=2.2]
            S[table-format=2.2]
            S[table-format=2.2]
            S[table-format=2.2]
            S[table-format=2.2]
            S[table-format=2.2]
            }
        \toprule
             &  \multicolumn{2}{c}{\textbf{ChimpACT}} & \multicolumn{2}{c}{\textbf{PanAf500}} & \multicolumn{2}{c}{\textbf{BaboonLand}} &  \multicolumn{2}{c}{\textbf{ChimpBehave}} \\
             \cmidrule(lr){2-3} \cmidrule(lr){4-5} \cmidrule(lr){6-7} \cmidrule(lr){8-9} 
             &  {mAP} & {mAP$_{w}$} & {Acc} & {B-Acc} & {Acc} & {B-Acc} & {Acc} & {B-Acc} \\
             \midrule
             X3D \hfill \cite{feichtenhofer_x3d_2020}&27.05 & 51.60&80.00&50.35& 64.89&31.41&89.3& 62.8\\
         VideoMAEv2 \hfill \cite{tong_videomae_2022}&&&&&&&92.3&\underline{74.8}\\
         UniformerV2-B \hfill \cite{li_uniformerv2_2023} &&&&& 63.45 & 28.67 && \\
              InternVideo-L \hfill \cite{wang_internvideo_2022}&25.7&&78.57&54.01&&&&\\
            ChimpVLM \hfill \cite{brookes_chimpvlm_2024}&&&84.91&61.94&&&&\\
          
            VideoPrism-g \hfill \cite{zhao_videoprism_2024}&31.5&&&&&&&\\

            \midrule
            \multicolumn{2}{l}{\textbf{Our Classifier}}&&&&&&&\\
            \; (\textit{V-JEPA}) & 36.33 & 55.50 &82.96 & 56.69& 74.91 &26.99	 & 94.99&  68.41\\
            \;  \textbf{\prmt{}} & \underline{39.25} & \underline{58.16}& \textbf{86.74} & \underline{62.75} &\underline{75.43} & \underline{33.99} & \underline{95.58} &	 71.30 
            \\
        \; \textbf{\prmt{} + \dalp{}} &  \textbf{40.00} & \textbf{59.29} & \underline{85.01} & \textbf{62.96} & \textbf{76.42} & \textbf{38.57} & \textbf{96.02} & \textbf{75.14} \\ %
            
            \bottomrule
        \end{tabular}
        
        \label{tab:main_comparison}
    \end{minipage}
    \hfill
    \begin{minipage}[t]{0.32\textwidth}
        \centering
        \vspace{0pt}
    \includegraphics[width=0.85\textwidth]{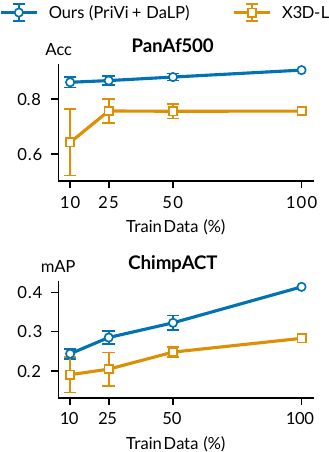}
    \captionof{figure}{\textbf{Our approach scales favorably with fewer labels.} Error bars are 95\,\% confidence intervals estimated over three subsets of the dataset. Results on \emph{val} sets.}
    \label{fig:head_data_ablations} 
    \end{minipage}
\end{figure*}

\paragraph{Dataset-level pretraining (\dalp{})}
\prmt{} includes diverse videos of primates, but not of the target datasets (ChimpACT, PanAf500, BaboonLand or ChimpBehave). To evaluate how much performance increase one could expect from perfectly in-domain unlabeled data, we also continue pretraining on the target dataset  \emph{after} pretraining on \prmt{}. Note that this pretraining is still entirely unsupervised, we do not use ground-truth bounding boxes. Chunking and primate detection is performed as for \rando{} (\cref{sec:data-curation-pipeline}) and we exclude the test set from pretraining except for ChimpBehave which does not have a dedicated test set. We pretrain as we do on \prmt{} but only train for 10k steps. 

\paragraph{Attentive Classifier}
For the attentive classifier, we always choose three layers, resulting in 220\,k parameters. When training it, we freeze the encoder $\enc$. On PanAf500, BaboonLand and ChimpBehave, we train for 40, 10 and 30 epochs, respectively, accounting for the different number of samples in each dataset. On ChimpACT, we train for only one epoch, as the dense annotation and miniclip sampling yields 500\,k highly-redundant miniclips. We use a learning rate of \num{1e-3} with warmup and cosine decay. 

On ChimpACT and PanAf500, we report performance at the best validation accuracy, for ChimpBehave and BaboonLand, we report performance at the end of training. We calculate class scores on three views per test sample. We always sample 16 frames uniformly from the input. We use cross-entropy loss except for BaboonLand where standard protocol is EQL loss~\cite{tan_equalization_2020}.

\subsection{Predicted Detections}

In this work, we mainly use ground truth detections to focus on behavior recognition and for compatibility with existing evaluation protocols.  However, for practical applications, end-to-end performance is highly relevant. To evaluate our approach for settings without ground truth detections, we apply the zero-shot detector SAM\,3~\cite{carion2025sam3segmentconcepts} and then apply our method on the predicted bounding boxes. We follow the recipe in \cref{subsec:hyperparameters}, except for classifier training, where we jointly train on the Chimp\-ACT train set and additional empty bounding boxes without any action labels. This is crucial to allow our model to deal with occasional false positive detections by SAM\,3.

\begin{table}[]
\centering
    \nicetable
    \caption{\textbf{Our method performs well on predicted bounding boxes.} AlphaChimp results are our reproduction due to incompatible evaluation protocols,
see \cref{sec:supp_evaluation}. Results are on \emph{val} sets. }
    \begin{tabular}{lrr}
    \toprule
   \textbf{ChimpACT (w/o GT detections)} &mAP  &mAP$_w$ \\
    \midrule 
    {AlphaChimp \hfill \cite{ma_alphachimp_2024}}& 25.35 & 40.23 \\ 
    \textbf{SAM3\,+\,Ours (\prmt{}\,+\,\dalp{})}  & \textbf{30.76}& \textbf{41.28} \\
 \bottomrule
 \end{tabular}

    \label{tab:scores_pred_bboxes}
\end{table}

\section{Results}

\subsection{\prmt{} pretraining improves performance over human-centric and dataset-level pretraining}
\label{sec:impact-dataset}

We first explore how varying pretraining data impacts behavior recognition for a fixed model (\cref{subsec:ssl}) and classifier (\cref{subsec:behavior-recognition}) architecture. We initialize using \mbox{V-JEPA} weights pretrained on VideoMix2M and then pretrain with only YT-Filtered, only \rando{} and both combined (i.\,e. \prmt{}). Both YT-Filtered and \rando{} individually improve performance over standard V-JEPA on both PanAf500 and ChimpACT, even though \rando{} shows smaller gains (\cref{tab:pretrain_data}). \rando{} contains camera trap recordings of chimpanzees, but no recordings of zoo-housed chimpanzees, which is mirrored in the larger performance improvements on PanAf500 compared to Chimp\-ACT (\cref{tab:pretrain_data}, \emph{V-JEPA} vs. \rando{}). Combining YT-Filtered and \rando{} to \prmt{} yields further gains on all metrics (\cref{tab:pretrain_data}), underscoring the importance of all parts of the dataset. \Cref{fig:eval_examples} shows example predictions of our model. Looking at per-action performance, we find an improvement in utilizing motion cues (e.g. distinguishing between climbing up, down, and hanging on PanAf500) over the V-JEPA baseline, suggesting better motion features in the backbone, see \cref{sec:supp_performance_breakdown} for the detailed breakdown.

\paragraph{YT relevance filtering boosts performance}
To measure the effect of our YouTube relevance filtering, we also compare to the baseline of randomly selected YouTube videos YT-Random. Our relevance filtering considerably improves performance (\cref{tab:pretrain_data}, YT-Filtered vs. YT-Random). To our surprise, even no relevance filtering still mostly outperforms the \emph{V-JEPA} baseline, suggesting that even noisy, partly primate-centric data is an improvement over VideoMix2M.

\paragraph{Broad \prmt{} pretraining outperforms dataset-level pretraining}
 We continue V-JEPA pretraining on only videos from {ChimpACT} and {PanAf500} (without labels and test samples) and compare with \prmt{}. Even though this dataset-level pretraining works, it is consistently outperformed by domain-level \prmt{} pretraining (\cref{tab:pretrain_data}). We do not observe transfer learning between \mbox{ChimpACT} and PanAf500 (\cref{tab:pretrain_data}, \crossdomain{gray} values), while \prmt{} pretraining performs well on both. %

\paragraph{Combining \prmt{} and dataset-level pretraining boosts performance further}
 Pretraining first on \prmt{} and then on the target datasets consistently improves performance, except for accuracy on PanAf500, where we see mixed results, likely due to the availability of chimpanzee camera trap recordings in \prmt{} (\cref{tab:pretrain_data} \& \ref{tab:main_comparison}; \prmt{} vs. \prmt{} + \dalp{}). \dalp{} deteriorates performance on other datasets, but this effect can be mitigated by jointly training on \prmt{} and the target dataset, see \cref{sec:supp_performance_breakdown}.

\subsection{Our method outperforms prior methods}
\label{sec:comparison-sota}

We compare our method against current state-of-the-art methods. Pretraining on \prmt{} followed by dataset-level pretraining (\prmt{} + \dalp{}) surpasses the state of the art on all four datasets across both class-balanced and unbalanced metrics (\cref{tab:main_comparison}). Our lightweight frozen attentive classifier using 220\,k parameters outperforms large specialist models like ChimpVLM with 167\,M parameters and full finetuning of human-centric models like VideoMAEv2. 

\paragraph{\prmt{} contributes across all datasets} When comparing V-JEPA pretrained on only human-centric data with pretraining on \prmt{}, \prmt{} brings considerable performance improvements across all datasets (\cref{tab:main_comparison}, Our Classifier).

\paragraph{Our method performs well on predicted bounding boxes} Despite using a zero-shot detector that was not trained on \mbox{Chimp}ACT, combining SAM\,3 with our method outperforms the current fully-finetuned state-of-the-art method AlphaChimp on ChimpACT without ground truth bounding boxes, see \cref{tab:scores_pred_bboxes}.

\subsection{Labeled Data Efficiency}
\label{sec:labeled-data-efficiency}

Next, we evaluate how well our method works on smaller datasets. We produce subsets of ChimpACT and PanAf500  with only 50\,\%, 25\,\%, and 10\,\% of training data, always including or excluding full sequences. See \cref{sec:supp_evaluation_datasets} for more details. PanAf500 contains 400 training video sequences, while ChimpACT contains only 127, making the task more challenging on ChimpACT. We compare against X3D~\cite{feichtenhofer_x3d_2020} as baseline.

We find that on PanAf500, our method scales very favorably with fewer labels, losing only 4.4\,\% accuracy from 100\,\% to 10\,\% training data. Ours with 10\,\% labeled data outperforms X3D with 100\,\% labeled data. X3D also scales very well up to 25\,\% without losing accuracy, but drops in performance at 10\,\%. On ChimpACT, both methods lose more performance when reducing labeled data, however our method at 25\,\% still outperforms X3D at 100\,\%. See \cref{fig:head_data_ablations}.

\subsection{Ablations}
\label{subsec:ablations}

Finally, we establish necessity of each component by ablating various design choices for pretraining and attentive classifier. Attentive classifiers without downprojection have far more parameters than our approach. To ensure a fair comparison, we explore learning rates between \num{1e-3} and \num{5e-5} and perform early stopping for each ablation.

\paragraph{Primate-centric cropping is beneficial} During pretraining, we crop videos to detected primate bounding boxes, because we assume that it is inefficient to spend computational resources on learning representations for background patches. Indeed, primate-centric cropping improves performance considerably on every metric compared with training on full video frames (\cref{tab:head_ablations}, w/o primate crop).

\paragraph{All components of our attentive classifier contribute} Removing the linear downprojection, reducing from three to one layers, and using one instead of $C$ class tokens reduces performance across all metrics (\cref{tab:head_ablations}), even though the effect of $C$ class tokens is small. Even using only one layer still achieves impressive performance with only 120\,k parameters, highlighting the quality of representation produced by our pretrained model.

\paragraph{Our pretrained model produces useful representations} A single cross-attention layer (\emph{Cross-Att.}) is commonly used for frozen evaluation to demonstrate representation quality \cite{bardes_revisiting_2024, zhao_videoprism_2024, sun_video_2024}. While this setup performs worse than our improved attentive classifier with more layers (except on PanAf500 balanced accuracy), it still shows good performance, demonstrating high feature quality.

\paragraph{Ours outperforms other deep attentive classifiers} \citet{assran_v-jepa_2025} present a deep attentive classifier with three self-attention layers followed by one cross-attention layer (\emph{Self- \& Cross-Att.}). Our models outperforms this design, despite theirs having 225 times more parameters. %

\begin{table}[]
    \nicetable
    \footnotesize
    \caption{\textbf{All components of our architecture contribute positively.} We ablate \textbf{(a)} pretraining and \textbf{(b)} classifier design decision and \textbf{(c)} compare our classifier against other common attentive classifiers. All models with pretraining on \prmt{} only, results reported on \textit{validation} sets.}
    \begin{tabular}{lrrrrr}
    \toprule
    &  \multirow{2}{*}{\thead{Train. \\ Params}} & \multicolumn{2}{c}{\thead{ChimpACT}}& \multicolumn{2}{c}{\thead{PanAf500}}\\
    \cmidrule(lr){3-4} \cmidrule(lr){5-6}
    &  &mAP  &mAP$_w$& Acc & B-Acc \\
    \midrule \midrule
    \textbf{Pretraining}&   &&&&\\
    \midrule
    \textbf{Ours}&  &\textbf{38.75}&\textbf{54.32}& \textbf{89.65}& \textbf{79.95}\\
    w/o primate crop& &32.62&47.74& 85.32& 72.93\\
    \midrule
    \midrule
    \textbf{Classifier}&   &&&&\\
    \midrule
    \textbf{Ours}&  0.22M&\textbf{38.75}&\textbf{54.32}& \textbf{89.65}& {79.95}\\
    \; w/o downproject &   37.84M&30.15&46.19&88.56&62.52\\
    \; single layer &  0.12M& 35.68	& 52.16& 89.21 & 75.81	\\
    \; single cls token &0.22M&38.54	&53.46	& 88.38	& 74.79\\
    \; single test-time view& 0.22M & 38.53 & 52.59 & 88.40 & 78.32 \\ 
    \midrule
    \multicolumn{1}{l}{\textbf{Baseline Classifier}} &&&&& \\
    Cross-Att. &  12.62M&32.02&50.64& 88.80 & \textbf{81.99} \\
        Self- \& Cross-Att.&   49.63M& 34.72 & 51.05 & 89.61 & 74.85 \\

 \bottomrule
 \end{tabular}
    
    \label{tab:head_ablations}
\end{table}

\section{Conclusion}
\label{sec:conclusion}

We introduced \prmt{}, a large-scale primate-centric pretraining dataset, and a simple framework for leveraging unlabeled video to improve behavior recognition. \prmt{} combines curated research footage from diverse species and settings with filtered YouTube videos, assembled through a scalable pipeline using CLIP-based relevance scoring and zero-shot primate detection. Pretraining V-JEPA on this dataset consistently boosts performance and surpasses prior work across all four benchmark datasets, showing the clear advantage of domain-level pretraining on primate videos over human-centric alternatives.

Our proposed framework consists of (a)~domain-level pretraining  on large-scale diverse animal-centric videos, (b)~optional dataset-level pretraining (\dalp{}) on the specific target dataset and (c)~frozen evaluation using a narrow, but deep attentive classifier. This framework has several desirable properties: Domain-level pretraining yields good behavior recognition performance across a wide variety of datasets; \dalp{} further boosts performance for individual datasets without requiring labeled data; and our frozen evaluation performs well even on few labeled samples. We hope that our framework and dataset will serve as a strong starting point for the community to move from specialist models towards unified primate behavior models, which could open new avenues for research in ethology, ecology, and conservation biology.

\paragraph{Limitations} Even though our dataset sets a new baseline for diversity in primate behavior datasets, it still captures only 11 different research setups. Due to the limited availability of labeled datasets, we could only evaluate on chimpanzees and baboons.

\section*{Acknowledgments}
{

\blind{The project was funded by the Deutsche Forschungsgemeinschaft (DFG, German Research Foundation) – ProjectID 454648639 – SFB 1528. Supported by the Deutsche Forschungsgemeinschaft (DFG, German Research Foundation) – GRK 2906 – project number 502807174 and GRK2070 - Project number 254142454. Data collection in Moyen Bafing was supported by the Leakey Foundation and the Wild Chimpanzee Foundation. The authors gratefully acknowledge the computing time granted by the Resource Allocation Board and provided on the supercomputer Emmy/Grete at NHR-Nord@Göttingen as part of the NHR infrastructure. The calculations for this research were conducted with computing resources under the project nib00021. }
}

{
    \small
    \bibliographystyle{ieeenat_fullname}
    \bibliography{main}

\begin{thebibliography}{74}
\providecommand{\natexlab}[1]{#1}
\providecommand{\url}[1]{\texttt{#1}}
\expandafter\ifx\csname urlstyle\endcsname\relax
  \providecommand{\doi}[1]{doi: #1}\else
  \providecommand{\doi}{doi: \begingroup \urlstyle{rm}\Url}\fi

\bibitem[Assran et~al.(2023)Assran, Duval, Misra, Bojanowski, Vincent, Rabbat,
  LeCun, and Ballas]{assran_self-supervised_2023}
Mahmoud Assran, Quentin Duval, Ishan Misra, Piotr Bojanowski, Pascal Vincent,
  Michael Rabbat, Yann LeCun, and Nicolas Ballas.
\newblock Self-supervised learning from images with a joint-embedding
  predictive architecture.
\newblock In \emph{Proceedings of the {IEEE}/{CVF} {Conference} on {Computer}
  {Vision} and {Pattern} {Recognition}}, pages 15619--15629, 2023.

\bibitem[Assran et~al.(2025)Assran, Bardes, Fan, Garrido, Howes, Komeili,
  Muckley, Rizvi, Roberts, Sinha, Zholus, Arnaud, Gejji, Martin, Hogan, Dugas,
  Bojanowski, Khalidov, Labatut, Massa, Szafraniec, Krishnakumar, Li, Ma,
  Chandar, Meier, LeCun, Rabbat, and Ballas]{assran_v-jepa_2025}
Mido Assran, Adrien Bardes, David Fan, Quentin Garrido, Russell Howes, Mojtaba
  Komeili, Matthew~J. Muckley, Ammar Rizvi, Claire Roberts, Koustuv Sinha,
  Artem Zholus, Sergio Arnaud, Abha Gejji, Ada Martin, Francois~Robert Hogan,
  Daniel Dugas, Piotr Bojanowski, Vasil Khalidov, Patrick Labatut, Francisco
  Massa, Marc Szafraniec, Kapil Krishnakumar, Yong Li, Xiaodong Ma, Sarath
  Chandar, Franziska Meier, Yann LeCun, Michael Rabbat, and Nicolas Ballas.
\newblock V-{JEPA} 2: {Self}-{Supervised} {Video} {Models} {Enable}
  {Understanding}, {Prediction} and {Planning}.
\newblock \emph{CoRR}, abs/2506.09985, 2025.
\newblock arXiv: 2506.09985.

\bibitem[Bain et~al.(2021)Bain, Nagrani, Schofield, Berdugo, Bessa, Owen,
  Hockings, Matsuzawa, Hayashi, Biro, Carvalho, and
  Zisserman]{bain_automated_2021}
Max Bain, Arsha Nagrani, Daniel Schofield, Sophie Berdugo, Joana Bessa, Jake
  Owen, Kimberley~J. Hockings, Tetsuro Matsuzawa, Misato Hayashi, Dora Biro,
  Susana Carvalho, and Andrew Zisserman.
\newblock Automated audiovisual behavior recognition in wild primates.
\newblock \emph{Science Advances}, 7\penalty0 (46):\penalty0 eabi4883, 2021.

\bibitem[Bardes et~al.(2024)Bardes, Garrido, Ponce, Chen, Rabbat, LeCun,
  Assran, and Ballas]{bardes_revisiting_2024}
Adrien Bardes, Quentin Garrido, Jean Ponce, Xinlei Chen, Michael Rabbat, Yann
  LeCun, Mido Assran, and Nicolas Ballas.
\newblock Revisiting {Feature} {Prediction} for {Learning} {Visual}
  {Representations} from {Video}.
\newblock \emph{Trans. Mach. Learn. Res.}, 2024, 2024.

\bibitem[Bolya et~al.(2025)Bolya, Huang, Sun, Cho, Madotto, Wei, Ma, Zhi,
  Rajasegaran, Rasheed, Wang, Monteiro, Xu, Dong, Ravi, Li, Dollár, and
  Feichtenhofer]{bolya_perception_2025}
Daniel Bolya, Po-Yao Huang, Peize Sun, Jang~Hyun Cho, Andrea Madotto, Chen Wei,
  Tengyu Ma, Jiale Zhi, Jathushan Rajasegaran, Hanoona Rasheed, Junke Wang,
  Marco Monteiro, Hu Xu, Shiyu Dong, Nikhila Ravi, Daniel Li, Piotr Dollár,
  and Christoph Feichtenhofer.
\newblock Perception {Encoder}: {The} best visual embeddings are not at the
  output of the network.
\newblock \emph{CoRR}, abs/2504.13181, 2025.
\newblock arXiv: 2504.13181.

\bibitem[Bosshard et~al.(2025)Bosshard, Mundry, and
  Fischer]{bosshard_ecological_2025}
Tiffany~Claire Bosshard, Roger Mundry, and Julia Fischer.
\newblock Ecological risk-taking across age in {{Barbary}} macaques.
\newblock \emph{Animal Behaviour}, 229:\penalty0 123337, 2025.

\bibitem[Brookes et~al.(2023)Brookes, Mirmehdi, Kühl, and
  Burghardt]{brookes_triple-stream_2023}
Otto Brookes, Majid Mirmehdi, Hjalmar~S. Kühl, and Tilo Burghardt.
\newblock Triple-stream {Deep} {Metric} {Learning} of {Great} {Ape}
  {Behavioural} {Actions}.
\newblock In \emph{Proceedings of the 18th {International} {Joint} {Conference}
  on {Computer} {Vision}, {Imaging} and {Computer} {Graphics} {Theory} and
  {Applications}, {VISIGRAPP} 2023, {Volume} 5: {VISAPP}, {Lisbon}, {Portugal},
  {February} 19-21, 2023}, pages 294--302. SCITEPRESS, 2023.

\bibitem[Brookes et~al.(2024{\natexlab{a}})Brookes, Mirmehdi, Kuhl, and
  Burghardt]{brookes_chimpvlm_2024}
Otto Brookes, Majid Mirmehdi, Hjalmar Kuhl, and Tilo Burghardt.
\newblock {ChimpVLM}: {Ethogram}-{Enhanced} {Chimpanzee} {Behaviour}
  {Recognition}, 2024{\natexlab{a}}.

\bibitem[Brookes et~al.(2024{\natexlab{b}})Brookes, Mirmehdi, Stephens,
  Angedakin, Corogenes, Dowd, Dieguez, Hicks, Jones, Lee, Leinert, Lapuente,
  McCarthy, Meier, Murai, Normand, Vergnes, Wessling, Wittig, Langergraber,
  Maldonado, Yang, Zuberbühler, Boesch, Arandjelovic, Kühl, and
  Burghardt]{brookes_panaf20k_2024}
Otto Brookes, Majid Mirmehdi, Colleen Stephens, Samuel Angedakin, Katherine
  Corogenes, Dervla Dowd, Paula Dieguez, Thurston~C. Hicks, Sorrel Jones, Kevin
  Lee, Vera Leinert, Juan Lapuente, Maureen~S. McCarthy, Amelia Meier, Mizuki
  Murai, Emmanuelle Normand, Virginie Vergnes, Erin~G. Wessling, Roman~M.
  Wittig, Kevin Langergraber, Nuria Maldonado, Xinyu Yang, Klaus Zuberbühler,
  Christophe Boesch, Mimi Arandjelovic, Hjalmar Kühl, and Tilo Burghardt.
\newblock {PanAf20K}: {A} {Large} {Video} {Dataset} for {Wild} {Ape}
  {Detection} and {Behaviour} {Recognition}.
\newblock \emph{International Journal of Computer Vision}, 2024{\natexlab{b}}.

\bibitem[Brookes et~al.(2025)Brookes, Kukushkin, Mirmehdi, Stephens, Dieguez,
  Hicks, Jones, Lee, McCarthy, Meier, Normand, Wessling, Wittig, Langergraber,
  Zuberbühler, Boesch, Schmid, Arandjelovic, Kühl, and
  Burghardt]{brookes_panaf-fgbg_2025}
Otto Brookes, Maksim Kukushkin, Majid Mirmehdi, Colleen Stephens, Paula
  Dieguez, Thurston~C. Hicks, Sorrel Jones, Kevin Lee, Maureen~S. McCarthy,
  Amelia Meier, Emmanuelle Normand, Erin~G. Wessling, Roman~M. Wittig, Kevin
  Langergraber, Klaus Zuberbühler, Lukas Boesch, Thomas Schmid, Mimi
  Arandjelovic, Hjalmar Kühl, and Tilo Burghardt.
\newblock The {PanAf}-{FGBG} {Dataset}: {Understanding} the {Impact} of
  {Backgrounds} in {Wildlife} {Behaviour} {Recognition}, 2025.

\bibitem[Brown et~al.(2020)Brown, Mann, Ryder, Subbiah, Kaplan, Dhariwal,
  Neelakantan, Shyam, Sastry, Askell, Agarwal, Herbert-Voss, Krueger, Henighan,
  Child, Ramesh, Ziegler, Wu, Winter, Hesse, Chen, Sigler, Litwin, Gray, Chess,
  Clark, Berner, McCandlish, Radford, Sutskever, and
  Amodei]{brown_language_2020}
Tom~B. Brown, Benjamin Mann, Nick Ryder, Melanie Subbiah, Jared Kaplan,
  Prafulla Dhariwal, Arvind Neelakantan, Pranav Shyam, Girish Sastry, Amanda
  Askell, Sandhini Agarwal, Ariel Herbert-Voss, Gretchen Krueger, T.~J.
  Henighan, Rewon Child, Aditya Ramesh, Daniel~M. Ziegler, Jeff Wu, Clemens
  Winter, Christopher Hesse, Mark Chen, Eric Sigler, Ma-teusz Litwin, Scott
  Gray, Benjamin Chess, Jack Clark, Christopher Berner, Sam McCandlish, Alec
  Radford, Ilya Sutskever, and Dario Amodei.
\newblock Language {Models} are {Few}-{Shot} {Learners}.
\newblock \emph{ArXiv}, abs/2005.14165, 2020.

\bibitem[Carion et~al.(2025)Carion, Gustafson, Hu, Debnath, Hu, Suris, Ryali,
  Alwala, Khedr, Huang, Lei, Ma, Guo, Kalla, Marks, Greer, Wang, Sun, Rädle,
  Afouras, Mavroudi, Xu, Wu, Zhou, Momeni, Hazra, Ding, Vaze, Porcher, Li, Li,
  Kamath, Cheng, Dollár, Ravi, Saenko, Zhang, and
  Feichtenhofer]{carion2025sam3segmentconcepts}
Nicolas Carion, Laura Gustafson, Yuan-Ting Hu, Shoubhik Debnath, Ronghang Hu,
  Didac Suris, Chaitanya Ryali, Kalyan~Vasudev Alwala, Haitham Khedr, Andrew
  Huang, Jie Lei, Tengyu Ma, Baishan Guo, Arpit Kalla, Markus Marks, Joseph
  Greer, Meng Wang, Peize Sun, Roman Rädle, Triantafyllos Afouras, Effrosyni
  Mavroudi, Katherine Xu, Tsung-Han Wu, Yu Zhou, Liliane Momeni, Rishi Hazra,
  Shuangrui Ding, Sagar Vaze, Francois Porcher, Feng Li, Siyuan Li, Aishwarya
  Kamath, Ho~Kei Cheng, Piotr Dollár, Nikhila Ravi, Kate Saenko, Pengchuan
  Zhang, and Christoph Feichtenhofer.
\newblock Sam 3: Segment anything with concepts, 2025.

\bibitem[Caron et~al.(2021)Caron, Touvron, Misra, Jégou, Mairal, Bojanowski,
  and Joulin]{caron_emerging_2021}
Mathilde Caron, Hugo Touvron, Ishan Misra, Hervé Jégou, Julien Mairal, Piotr
  Bojanowski, and Armand Joulin.
\newblock Emerging {Properties} in {Self}-{Supervised} {Vision} {Transformers}.
\newblock In \emph{2021 {IEEE}/{CVF} {International} {Conference} on {Computer}
  {Vision}, {ICCV} 2021, {Montreal}, {QC}, {Canada}, {October} 10-17, 2021},
  pages 9630--9640. IEEE, 2021.

\bibitem[Castellano()]{castellano_pyscenedetect_nodate}
Brandon Castellano.
\newblock {PySceneDetect}.

\bibitem[Chen et~al.(2023)Chen, Hu, Coker, Berumen, Costelloe, Beery, Rohrbach,
  and Elhoseiny]{chen_mammalnet_2023}
Jun Chen, Ming Hu, Darren~J. Coker, Michael~L. Berumen, Blair~R. Costelloe,
  Sara Beery, Anna Rohrbach, and Mohamed Elhoseiny.
\newblock {MammalNet}: {A} {Large}-{Scale} {Video} {Benchmark} for {Mammal}
  {Recognition} and {Behavior} {Understanding}.
\newblock In \emph{{IEEE}/{CVF} {Conference} on {Computer} {Vision} and
  {Pattern} {Recognition}, {CVPR} 2023, {Vancouver}, {BC}, {Canada}, {June}
  17-24, 2023}, pages 13052--13061. IEEE, 2023.

\bibitem[Chen et~al.(2020)Chen, Kornblith, Norouzi, and
  Hinton]{chen_simple_2020}
Ting Chen, Simon Kornblith, Mohammad Norouzi, and Geoffrey~E. Hinton.
\newblock A {Simple} {Framework} for {Contrastive} {Learning} of {Visual}
  {Representations}.
\newblock In \emph{Proceedings of the 37th {International} {Conference} on
  {Machine} {Learning}, {ICML} 2020, 13-18 {July} 2020, {Virtual} {Event}},
  pages 1597--1607. PMLR, 2020.

\bibitem[Devlin et~al.(2019)Devlin, Chang, Lee, and
  Toutanova]{devlin_bert_2019}
Jacob Devlin, Ming-Wei Chang, Kenton Lee, and Kristina Toutanova.
\newblock {BERT}: {Pre}-training of {Deep} {Bidirectional} {Transformers} for
  {Language} {Understanding}.
\newblock In \emph{Proceedings of the 2019 {Conference} of the {North}
  {American} {Chapter} of the {Association} for {Computational} {Linguistics}:
  {Human} {Language} {Technologies}, {NAACL}-{HLT} 2019, {Minneapolis}, {MN},
  {USA}, {June} 2-7, 2019, {Volume} 1 ({Long} and {Short} {Papers})}, pages
  4171--4186. Association for Computational Linguistics, 2019.

\bibitem[Duporge et~al.(2025)Duporge, Kholiavchenko, Harel, Wolf, Rubenstein,
  Crofoot, Berger-Wolf, Lee, Barreau, Kline, Ramirez, and
  Stewart]{duporge_baboonland_2025}
Isla Duporge, Maksim Kholiavchenko, Roi Harel, Scott Wolf, Daniel~I.
  Rubenstein, Margaret~C. Crofoot, Tanya~Y. Berger-Wolf, Stephen~J. Lee, Julie
  Barreau, Jenna Kline, Michelle Ramirez, and Charles~V. Stewart.
\newblock {BaboonLand} {Dataset}: {Tracking} {Primates} in the {Wild} and
  {Automating} {Behaviour} {Recognition} from {Drone} {Videos}.
\newblock \emph{Int. J. Comput. Vis.}, 133\penalty0 (9):\penalty0 6578--6589,
  2025.

\bibitem[Feichtenhofer(2020)]{feichtenhofer_x3d_2020}
Christoph Feichtenhofer.
\newblock {X3D}: {Expanding} {Architectures} for {Efficient} {Video}
  {Recognition}.
\newblock In \emph{2020 {IEEE}/{CVF} {Conference} on {Computer} {Vision} and
  {Pattern} {Recognition}, {CVPR} 2020, {Seattle}, {WA}, {USA}, {June} 13-19,
  2020}, pages 200--210. Computer Vision Foundation / IEEE, 2020.

\bibitem[Feichtenhofer et~al.(2022)Feichtenhofer, Fan, Li, and
  He]{feichtenhofer_masked_2022}
Christoph Feichtenhofer, Haoqi Fan, Yanghao Li, and Kaiming He.
\newblock Masked {Autoencoders} {As} {Spatiotemporal} {Learners}.
\newblock \emph{Advances in Neural Information Processing Systems},
  35:\penalty0 35946--35958, 2022.

\bibitem[Fuchs et~al.(2023)Fuchs, Genty, Zuberbühler, and
  Cotofrei]{fuchs_asbar_2023}
Michael Fuchs, Emilie Genty, Klaus Zuberbühler, and Paul Cotofrei.
\newblock {ASBAR}: an {Animal} {Skeleton}-{Based} {Action} {Recognition}
  framework. {Recognizing} great ape behaviors in the wild using pose
  estimation with domain adaptation, 2023.

\bibitem[Fuchs et~al.(2025)Fuchs, Genty, Bangerter, Zuberbühler, Odobez, and
  Cotofrei]{fuchs_forest_2025}
Michael Fuchs, Emilie Genty, Adrian Bangerter, Klaus Zuberbühler, Jean-Marc
  Odobez, and Paul Cotofrei.
\newblock From {Forest} to {Zoo}: {Great} {Ape} {Behavior} {Recognition} with
  {ChimpBehave}.
\newblock \emph{Int. J. Comput. Vis.}, 133\penalty0 (10):\penalty0 6668--6688,
  2025.

\bibitem[Gabeff et~al.(2025)Gabeff, Qi, Flaherty, Sumbul, Mathis, and
  Tuia]{gabeff_mammalps_2025}
Valentin Gabeff, Haozhe Qi, Brendan Flaherty, Gencer Sumbul, Alexander Mathis,
  and Devis Tuia.
\newblock {MammAlps}: {A} {Multi}-view {Video} {Behavior} {Monitoring}
  {Dataset} of {Wild} {Mammals} in the {Swiss} {Alps}.
\newblock In \emph{{IEEE}/{CVF} {Conference} on {Computer} {Vision} and
  {Pattern} {Recognition}, {CVPR} 2025, {Nashville}, {TN}, {USA}, {June} 11-15,
  2025}, pages 13854--13864. Computer Vision Foundation / IEEE, 2025.

\bibitem[Gadre et~al.(2023)Gadre, Ilharco, Fang, Hayase, Smyrnis, Nguyen,
  Marten, Wortsman, Ghosh, Zhang, Orgad, Entezari, Daras, Pratt, Ramanujan,
  Bitton, Marathe, Mussmann, Vencu, Cherti, Krishna, Koh, Saukh, Ratner, Song,
  Hajishirzi, Farhadi, Beaumont, Oh, Dimakis, Jitsev, Carmon, Shankar, and
  Schmidt]{gadre_datacomp_2023}
Samir~Yitzhak Gadre, Gabriel Ilharco, Alex Fang, Jonathan Hayase, Georgios
  Smyrnis, Thao Nguyen, Ryan Marten, Mitchell Wortsman, Dhruba Ghosh, Jieyu
  Zhang, Eyal Orgad, Rahim Entezari, Giannis Daras, Sarah~M. Pratt, Vivek
  Ramanujan, Yonatan Bitton, Kalyani Marathe, Stephen Mussmann, Richard Vencu,
  Mehdi Cherti, Ranjay Krishna, Pang~Wei Koh, Olga Saukh, Alexander~J. Ratner,
  Shuran Song, Hannaneh Hajishirzi, Ali Farhadi, Romain Beaumont, Sewoong Oh,
  Alex Dimakis, Jenia Jitsev, Yair Carmon, Vaishaal Shankar, and Ludwig
  Schmidt.
\newblock {DataComp}: {In} search of the next generation of multimodal
  datasets.
\newblock In \emph{Advances in {Neural} {Information} {Processing} {Systems}
  36: {Annual} {Conference} on {Neural} {Information} {Processing} {Systems}
  2023, {NeurIPS} 2023, {New} {Orleans}, {LA}, {USA}, {December} 10 - 16,
  2023}, 2023.

\bibitem[Gu et~al.(2017)Gu, Sun, Ross, Vondrick, Pantofaru, Li,
  Vijayanarasimhan, Toderici, Ricco, Sukthankar, Schmid, and
  Malik]{gu_ava_2017}
Chunhui Gu, Chen Sun, David~A. Ross, Carl Vondrick, Caroline Pantofaru, Yeqing
  Li, Sudheendra Vijayanarasimhan, George Toderici, Susanna Ricco, Rahul
  Sukthankar, Cordelia Schmid, and Jitendra Malik.
\newblock {AVA}: {A} {Video} {Dataset} of {Spatio}-temporally {Localized}
  {Atomic} {Visual} {Actions}, 2017.

\bibitem[Gururangan et~al.(2020)Gururangan, Marasovic, Swayamdipta, Lo,
  Beltagy, Downey, and Smith]{DBLP:conf/acl/GururanganMSLBD20}
Suchin Gururangan, Ana Marasovic, Swabha Swayamdipta, Kyle Lo, Iz Beltagy, Doug
  Downey, and Noah~A. Smith.
\newblock Don't stop pretraining: Adapt language models to domains and tasks.
\newblock In \emph{Proceedings of the 58th Annual Meeting of the Association
  for Computational Linguistics, {ACL} 2020, Online, July 5-10, 2020}, pages
  8342--8360. Association for Computational Linguistics, 2020.

\bibitem[He et~al.(2021)He, Chen, Xie, Li, Doll'ar, and
  Girshick]{he_masked_2021}
Kaiming He, Xinlei Chen, Saining Xie, Yanghao Li, Piotr Doll'ar, and Ross~B.
  Girshick.
\newblock Masked {Autoencoders} {Are} {Scalable} {Vision} {Learners}.
\newblock \emph{2022 IEEE/CVF Conference on Computer Vision and Pattern
  Recognition (CVPR)}, pages 15979--15988, 2021.

\bibitem[Hinton and Salakhutdinov(2006)]{hinton_reducing_2006}
Geoffrey~E Hinton and Ruslan~R Salakhutdinov.
\newblock Reducing the dimensionality of data with neural networks.
\newblock \emph{Science}, 313\penalty0 (5786):\penalty0 504--507, 2006.
\newblock Publisher: American Association for the Advancement of Science.

\bibitem[Hinton et~al.(2006)Hinton, Osindero, and Teh]{hinton_fast_2006}
Geoffrey~E Hinton, Simon Osindero, and Yee-Whye Teh.
\newblock A fast learning algorithm for deep belief nets.
\newblock \emph{Neural computation}, 18\penalty0 (7):\penalty0 1527--1554,
  2006.
\newblock Publisher: MIT Press.

\bibitem[Iashin et~al.(2025)Iashin, Lee, Schofield, and
  Zisserman]{iashin_self-supervised_2025}
Vladimir Iashin, Horace Lee, Dan Schofield, and Andrew Zisserman.
\newblock Self-supervised {Learning} on {Camera} {Trap} {Footage} {Yields} a
  {Strong} {Universal} {Face} {Embedder}.
\newblock \emph{CoRR}, abs/2507.10552, 2025.
\newblock arXiv: 2507.10552.

\bibitem[Karako{\c c} et~al.(2025)Karako{\c c}, Vogg, Marziliano, {von
  Petersdorff}, Ecker, Kappeler, and Fichtel]{karakoc_foraging_2025}
Elif Karako{\c c}, Richard Vogg, Michele Marziliano, Jacob {von Petersdorff},
  Alexander~S. Ecker, Peter~M. Kappeler, and Claudia Fichtel.
\newblock Foraging competence and its impact on social relationships in a
  socially tolerant wild primate.
\newblock \emph{Animal Cognition}, 28\penalty0 (1):\penalty0 86, 2025.

\bibitem[Kay et~al.(2017)Kay, Carreira, Simonyan, Zhang, Hillier,
  Vijayanarasimhan, Viola, Green, Back, Natsev, Suleyman, and
  Zisserman]{kay_kinetics_2017}
Will Kay, João Carreira, Karen Simonyan, Brian Zhang, Chloe Hillier,
  Sudheendra Vijayanarasimhan, Fabio Viola, Tim Green, Trevor Back, Paul
  Natsev, Mustafa Suleyman, and Andrew Zisserman.
\newblock The {Kinetics} {Human} {Action} {Video} {Dataset}.
\newblock \emph{CoRR}, abs/1705.06950, 2017.
\newblock arXiv: 1705.06950.

\bibitem[Kholiavchenko et~al.(2024)Kholiavchenko, Kline, Ramirez, Stevens,
  Sheets, Babu, Banerji, Campolongo, Thompson, Van~Tiel, Miliko, Bessa,
  Duporge, Berger-Wolf, Rubenstein, and Stewart]{kholiavchenko_kabr_2024}
Maksim Kholiavchenko, Jenna Kline, Michelle Ramirez, Sam Stevens, Alec Sheets,
  Reshma Babu, Namrata Banerji, Elizabeth Campolongo, Matthew Thompson, Nina
  Van~Tiel, Jackson Miliko, Eduardo Bessa, Isla Duporge, Tanya Berger-Wolf,
  Daniel Rubenstein, and Charles Stewart.
\newblock {KABR}: {In}-{Situ} {Dataset} for {Kenyan} {Animal} {Behavior}
  {Recognition} from {Drone} {Videos}.
\newblock In \emph{Proceedings of the {IEEE}/{CVF} {Winter} {Conference} on
  {Applications} of {Computer} {Vision}}, pages 31--40, 2024.

\bibitem[Kingma and Welling(2014)]{kingma_auto-encoding_2014}
Diederik~P. Kingma and Max Welling.
\newblock Auto-{Encoding} {Variational} {Bayes}.
\newblock In \emph{2nd {International} {Conference} on {Learning}
  {Representations}, {ICLR} 2014, {Banff}, {AB}, {Canada}, {April} 14-16, 2014,
  {Conference} {Track} {Proceedings}}, 2014.

\bibitem[Kirillov et~al.(2023)Kirillov, Mintun, Ravi, Mao, Rolland, Gustafson,
  Xiao, Whitehead, Berg, Lo, Dollár, and Girshick]{kirillov_segment_2023}
Alexander Kirillov, Eric Mintun, Nikhila Ravi, Hanzi Mao, Chloé Rolland, Laura
  Gustafson, Tete Xiao, Spencer Whitehead, Alexander~C. Berg, Wan-Yen Lo, Piotr
  Dollár, and Ross~B. Girshick.
\newblock Segment {Anything}.
\newblock In \emph{{IEEE}/{CVF} {International} {Conference} on {Computer}
  {Vision}, {ICCV} 2023, {Paris}, {France}, {October} 1-6, 2023}, pages
  3992--4003. IEEE, 2023.

\bibitem[Kuznetsova et~al.(2020)Kuznetsova, Rom, Alldrin, Uijlings, Krasin,
  Pont-Tuset, Kamali, Popov, Malloci, Kolesnikov, Duerig, and
  Ferrari]{kuznetsova_open_2020}
Alina Kuznetsova, Hassan Rom, Neil Alldrin, Jasper R.~R. Uijlings, Ivan Krasin,
  Jordi Pont-Tuset, Shahab Kamali, Stefan Popov, Matteo Malloci, Alexander
  Kolesnikov, Tom Duerig, and Vittorio Ferrari.
\newblock The {Open} {Images} {Dataset} {V4}.
\newblock \emph{Int. J. Comput. Vis.}, 128\penalty0 (7):\penalty0 1956--1981,
  2020.

\bibitem[Li et~al.(2023)Li, Wang, He, Li, Wang, Wang, and
  Qiao]{li_uniformerv2_2023}
Kunchang Li, Yali Wang, Yinan He, Yizhuo Li, Yi Wang, Limin Wang, and Yu Qiao.
\newblock {UniFormerV2}: {Unlocking} the {Potential} of {Image} {ViTs} for
  {Video} {Understanding}.
\newblock In \emph{{IEEE}/{CVF} {International} {Conference} on {Computer}
  {Vision}, {ICCV} 2023, {Paris}, {France}, {October} 1-6, 2023}, pages
  1632--1643. IEEE, 2023.

\bibitem[Liu et~al.(2024)Liu, Zeng, Ren, Li, Zhang, Yang, Jiang, Li, Yang, Su,
  Zhu, and Zhang]{liu_grounding_2024}
Shilong Liu, Zhaoyang Zeng, Tianhe Ren, Feng Li, Hao Zhang, Jie Yang, Qing
  Jiang, Chunyuan Li, Jianwei Yang, Hang Su, Jun Zhu, and Lei Zhang.
\newblock Grounding {DINO}: {Marrying} {DINO} with {Grounded} {Pre}-training
  for {Open}-{Set} {Object} {Detection}.
\newblock In \emph{Computer {Vision} - {ECCV} 2024 - 18th {European}
  {Conference}, {Milan}, {Italy}, {September} 29-{October} 4, 2024,
  {Proceedings}, {Part} {XLVII}}, pages 38--55. Springer, 2024.

\bibitem[Lüddecke and Ecker(2022)]{luddecke_image_2022}
Timo Lüddecke and Alexander Ecker.
\newblock Image {Segmentation} {Using} {Text} and {Image} {Prompts}.
\newblock pages 7086--7096, 2022.

\bibitem[Ma et~al.(2023)Ma, Kaufhold, Su, Zhu, Terwilliger, Meza, Zhu, Rossano,
  and Wang]{ma_chimpact_2023}
Xiaoxuan Ma, Stephan~P. Kaufhold, Jiajun Su, Wentao Zhu, Jack Terwilliger,
  Andres Meza, Yixin Zhu, Federico Rossano, and Yizhou Wang.
\newblock {ChimpACT}: {A} {Longitudinal} {Dataset} for {Understanding}
  {Chimpanzee} {Behaviors}, 2023.

\bibitem[Ma et~al.(2024)Ma, Lin, Xu, Kaufhold, Terwilliger, Meza, Zhu, Rossano,
  and Wang]{ma_alphachimp_2024}
Xiaoxuan Ma, Yutang Lin, Yuan Xu, Stephan~P. Kaufhold, Jack Terwilliger, Andres
  Meza, Yixin Zhu, Federico Rossano, and Yizhou Wang.
\newblock {AlphaChimp}: {Tracking} and {Behavior} {Recognition} of
  {Chimpanzees}, 2024.

\bibitem[Mamooler et~al.(2025)Mamooler, Qi, Gabeff, Montariol, Bosselut, and
  Mathis]{mamooler_fine-tuning_2025}
Sepideh Mamooler, Haozhe Qi, Valentin Gabeff, Syrielle Montariol, Antoine
  Bosselut, and Alexander Mathis.
\newblock Fine-tuning {Vision}-{Language} {Models} for {Animal} {Behavior}
  {Analysis}.
\newblock In \emph{{LLM} for {Scientific} {Discovery}: {Reasoning},
  {Assistance}, and {Collaboration}}, 2025.

\bibitem[Miech et~al.(2019)Miech, Zhukov, Alayrac, Tapaswi, Laptev, and
  Sivic]{miech_howto100m_2019}
Antoine Miech, Dimitri Zhukov, Jean-Baptiste Alayrac, Makarand Tapaswi, Ivan
  Laptev, and Josef Sivic.
\newblock {HowTo100M}: {Learning} a {Text}-{Video} {Embedding} by {Watching}
  {Hundred} {Million} {Narrated} {Video} {Clips}.
\newblock In \emph{2019 {IEEE}/{CVF} {International} {Conference} on {Computer}
  {Vision}, {ICCV} 2019, {Seoul}, {Korea} ({South}), {October} 27 - {November}
  2, 2019}, pages 2630--2640. IEEE, 2019.

\bibitem[Mueller et~al.(2025)Mueller, Lüddecke, Vogg, and
  Ecker]{mueller_domain-adaptive_2025}
Felix~B. Mueller, Timo Lüddecke, Richard Vogg, and Alexander~S. Ecker.
\newblock Domain-{Adaptive} {Pretraining} {Improves} {Primate} {Behavior}
  {Recognition}.
\newblock \emph{CoRR}, abs/2509.12193, 2025.
\newblock arXiv: 2509.12193.

\bibitem[Ng et~al.(2022)Ng, Ong, Zheng, Ni, Yeo, and Liu]{ng_animal_2022}
Xun~Long Ng, Kian~Eng Ong, Qichen Zheng, Yun Ni, Si~Yong Yeo, and Jun Liu.
\newblock Animal {Kingdom}: {A} {Large} and {Diverse} {Dataset} for {Animal}
  {Behavior} {Understanding}.
\newblock pages 19023--19034, 2022.

\bibitem[O'Hearn et~al.(2024)O'Hearn, Frank, Keupp, and
  Fischer]{ohearn_lessons_2024}
William O'Hearn, Louis Frank, Stefanie Keupp, and Julia Fischer.
\newblock Lessons learned from a cooperative box experiment with wild baboons.
\newblock \emph{Animal Behavior and Cognition}, 11\penalty0 (4):\penalty0
  330--348, 2024.

\bibitem[O'Hearn et~al.(2025)O'Hearn, Beckmann, Von~Fersen, Dal~Pesco, Mundry,
  Keupp, Diakhate, Niederbremer, and Fischer]{ohearn_increased_2025}
William~John O'Hearn, J{\"o}rg Beckmann, Lorenzo Von~Fersen, Federica
  Dal~Pesco, Roger Mundry, Stefanie Keupp, Ndiouga Diakhate, Carolin
  Niederbremer, and Julia Fischer.
\newblock Increased female competition for males with enhanced foraging skills
  in {{Guinea}} baboons.
\newblock \emph{Proceedings of the Royal Society B: Biological Sciences},
  292\penalty0 (2042):\penalty0 20242925, 2025.

\bibitem[Oquab et~al.(2024)Oquab, Darcet, Moutakanni, Vo, Szafraniec, Khalidov,
  Fernandez, Haziza, Massa, El-Nouby, Assran, Ballas, Galuba, Howes, Huang, Li,
  Misra, Rabbat, Sharma, Synnaeve, Xu, Jégou, Mairal, Labatut, Joulin, and
  Bojanowski]{oquab_dinov2_2024}
Maxime Oquab, Timothée Darcet, Théo Moutakanni, Huy~V. Vo, Marc Szafraniec,
  Vasil Khalidov, Pierre Fernandez, Daniel Haziza, Francisco Massa, Alaaeldin
  El-Nouby, Mido Assran, Nicolas Ballas, Wojciech Galuba, Russell Howes, Po-Yao
  Huang, Shang-Wen Li, Ishan Misra, Michael Rabbat, Vasu Sharma, Gabriel
  Synnaeve, Hu Xu, Hervé Jégou, Julien Mairal, Patrick Labatut, Armand
  Joulin, and Piotr Bojanowski.
\newblock {DINOv2}: {Learning} {Robust} {Visual} {Features} without
  {Supervision}.
\newblock \emph{Trans. Mach. Learn. Res.}, 2024, 2024.

\bibitem[Pereira et~al.(2025)Pereira, Meesawat, Malaivijitnond, Ostner, and
  Sch{\"u}lke]{pereira_complex_2025}
Sofia~M. Pereira, Suthirote Meesawat, Suchinda Malaivijitnond, Julia Ostner,
  and Oliver Sch{\"u}lke.
\newblock Complex modulation of visual attention to 3rd person interactions in
  wild macaques, 2025.

\bibitem[Radford and Narasimhan(2018)]{radford_improving_2018}
Alec Radford and Karthik Narasimhan.
\newblock Improving {Language} {Understanding} by {Generative}
  {Pre}-{Training}.
\newblock 2018.

\bibitem[Radford et~al.(2021)Radford, Kim, Hallacy, Ramesh, Goh, Agarwal,
  Sastry, Askell, Mishkin, Clark, and {others}]{radford_learning_2021}
Alec Radford, Jong~Wook Kim, Chris Hallacy, Aditya Ramesh, Gabriel Goh,
  Sandhini Agarwal, Girish Sastry, Amanda Askell, Pamela Mishkin, Jack Clark,
  and {others}.
\newblock Learning transferable visual models from natural language
  supervision.
\newblock In \emph{International conference on machine learning}, pages
  8748--8763. PmLR, 2021.

\bibitem[Ravi et~al.(2025)Ravi, Gabeur, Hu, Hu, Ryali, Ma, Khedr, Rädle,
  Rolland, Gustafson, Mintun, Pan, Alwala, Carion, Wu, Girshick, Dollár, and
  Feichtenhofer]{ravi_sam_2025}
Nikhila Ravi, Valentin Gabeur, Yuan-Ting Hu, Ronghang Hu, Chaitanya Ryali,
  Tengyu Ma, Haitham Khedr, Roman Rädle, Chloé Rolland, Laura Gustafson, Eric
  Mintun, Junting Pan, Kalyan~Vasudev Alwala, Nicolas Carion, Chao-Yuan Wu,
  Ross~B. Girshick, Piotr Dollár, and Christoph Feichtenhofer.
\newblock {SAM} 2: {Segment} {Anything} in {Images} and {Videos}.
\newblock In \emph{The {Thirteenth} {International} {Conference} on {Learning}
  {Representations}, {ICLR} 2025, {Singapore}, {April} 24-28, 2025}.
  OpenReview.net, 2025.

\bibitem[Rezende et~al.(2014)Rezende, Mohamed, and
  Wierstra]{rezende_stochastic_2014}
Danilo~Jimenez Rezende, Shakir Mohamed, and Daan Wierstra.
\newblock Stochastic {Backpropagation} and {Approximate} {Inference} in {Deep}
  {Generative} {Models}.
\newblock In \emph{Proceedings of the 31th {International} {Conference} on
  {Machine} {Learning}, {ICML} 2014, {Beijing}, {China}, 21-26 {June} 2014},
  pages 1278--1286. JMLR.org, 2014.

\bibitem[Rodriguez-Juan et~al.(2025)Rodriguez-Juan, Ortiz-Perez,
  Benavent-Lledo, Mulero-Pérez, Ruiz-Ponce, Orihuela-Torres, Garcia-Rodriguez,
  and Sebastián-González]{rodriguez-juan_visual_2025}
Javier Rodriguez-Juan, David Ortiz-Perez, Manuel Benavent-Lledo, David
  Mulero-Pérez, Pablo Ruiz-Ponce, Adrian Orihuela-Torres, Jose
  Garcia-Rodriguez, and Esther Sebastián-González.
\newblock Visual {WetlandBirds} {Dataset}: {Bird} {Species} {Identification}
  and {Behavior} {Recognition} in {Videos}, 2025.

\bibitem[Ryan et~al.(2025)Ryan, Bati, Lee, Bolya, Hoffman, and
  Rehg]{ryan_gaze-lle_2025}
Fiona Ryan, Ajay Bati, Sangmin Lee, Daniel Bolya, Judy Hoffman, and James~M.
  Rehg.
\newblock Gaze-{LLE}: {Gaze} {Target} {Estimation} via {Large}-{Scale}
  {Learned} {Encoders}.
\newblock pages 28874--28884, 2025.

\bibitem[Sakib and Burghardt(2020)]{sakib_visual_2020}
Faizaan Sakib and Tilo Burghardt.
\newblock Visual {Recognition} of {Great} {Ape} {Behaviours} in the {Wild}.
\newblock \emph{CoRR}, abs/2011.10759, 2020.
\newblock arXiv: 2011.10759.

\bibitem[Salehi et~al.(2024)Salehi, Dorkenwald, Thoker, Gavves, Snoek, and
  Asano]{salehi_sigma_2024}
Mohammadreza Salehi, Michael Dorkenwald, Fida~Mohammad Thoker, Efstratios
  Gavves, Cees~GM Snoek, and Yuki~M Asano.
\newblock {SIGMA}: {Sinkhorn}-{Guided} {Masked} {Video} {Modeling}.
\newblock In \emph{European {Conference} on {Computer} {Vision}}, pages
  293--312. Springer, 2024.

\bibitem[Santo et~al.(2025)Santo, Izar, Delval, Gregolin, and
  Hirata]{santo_fine-tuning_2025}
Giulio Cesare~Mastrocinque Santo, Patrícia Izar, Irene Delval, Victor
  de~Napole Gregolin, and Nina S.~T. Hirata.
\newblock Fine-{Tuning} {Video}-{Text} {Contrastive} {Model} for {Primate}
  {Behavior} {Retrieval} from {Unlabeled} {Raw} {Videos}.
\newblock \emph{CoRR}, abs/2505.05681, 2025.
\newblock arXiv: 2505.05681.

\bibitem[Schuhmann et~al.(2022)Schuhmann, Beaumont, Vencu, Gordon, Wightman,
  Cherti, Coombes, Katta, Mullis, Wortsman, Schramowski, Kundurthy, Crowson,
  Schmidt, Kaczmarczyk, and Jitsev]{schuhmann_laion-5b_2022}
Christoph Schuhmann, Romain Beaumont, Richard Vencu, Cade Gordon, Ross
  Wightman, Mehdi Cherti, Theo Coombes, Aarush Katta, Clayton Mullis, Mitchell
  Wortsman, Patrick Schramowski, Srivatsa Kundurthy, Katherine Crowson, Ludwig
  Schmidt, Robert Kaczmarczyk, and Jenia Jitsev.
\newblock {LAION}-{5B}: {An} open large-scale dataset for training next
  generation image-text models.
\newblock In \emph{Advances in {Neural} {Information} {Processing} {Systems}
  35: {Annual} {Conference} on {Neural} {Information} {Processing} {Systems}
  2022, {NeurIPS} 2022, {New} {Orleans}, {LA}, {USA}, {November} 28 -
  {December} 9, 2022}, 2022.

\bibitem[Sehner et~al.(2025)Sehner, Willems, Baumeyer, Davis, {van Schaik}, and
  Burkart]{sehner_sensitivity_2025}
Sandro Sehner, Erik~P. Willems, Adrian Baumeyer, Leyla Davis, Carel~P. {van
  Schaik}, and Judith~M. Burkart.
\newblock Sensitivity to immature skill deficits. {{Food}} sharing experiments
  in squirrel monkeys ({{Saimiri}} boliviensis) and common marmosets
  ({{Callithrix}} jacchus).
\newblock \emph{Journal of Comparative Psychology}, 139\penalty0 (3):\penalty0
  178--191, 2025.

\bibitem[Shahidi et~al.(2026)Shahidi, Ahmed, Badayeva, Lacal, and
  Gail]{shahidi_freely_2026}
Neda Shahidi, Zurna Ahmed, Yuliya Badayeva, Irene Lacal, and Alexander Gail.
\newblock Freely foraging macaques value information in ambiguous terrains.
\newblock \emph{Scientific Reports}, 16\penalty0 (1):\penalty0 881, 2026.

\bibitem[Singh et~al.(2025)Singh, Janson, Mehrbod, Ibrahim, Rish, Belilovsky,
  and Thérien]{singh_beyond_2025}
Vaibhav Singh, Paul Janson, Paria Mehrbod, Adam Ibrahim, Irina Rish, Eugene
  Belilovsky, and Benjamin Thérien.
\newblock Beyond {Cosine} {Decay}: {On} the effectiveness of {Infinite}
  {Learning} {Rate} {Schedule} for {Continual} {Pre}-training.
\newblock \emph{CoRR}, abs/2503.02844, 2025.
\newblock arXiv: 2503.02844.

\bibitem[Sun et~al.(2024)Sun, Zhou, Zhao, Yuan, Seybold, Hendon, Schroff, Ross,
  Adam, Hu, and {others}]{sun_video_2024}
Jennifer~J Sun, Hao Zhou, Long Zhao, Liangzhe Yuan, Bryan Seybold, David
  Hendon, Florian Schroff, David~A Ross, Hartwig Adam, Bo Hu, and {others}.
\newblock Video foundation models for animal behavior analysis.
\newblock \emph{bioRxiv}, pages 2024--07, 2024.
\newblock Publisher: Cold Spring Harbor Laboratory.

\bibitem[Tan et~al.(2020)Tan, Wang, Li, Li, Ouyang, Yin, and
  Yan]{tan_equalization_2020}
Jingru Tan, Changbao Wang, Buyu Li, Quanquan Li, Wanli Ouyang, Changqing Yin,
  and Junjie Yan.
\newblock Equalization {Loss} for {Long}-{Tailed} {Object} {Recognition}.
\newblock In \emph{2020 {IEEE}/{CVF} {Conference} on {Computer} {Vision} and
  {Pattern} {Recognition}, {CVPR} 2020, {Seattle}, {WA}, {USA}, {June} 13-19,
  2020}, pages 11659--11668. Computer Vision Foundation / IEEE, 2020.

\bibitem[Tenenbaum et~al.(2000)Tenenbaum, Silva, and
  Langford]{tenenbaum_global_2000}
Joshua~B. Tenenbaum, Vin~de Silva, and John~C. Langford.
\newblock A {Global} {Geometric} {Framework} for {Nonlinear} {Dimensionality}
  {Reduction}.
\newblock \emph{Science}, 290\penalty0 (5500):\penalty0 2319--2323, 2000.

\bibitem[Tong et~al.(2022)Tong, Song, Wang, and Wang]{tong_videomae_2022}
Zhan Tong, Yibing Song, Jue Wang, and Limin Wang.
\newblock {VideoMAE}: {Masked} {Autoencoders} are {Data}-{Efficient} {Learners}
  for {Self}-{Supervised} {Video} {Pre}-{Training}.
\newblock \emph{Advances in Neural Information Processing Systems},
  35:\penalty0 10078--10093, 2022.

\bibitem[Vincent et~al.(2008)Vincent, Larochelle, Bengio, and
  Manzagol]{vincent_extracting_2008}
Pascal Vincent, Hugo Larochelle, Yoshua Bengio, and Pierre-Antoine Manzagol.
\newblock Extracting and composing robust features with denoising autoencoders.
\newblock In \emph{Proceedings of the 25th international conference on
  {Machine} learning}, pages 1096--1103, 2008.

\bibitem[Wang et~al.(2022)Wang, Li, Li, He, Huang, Zhao, Zhang, Xu, Liu, Wang,
  Xing, Chen, Pan, Yu, Wang, Wang, and Qiao]{wang_internvideo_2022}
Yi Wang, Kunchang Li, Yizhuo Li, Yinan He, Bingkun Huang, Zhiyu Zhao, Hongjie
  Zhang, Jilan Xu, Yi Liu, Zun Wang, Sen Xing, Guo Chen, Junting Pan, Jiashuo
  Yu, Yali Wang, Limin Wang, and Yu Qiao.
\newblock {InternVideo}: {General} {Video} {Foundation} {Models} via
  {Generative} and {Discriminative} {Learning}.
\newblock \emph{CoRR}, abs/2212.03191, 2022.
\newblock arXiv: 2212.03191.

\bibitem[Wang et~al.(2025)Wang, Yu, Blau, Zhang, Laboratory, Paninski, Hurwitz,
  and Whiteway]{wang_self-supervised_2025}
Yanchen Wang, Han Yu, Ari Blau, Yizi Zhang, International~Brain Laboratory,
  Liam Paninski, Cole~L. Hurwitz, and Matthew~R. Whiteway.
\newblock Self-supervised pretraining of vision transformers for animal
  behavioral analysis and neural encoding.
\newblock \emph{CoRR}, abs/2507.09513, 2025.
\newblock arXiv: 2507.09513.

\bibitem[Xu et~al.(2021)Xu, Ghosh, Huang, Okhonko, Aghajanyan, Metze,
  Zettlemoyer, and Feichtenhofer]{xu_videoclip_2021}
Hu Xu, Gargi Ghosh, Po-Yao Huang, Dmytro Okhonko, Armen Aghajanyan, Florian
  Metze, Luke Zettlemoyer, and Christoph Feichtenhofer.
\newblock {VideoCLIP}: {Contrastive} {Pre}-training for {Zero}-shot
  {Video}-{Text} {Understanding}.
\newblock In \emph{Proceedings of the 2021 {Conference} on {Empirical}
  {Methods} in {Natural} {Language} {Processing}, {EMNLP} 2021, {Virtual}
  {Event} / {Punta} {Cana}, {Dominican} {Republic}, 7-11 {November}, 2021},
  pages 6787--6800. Association for Computational Linguistics, 2021.

\bibitem[Yang et~al.(2022)Yang, Yang, Xu, Zhang, Lan, and
  Tao]{yang_apt-36k_2022}
Yuxiang Yang, Junjie Yang, Yufei Xu, Jing Zhang, Long Lan, and Dacheng Tao.
\newblock {APT}-{36K}: {A} {Large}-scale {Benchmark} for {Animal} {Pose}
  {Estimation} and {Tracking}.
\newblock In \emph{Advances in {Neural} {Information} {Processing} {Systems}
  35: {Annual} {Conference} on {Neural} {Information} {Processing} {Systems}
  2022, {NeurIPS} 2022, {New} {Orleans}, {LA}, {USA}, {November} 28 -
  {December} 9, 2022}, 2022.

\bibitem[Zhao et~al.(2025)Zhao, Wu, and Wu]{zhao_web-scale_2025}
Brian~Nlong Zhao, Jiajun Wu, and Shangzhe Wu.
\newblock Web-{Scale} {Collection} of {Video} {Data} for {4D} {Animal}
  {Reconstruction}.
\newblock \emph{arXiv preprint arXiv:2511.01169}, 2025.

\bibitem[Zhao et~al.(2024)Zhao, Gundavarapu, Yuan, Zhou, Yan, Sun, Friedman,
  Qian, Weyand, Zhao, Hornung, Schroff, Yang, Ross, Wang, Adam, Sirotenko, Liu,
  and Gong]{zhao_videoprism_2024}
Long Zhao, Nitesh~Bharadwaj Gundavarapu, Liangzhe Yuan, Hao Zhou, Shen Yan,
  Jennifer~J. Sun, Luke Friedman, Rui Qian, Tobias Weyand, Yue Zhao, Rachel
  Hornung, Florian Schroff, Ming-Hsuan Yang, David~A. Ross, Huisheng Wang,
  Hartwig Adam, Mikhail Sirotenko, Ting Liu, and Boqing Gong.
\newblock {VideoPrism}: {A} {Foundational} {Visual} {Encoder} for {Video}
  {Understanding}.
\newblock In \emph{Forty-first {International} {Conference} on {Machine}
  {Learning}, {ICML} 2024, {Vienna}, {Austria}, {July} 21-27, 2024}.
  OpenReview.net, 2024.

\bibitem[Zhou et~al.(2022)Zhou, Wei, Wang, Shen, Xie, Yuille, and
  Kong]{zhou_image_2022}
Jinghao Zhou, Chen Wei, Huiyu Wang, Wei Shen, Cihang Xie, Alan~L. Yuille, and
  Tao Kong.
\newblock Image {BERT} {Pre}-training with {Online} {Tokenizer}.
\newblock In \emph{The {Tenth} {International} {Conference} on {Learning}
  {Representations}, {ICLR} 2022, {Virtual} {Event}, {April} 25-29, 2022}.
  OpenReview.net, 2022.

\end{thebibliography}
}

\onecolumn
\appendix
\clearpage

\setcounter{page}{1}
\maketitlesupplementary

\begin{table*}[t]
    \nicetable
    \footnotesize
    \caption{\textbf{Detailed composition of our dataset.} 
    The proportion of the different sub-datasets corresponds to the number of 3-s snippets in each sub-dataset. Note that we vary the sampling stride between datasets and sample overlapping snippets for small datasets. We also report the duration of unique video material of each dataset. We publish all data except for the chimpanzee subset.  } %
    \begin{tabular}{
    l
    r
    rr
    >{\raggedright\arraybackslash}p{3cm}
    >{\raggedright\arraybackslash}p{2cm}
    >{\raggedright\arraybackslash}p{2.5cm}
    r}
    \toprule
         \textbf{Name} &     \textbf{Weight [\%]}& \textbf{\#Snippets}&\textbf{Unique Hours} &\textbf{Location}& \textbf{Setting}& \textbf{Species}&\textbf{Raw [h]} \\
         \midrule
         \textbf{YT-Filtered} & \textbf{63.0}&\textbf{454k}& \textbf{250.0}&diverse&diverse&diverse& \textbf{458} \\
 \midrule
         \textbf{\rando{}} & \textbf{37.0}& \textbf{267k}& \textbf{173.7}&&&& \textbf{721} \\

eulemur \hfill \cite{karakoc_foraging_2025}& 
       3.9& 28k&23.2&\blind{Kirindy Forest}, Madagascar & Wild, Experiment, Multi-Cam& Red-fronted lemurs (\textit{Eulemur rufifrons})&311\\ %

        baboon\_w \hfill \cite{ohearn_increased_2025, ohearn_lessons_2024}& 4.4&  31k&26.1& \blind{CRP Simenti}, Senegal& Wild, Experiment& Guinea baboons (\textit{Papio papio})&80\\ %

    baboon\_a & 1.1&  8k&2.2& \blind{CRP Simenti}, Senegal& Wild& Guinea baboons (\textit{Papio papio})&2\\ %

    lemur & 4.5&  32k&26.9& \blind{Affenwald Straußberg}, Germany& Semi-free ranging& Ring-tailed lemurs (\textit{Lemur catta})&38\\ %

 assamese \hfill \cite{pereira_complex_2025}& 0.1&  1k&0.3& \blind{Phu Khieo Wildlife Sanctuary}, Thailand& Wild& Assamese macaques (\textit{Macaca assamensis})&8\\ %

    barbary\_a & 2.2&  16k&9.0& \blind{La Forêt des Singes, Rocamadour}, France& Semi-free ranging& Barbary macaques (\textit{Macaca sylvanus})&21\\ %

    saimiri \hfill \cite{sehner_sensitivity_2025}& 2.0&  15k&12& \blind{Basel and Zurich Zoo}, Switzerland& Captive& Black-capped squirrel monkeys (\textit{Saimiri boliviensis})&30\\ %
    barbary\_t \hfill \cite{bosshard_ecological_2025} & 1.6&  12k&6.3& \blind{La Forêt des Singes, Rocamadour}, France& Semi-free ranging, Experiment& Barbary macaques (\textit{Macaca sylvanus})&6\\ %
    marmoset \hfill \cite{sehner_sensitivity_2025}& 2.1&  15k&12.5& \blind{University of Zurich}, Switzerland& Captive, Multi-Cam& Common marmosets (\textit{Callithrix jacchus})&100\\ %
    rhesus \hfill \cite{shahidi_freely_2026}& 1.3&  9k&7.7& \blind{German Primate Center}, Germany & Captive, Multi-Cam& Rhesus macaques (\textit{Macaca mulatta})&62\\ %

 chimpanzee (private) & 13.7&  99k&47.3& \blind{Moyen-Bafing National Park}, Guinea& Wild & 96.5\,\% Chimpanzees (\textit{Pan troglodytes}), 3.5\,\% Guinea baboons (\textit{Papio papio}) &63\\ %
 \midrule
 
 \textbf{\prmt{}}& \textbf{100.0} & \textbf{721k} & \textbf{423.7} & & & & \textbf{1179} \\
 \bottomrule
 \end{tabular}
    
    \label{tab:supp_data_subsets}
\end{table*}

\section{\prmt{} Dataset}
\label{sec:supp_dataset}

See \cref{tab:supp_data_subsets} for details about all subsets of \prmt{} and \cref{fig:supp_privi_examples} for example frames from the video snippets. 

In the following paragraphs, we will give a brief description of each of the source datasets in \rando{}.

\paragraph{eulemur} Videos are collected from a social learning experiment with red-fronted lemurs which involved feeding boxes with different opening techniques. Lemurs wear collars for identification. Observed behaviors are various types of box interactions, scrounging, and looking at other lemurs. The experiments were filmed with fixed cameras from various camera angles (close-up, bird's eye view, and tree-mounted).

\paragraph{baboon\_w} This dataset contains videos of a foraging experiment with wild Guinea baboons (\textit{Papio papio}) in \blind{Niokolo Koba National Park} in Senegal. The experiment consists of baboons learning to operate a food box that rewards them with peanuts for pulling a lever on its side. Once a sufficient number of baboons were trained to operate the lever one was chosen to be the sole operator for the remainder of the experiment and his social interactions were monitored. Video cameras were placed on fixed tripods a set distance from the box (around 8\,m) as part of the set up for both training and experimental sessions with the box and remained on until the box was empty of food. During sessions the primary behavior of the baboons is feeding, and resting, with occasional bouts of aggression and allogrooming. 

\paragraph{baboon\_a} Miscellaneous, diverse footage of baboons at \blind{Niokolo Koba National Park} in Senegal. Recordings are not from a single experiment, but focus on social behaviors like grooming and sitting together. Recordings contain some fights. 

\paragraph{lemur} This dataset was collected during short focal follows of semi–free-ranging ring-tailed lemurs (\textit{Lemur catta}). Video recordings were obtained using hand-held cameras, and subjects wore tracking collars for identification and monitoring.

\paragraph{assamese} Miscellaneous footage of Assamese macaques (\textit{Macaca assamensis}) expressing natural behavior in the wild at Phu Khieo Wildlife Sanctuary, Thailand. Video data was collected only when visibility was good, i.e. often when the individuals were on the ground or at low canopy levels. Common behaviors include feeding, resting and social interactions. The camera was either hand-held or head-mounted.

\paragraph{barbary\_a} Miscellaneous footage of semi-free ranging Barbary macaques (\textit{Macaca sylvanus}) at \blind{La Forêt des Singes, Rocamadour} in France. Most videos are of macaques walking through a tunnel made of wire mesh. These recordings were done to capture standardized videos of individuals in the population for individual identification and for determining group structure. The dataset also contains recordings not containing the wire mesh tunnel, most of which are of social interactions.

\paragraph{saimiri}
Squirrel monkeys were tested at \blind{Zoo Basel} and \blind{Zoo Zurich}. At \blind{Basel}, they lived in three connected indoor and two outdoor enclosures enriched with climbing structures. At \blind{Zurich}, they were  housed in two connected indoor and two outdoor enclosures, with indoor vegetation and large outdoor trees. Both groups were tested in the morning after the first feeding and filmed during an extractive foraging task in one of their home enclosures, with all group members able to access the apparatus mounted on a wooden table in the enclosure center.

\paragraph{barbary\_t} This is footage of a risk assessment experiment with barbary macaques. A wooden box contains a rubber snake (high risk) or a cube with snake-like visual appearance (low risk) and a peanut (high reward) or a popcorn (low reward). It was studied whether macaques take the food or not. The camera is fixed in the wooden box, recording outwards.

\paragraph{marmoset} 
Marmosets were housed in heated indoor enclosures with ad libitum access to outdoor spaces. Both areas were enriched with natural branches, climbing structures, wood chips, or soil with vegetation.  Animals were filmed during an extractive foraging task in their home enclosure, with all group members able to access the apparatus mounted on a wooden table placed in the enclosure center.

\paragraph{rhesus} Recordings are of a foraging experiment with one or two rhesus macaques in a white experiment room. The floor in the experiment room is covered with piles of wood chips, some of which contain varying amounts of food. It was observed whether macaques learn in which piles they can expect food.  Recordings were captured with several fixed cameras on walls and ceiling.

\paragraph{chimpanzee} This dataset was collected by the Moyen Bafing Chimpanzee Project and consists of camera‑trap footage of chimpanzees in a savanna mosaic in Guinea. Bushnell 4K no‑glow cameras were deployed to document a wide range of behaviors, including tool-use. This dataset cannot be published yet as it is part of an ongoing research project.

\begin{figure*}[htbp]
    \centering
    \begin{subfigure}{\textwidth}
        \centering
        \includegraphics[width=0.32\textwidth]{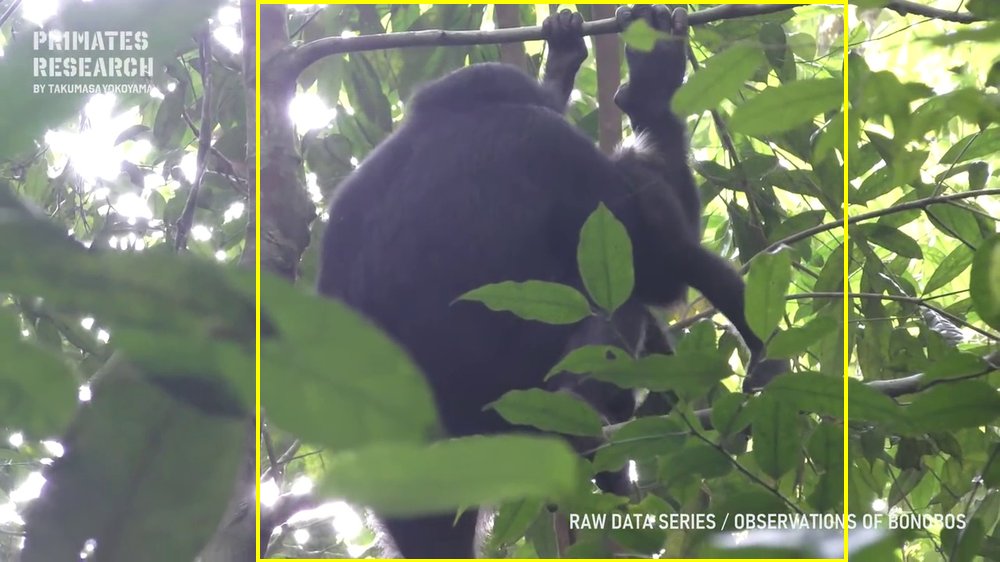} \hfill \includegraphics[width=0.32\textwidth]{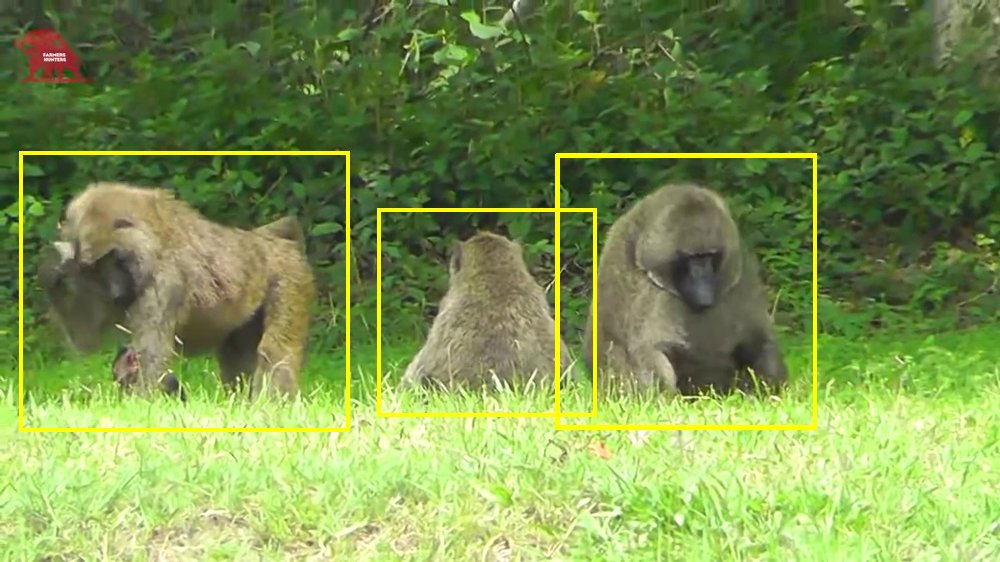} \hfill \includegraphics[width=0.32\textwidth]{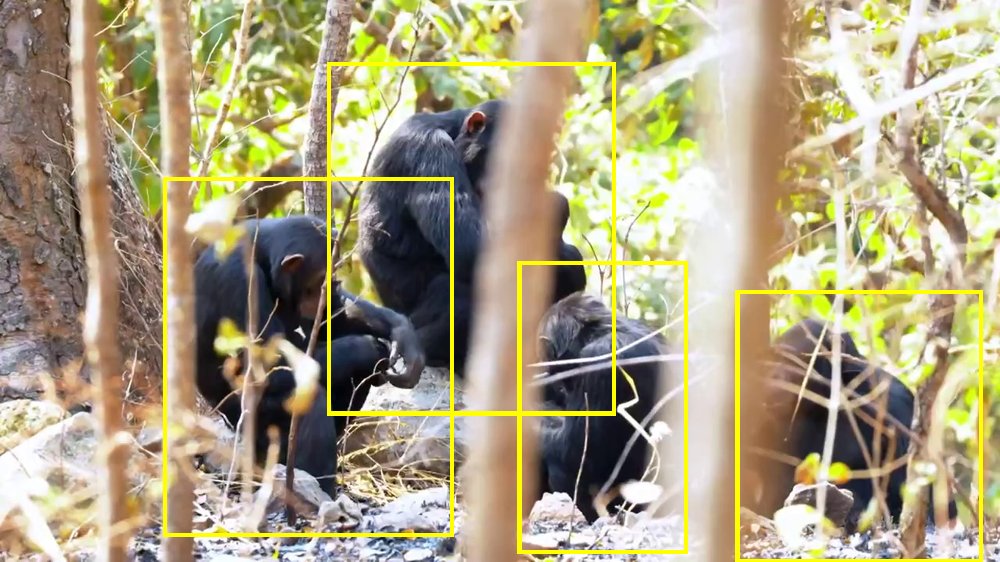}
    \end{subfigure}
    \vspace{1em}
    \begin{subfigure}{\textwidth}
        \centering
        \includegraphics[width=0.32\textwidth]{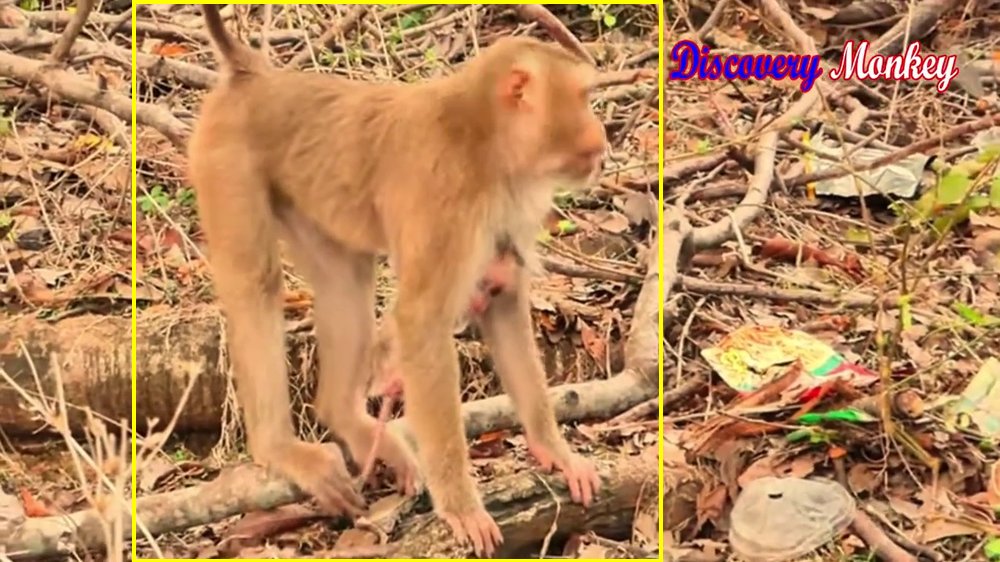} \hfill \includegraphics[width=0.32\textwidth]{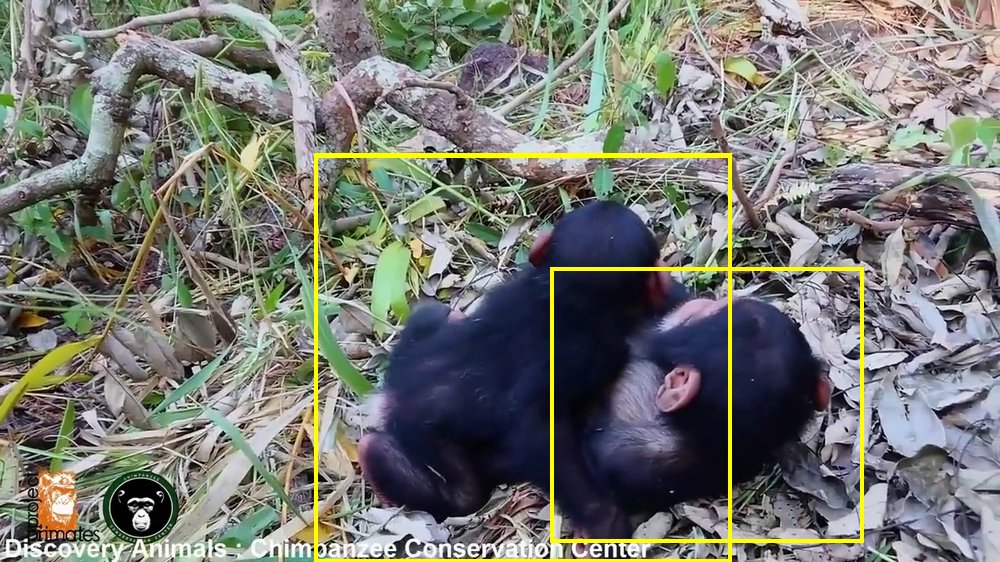} \hfill \includegraphics[width=0.32\textwidth]{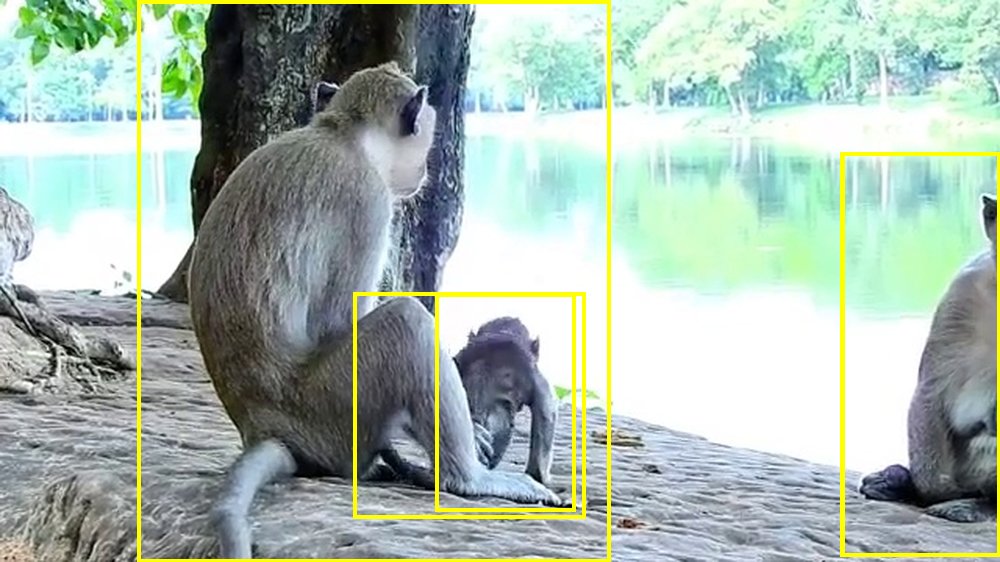}
    \end{subfigure}
    \vspace{1em}
    \begin{subfigure}{\textwidth}
        \centering
        \includegraphics[width=0.32\textwidth]{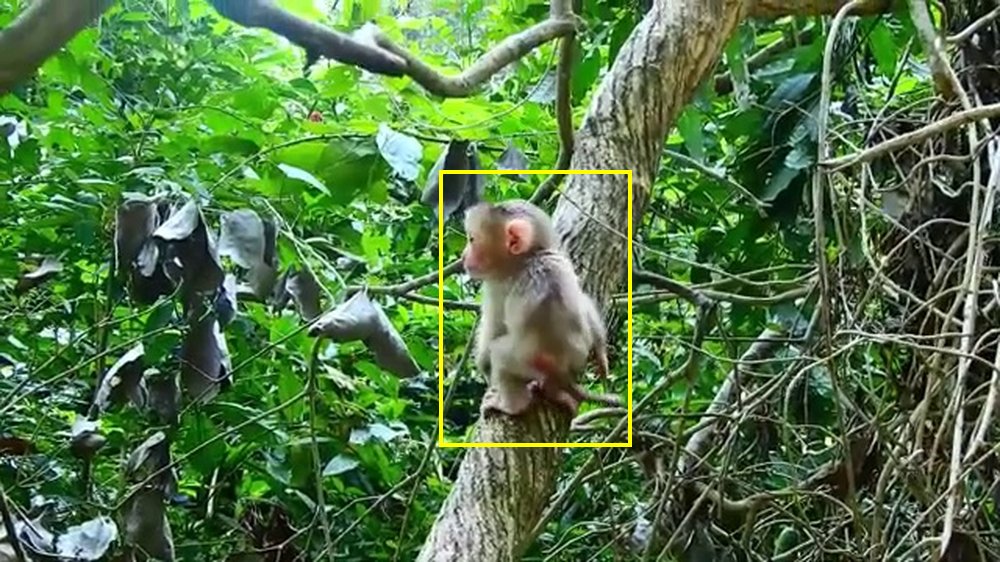} \hfill \includegraphics[width=0.32\textwidth]{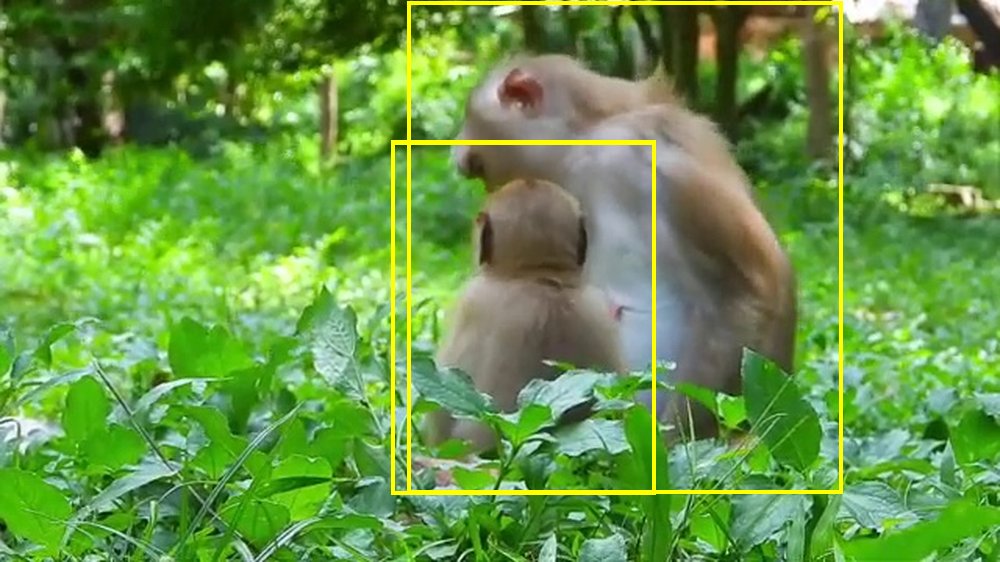} \hfill \includegraphics[width=0.32\textwidth]{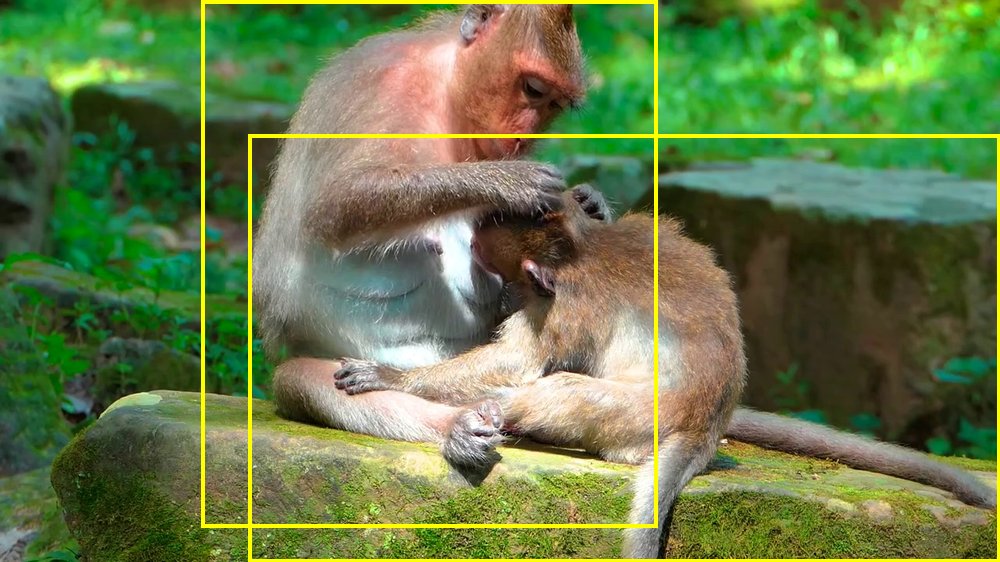}
        \caption{YT-Filtered}
    \end{subfigure}
    \vspace{1em}
    \begin{subfigure}{\textwidth}
        \centering
        \includegraphics[width=0.32\textwidth]{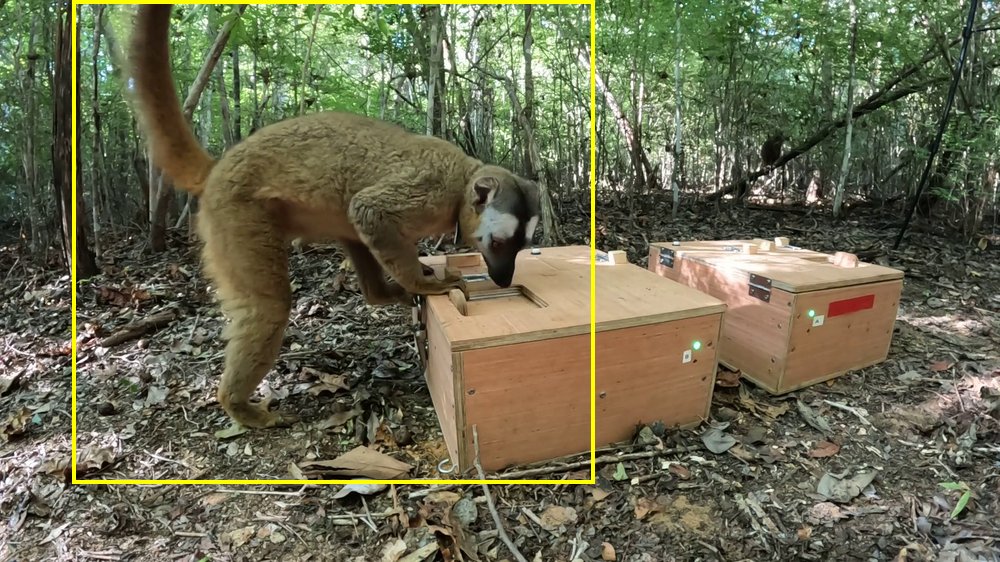} \hfill \includegraphics[width=0.32\textwidth]{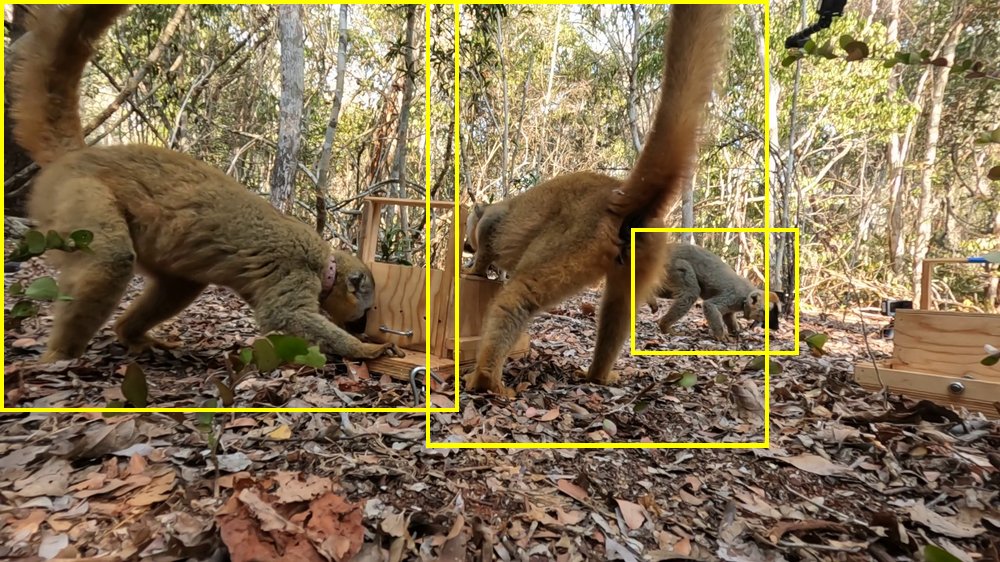} \hfill \includegraphics[width=0.32\textwidth]{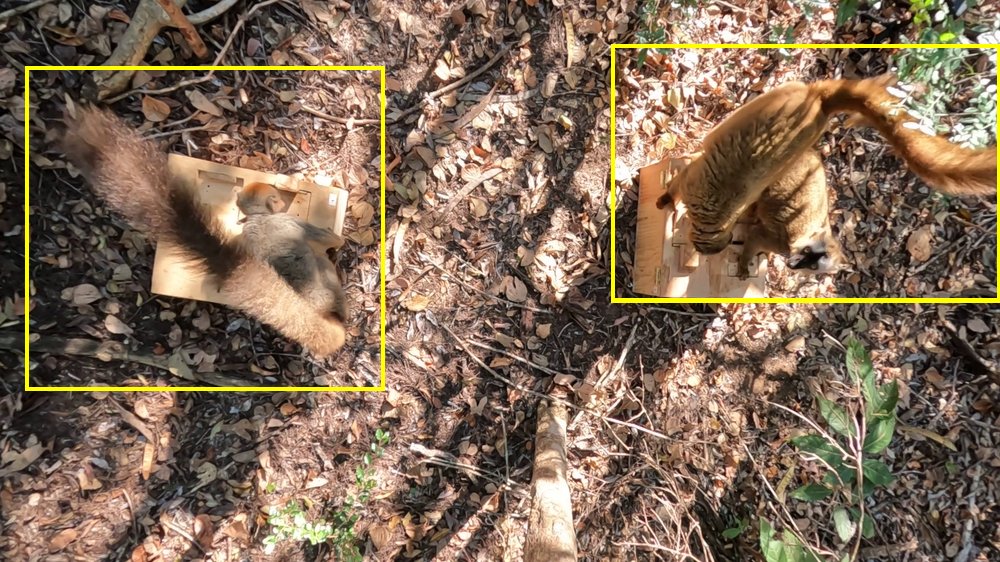}
        \caption{eulemur}
    \end{subfigure}
    \vspace{1em}
    \begin{subfigure}{\textwidth}
        \centering
        \includegraphics[width=0.32\textwidth]{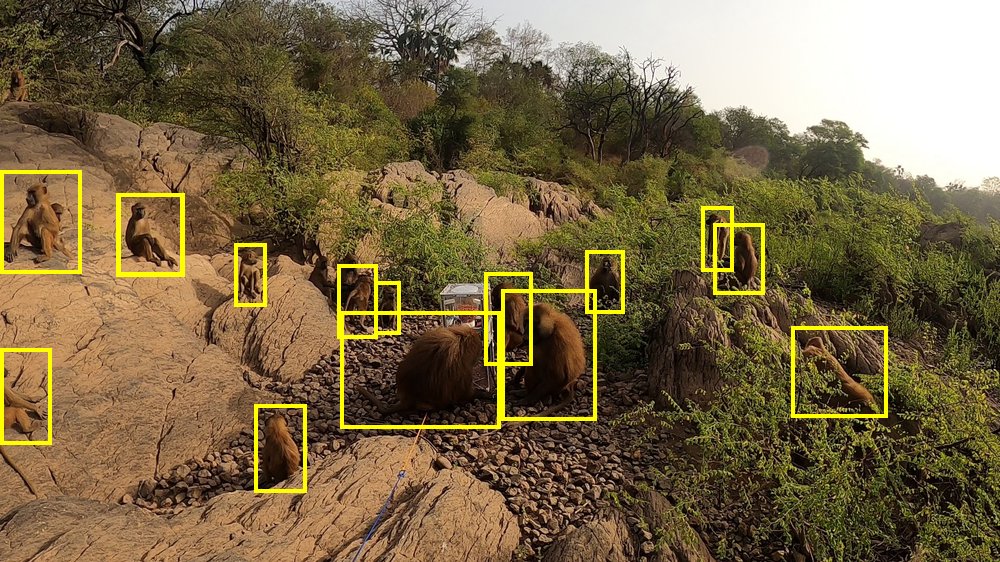} \hfill \includegraphics[width=0.32\textwidth]{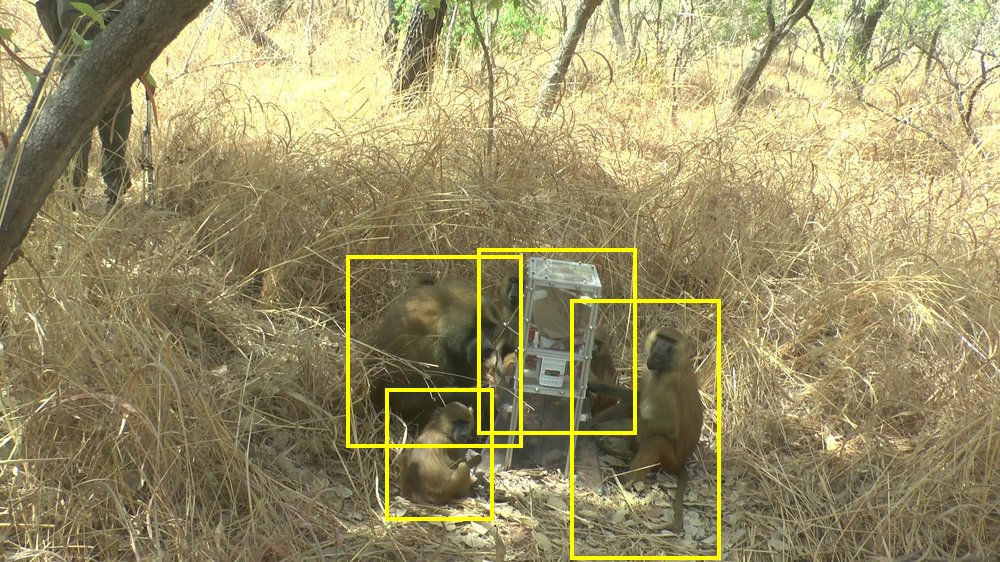} \hfill \includegraphics[width=0.32\textwidth]{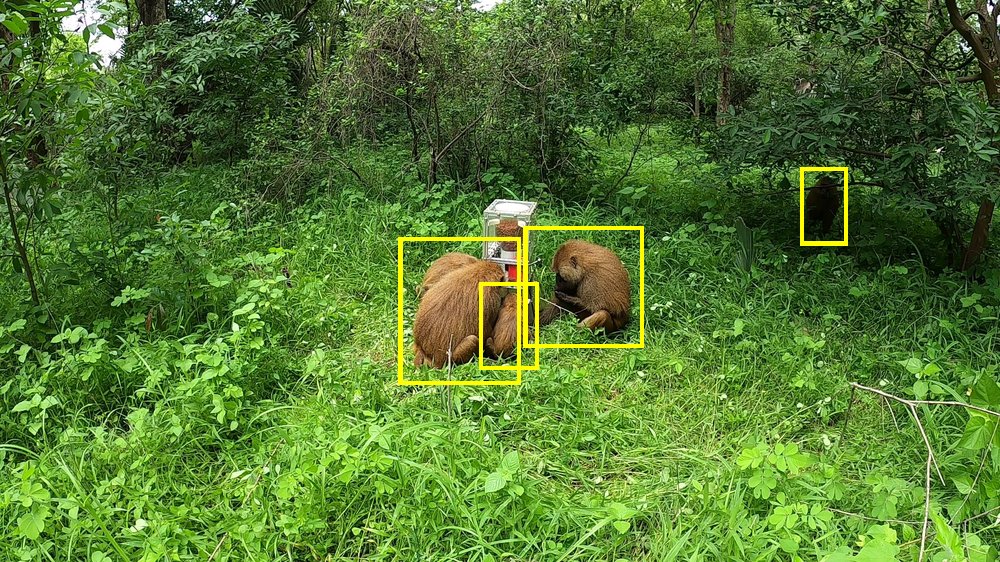}
        \caption{baboon\_w}
    \end{subfigure}
    \vspace{1em}

    \caption{\textbf{Examples of \prmt{} dataset samples.} Examples sorted by sub-dataset. We only show center frames of the video snippets, please see the supplementary material for videos. Bounding boxes of detected primates in yellow. (continued on next page)}
\end{figure*}
\clearpage

\begin{figure*}[htbp]
    \ContinuedFloat
    \centering
    \begin{subfigure}{\textwidth}
        \centering
        \includegraphics[width=0.32\textwidth]{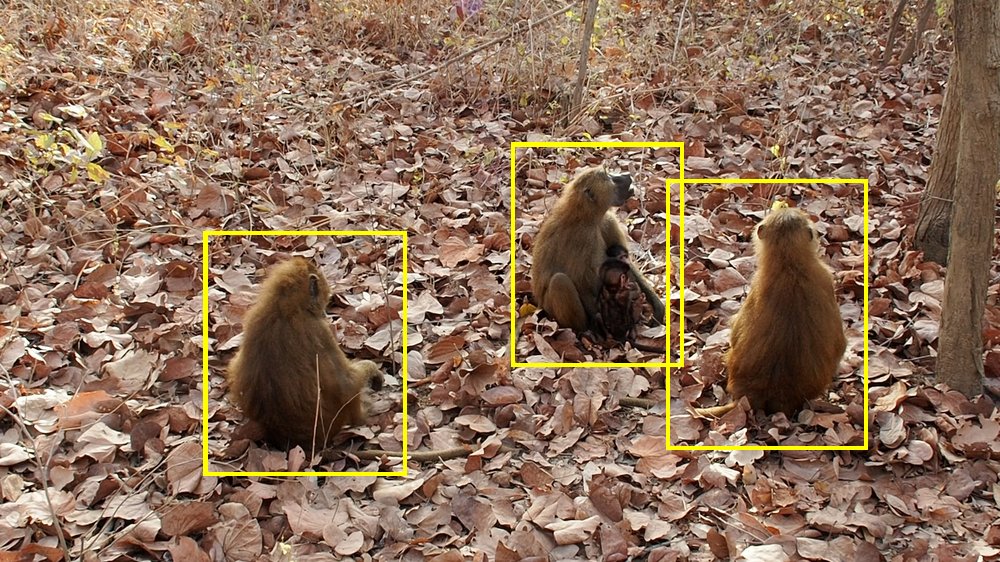} \hfill \includegraphics[width=0.32\textwidth]{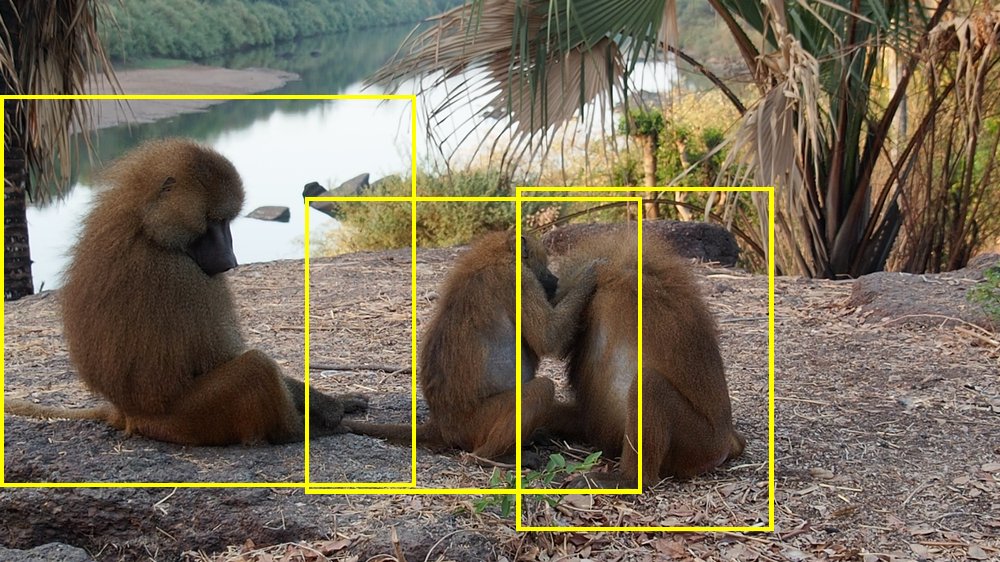} \hfill \includegraphics[width=0.32\textwidth]{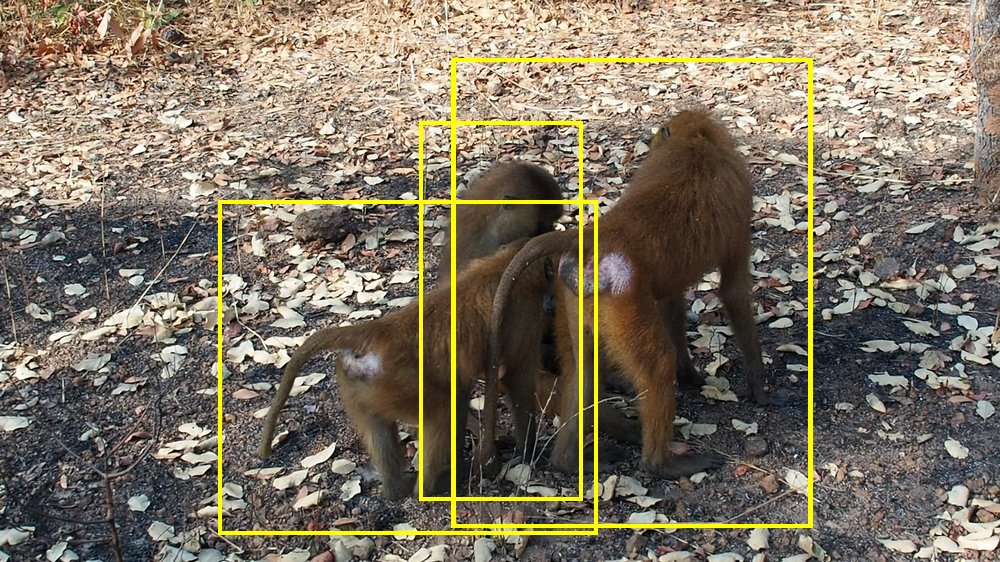}
        \caption{baboon\_a}
    \end{subfigure}
    \vspace{1em}
    \begin{subfigure}{\textwidth}
        \centering
        \includegraphics[width=0.32\textwidth]{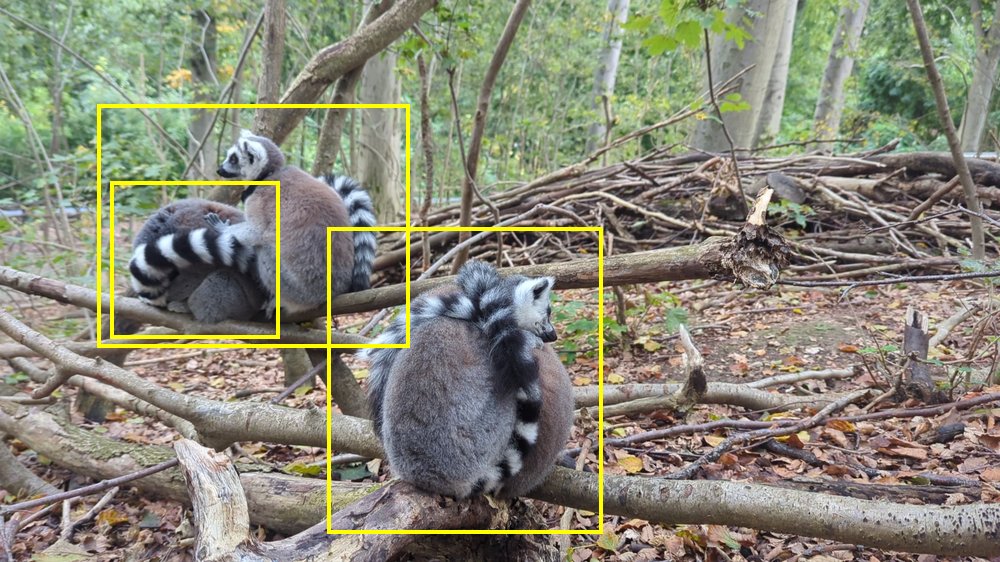} \hfill \includegraphics[width=0.32\textwidth]{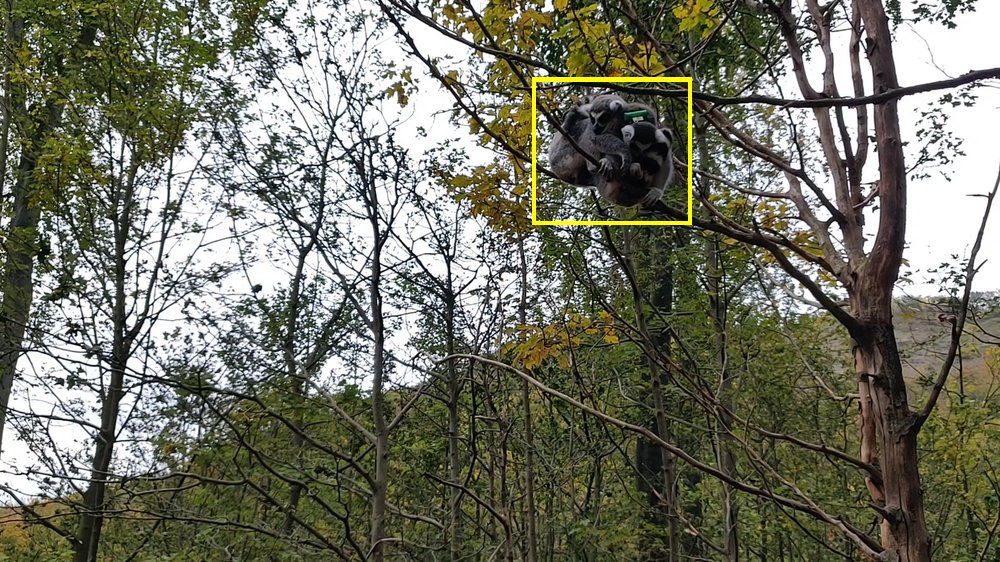} \hfill \includegraphics[width=0.32\textwidth]{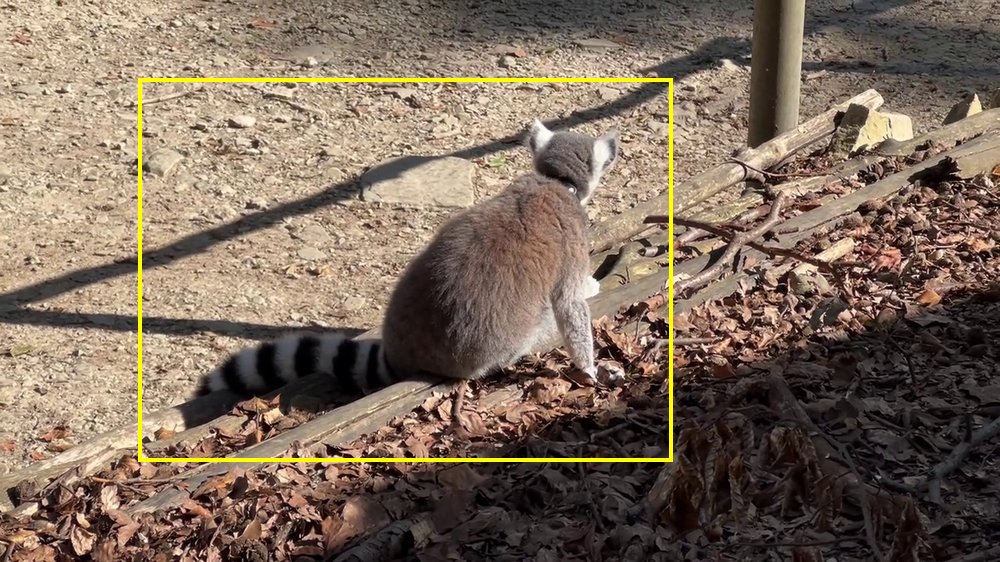}
        \caption{lemur}
    \end{subfigure}
    \vspace{1em}
    \begin{subfigure}{\textwidth}
        \centering
        \includegraphics[width=0.32\textwidth]{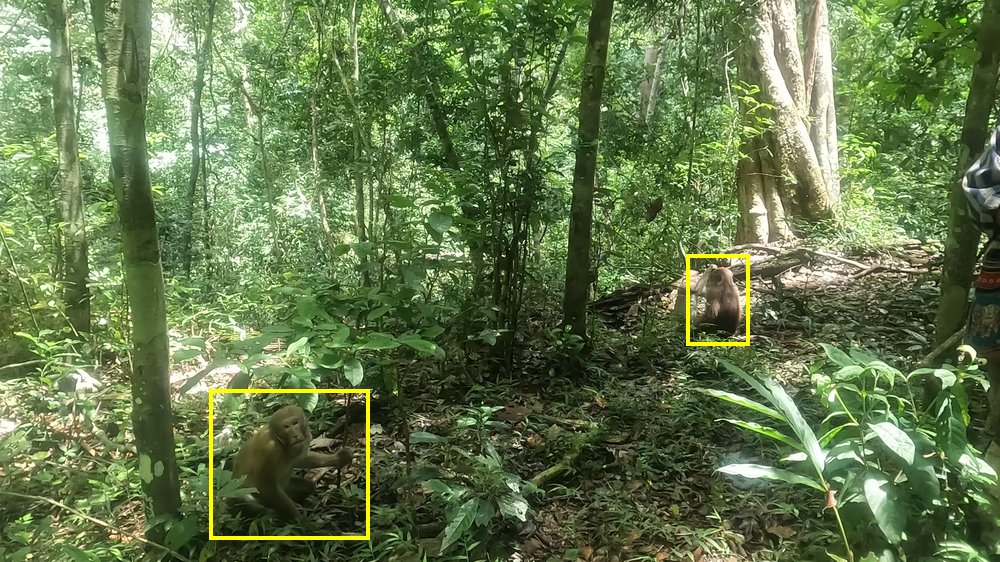} \hfill \includegraphics[width=0.32\textwidth]{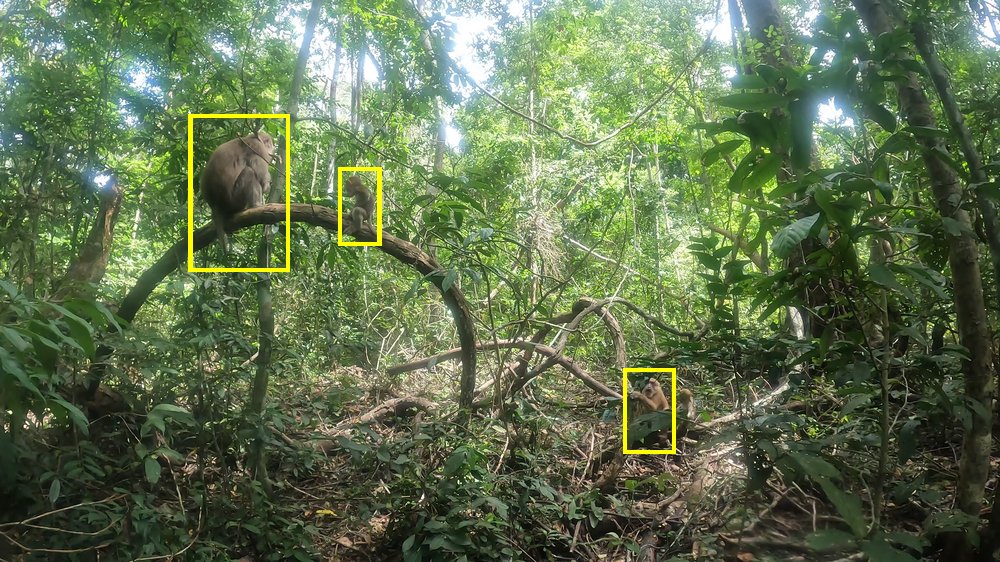} \hfill \includegraphics[width=0.32\textwidth]{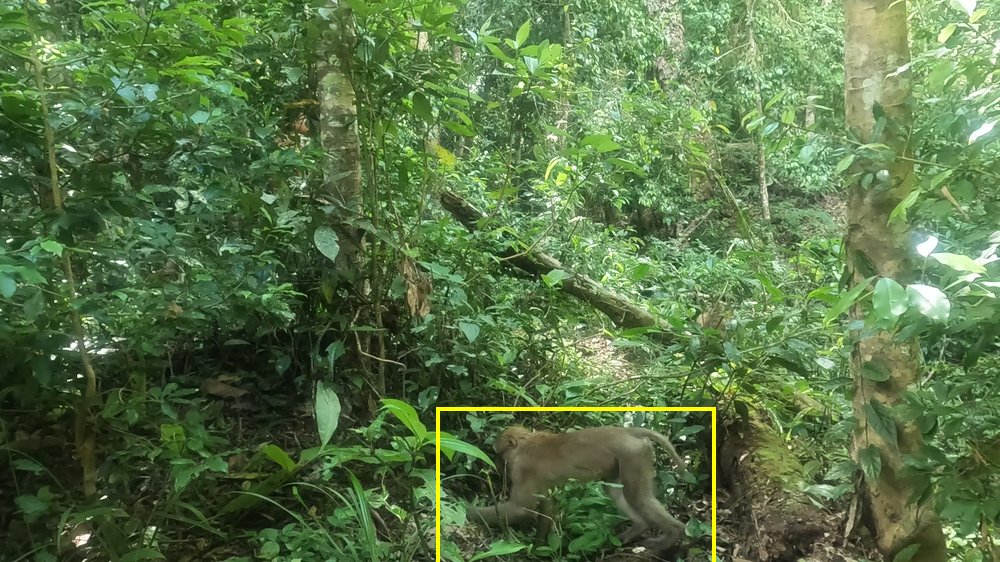}
        \caption{assamese}
    \end{subfigure}
    \vspace{1em}
    \begin{subfigure}{\textwidth}
        \centering
        \includegraphics[width=0.32\textwidth]{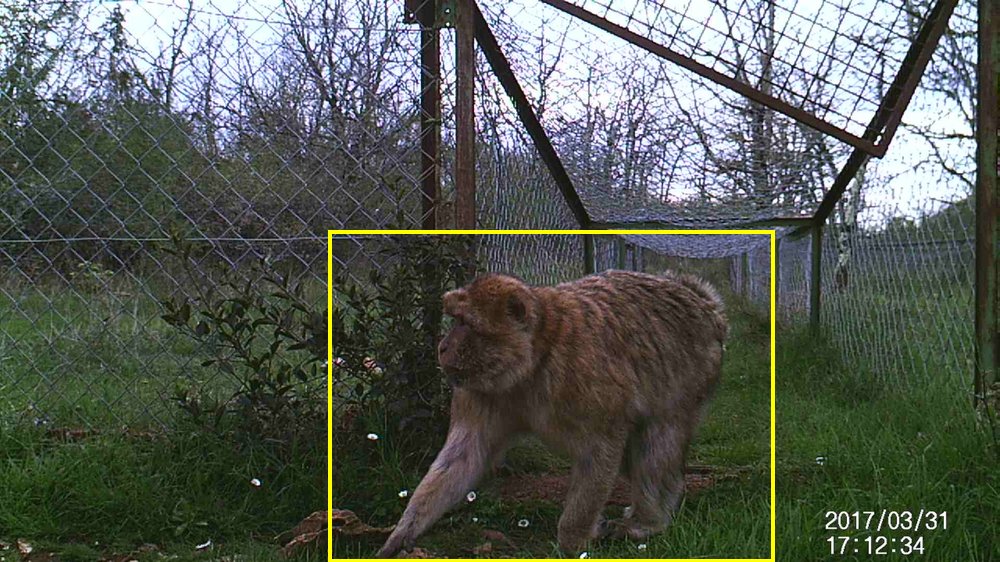} \hfill \includegraphics[width=0.32\textwidth]{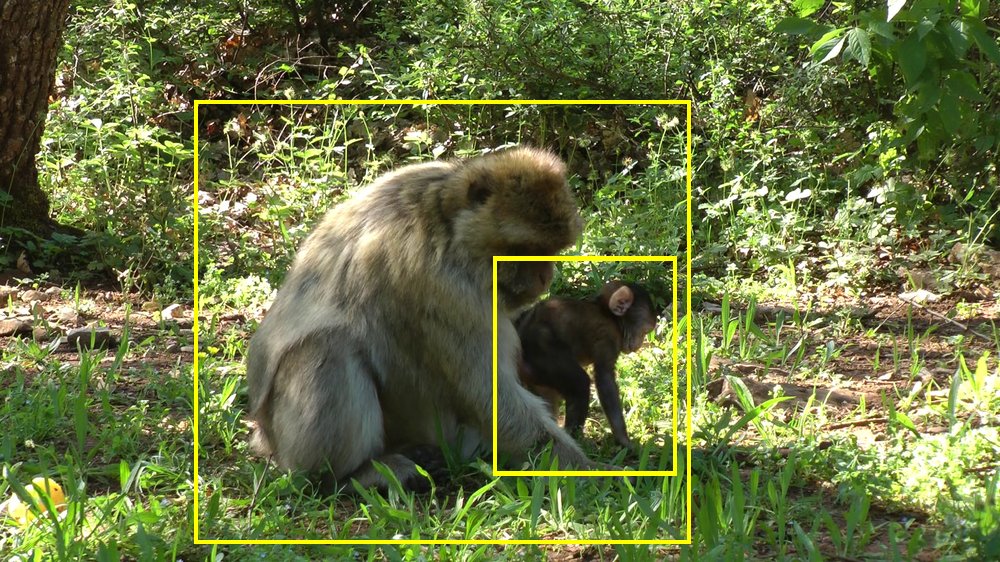} \hfill \includegraphics[width=0.32\textwidth]{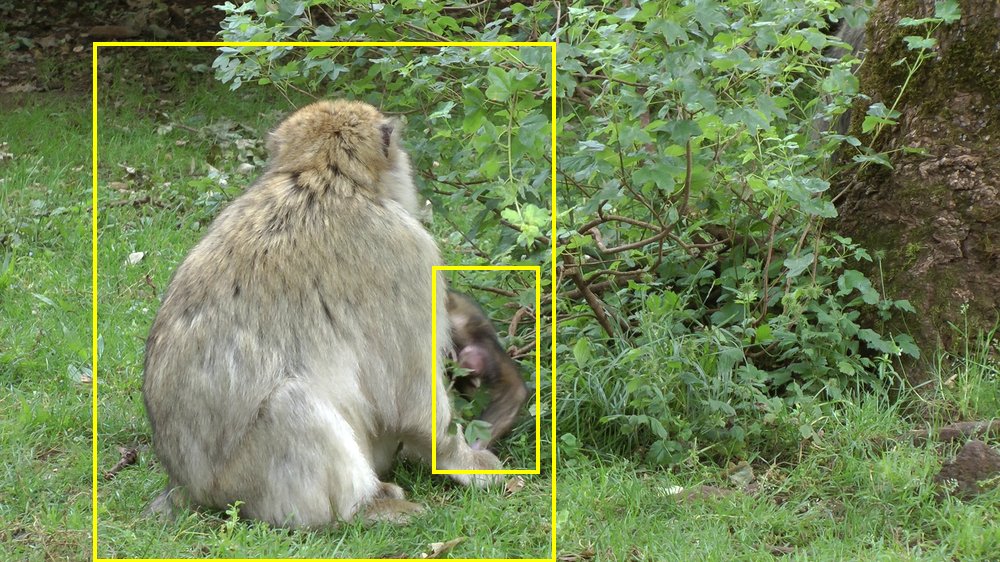}
        \caption{barbary\_a}
    \end{subfigure}
    \vspace{1em}
    
    \begin{subfigure}{\textwidth}
        \centering
        \includegraphics[width=0.32\textwidth]{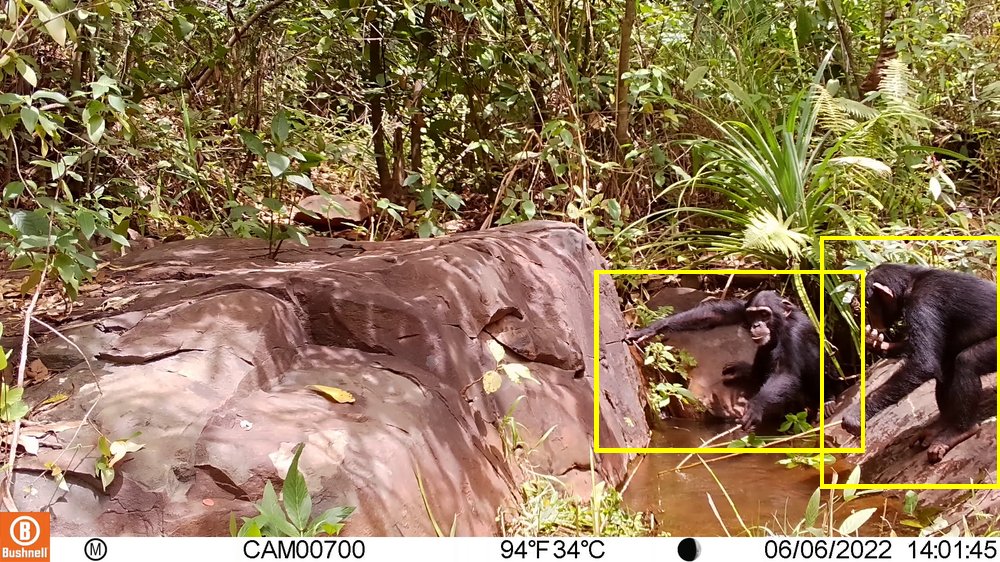} \hfill \includegraphics[width=0.32\textwidth]{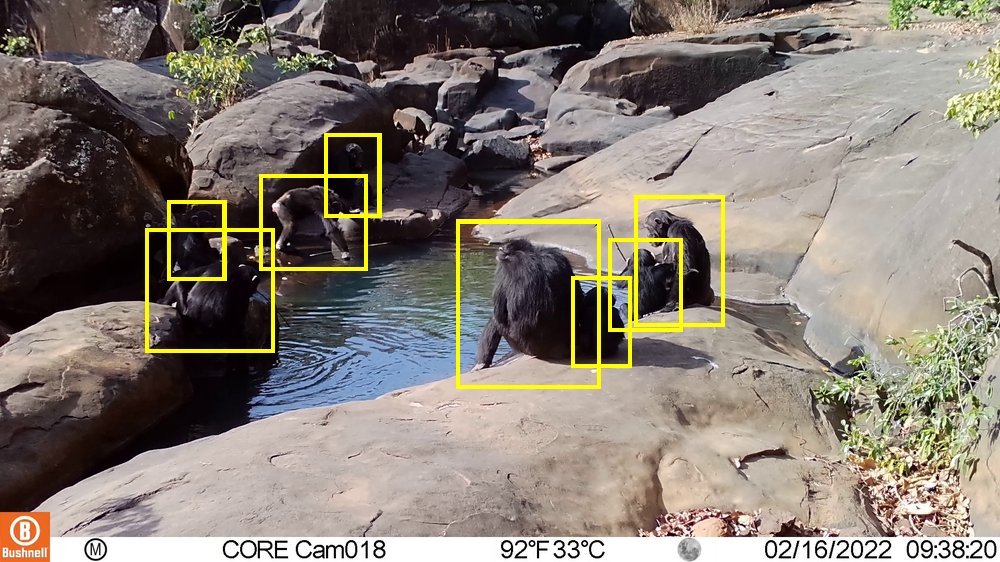} \hfill \includegraphics[width=0.32\textwidth]{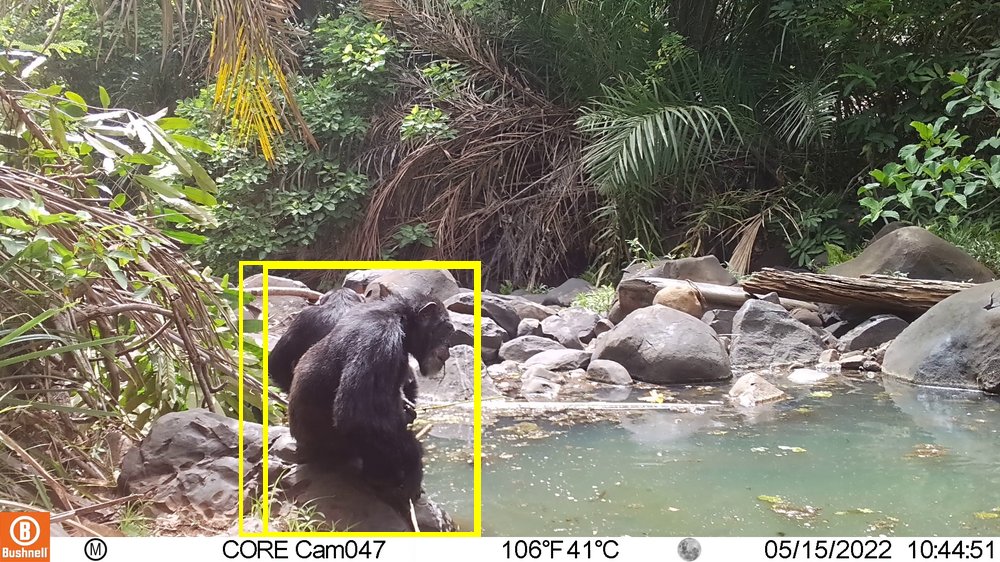}
        \caption{chimpanzee}
    \end{subfigure}
    
    \vspace{1em}
    
    \caption{\textbf{Examples of \prmt{} dataset samples.} Examples sorted by sub-dataset. We only show center frames of the video snippets, please see the supplementary material for videos. Bounding boxes of detected primates in yellow. (continued on next page)}
\end{figure*}
\clearpage

\begin{figure*}[htbp]
    \ContinuedFloat
    \centering
    \begin{subfigure}{\textwidth}
        \centering
        \includegraphics[width=0.32\textwidth]{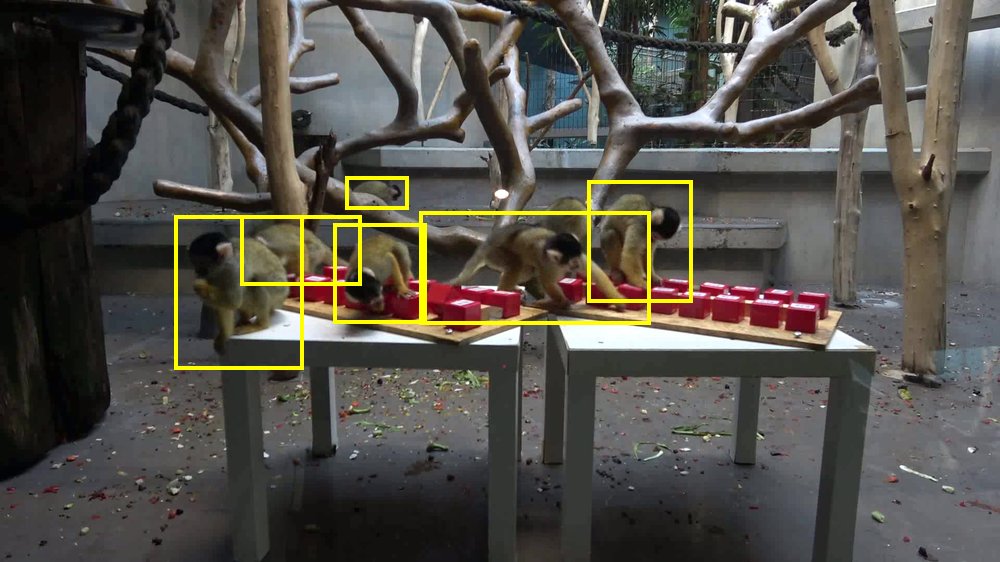} \hfill \includegraphics[width=0.32\textwidth]{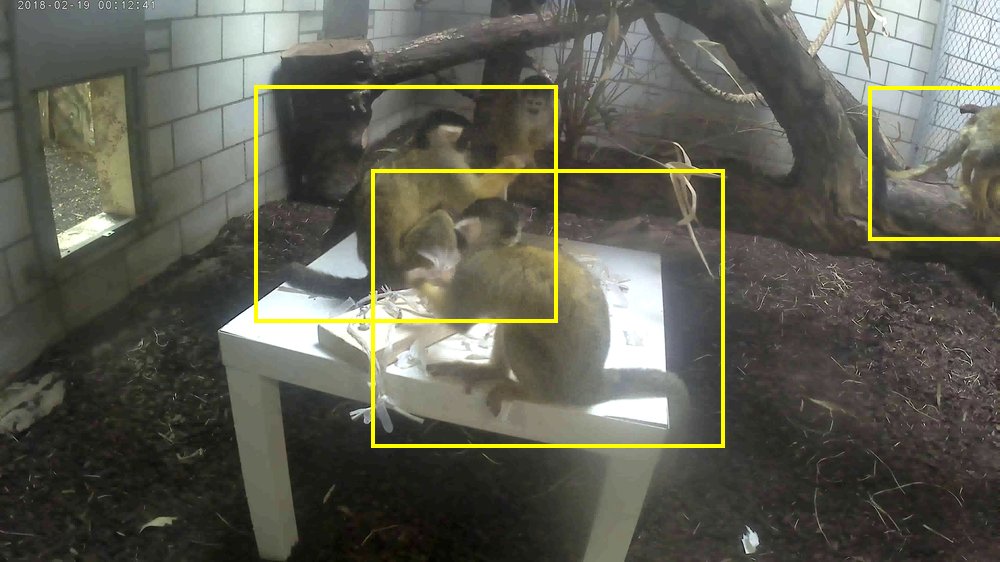} \hfill \includegraphics[width=0.32\textwidth]{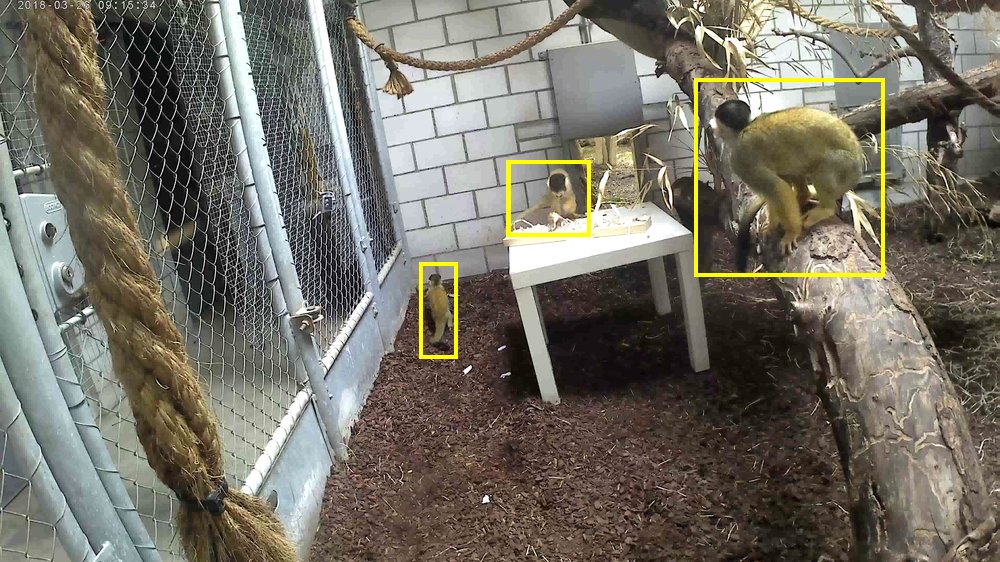}
        \caption{saimiri}
    \end{subfigure}
    \vspace{1em}
    \begin{subfigure}{0.32\textwidth}
        \centering
        \includegraphics[width=\textwidth]{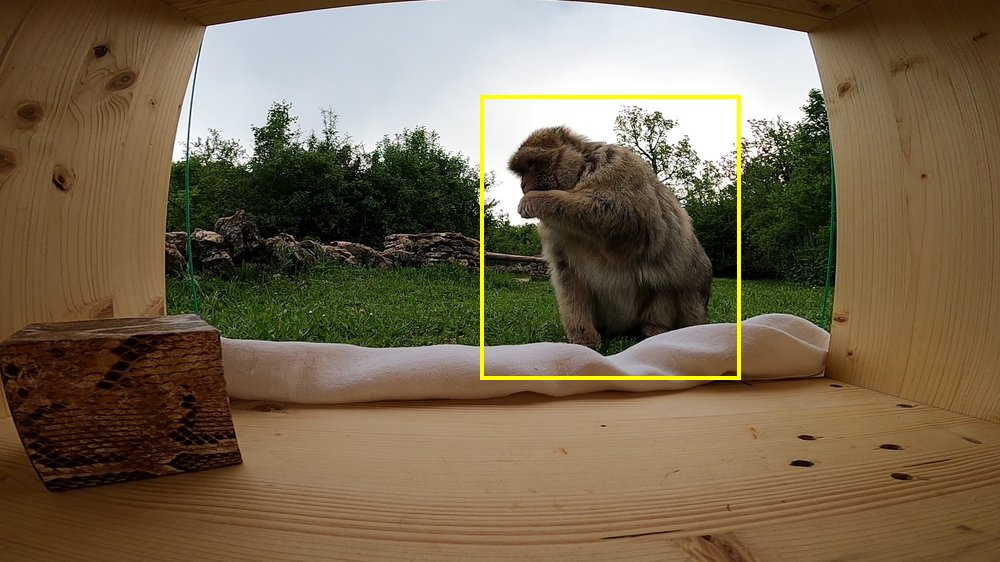} 
        \caption{barbary\_t}
    \end{subfigure}
    \hfill
    \begin{subfigure}{0.32\textwidth}
        \centering
        \includegraphics[width=\textwidth]{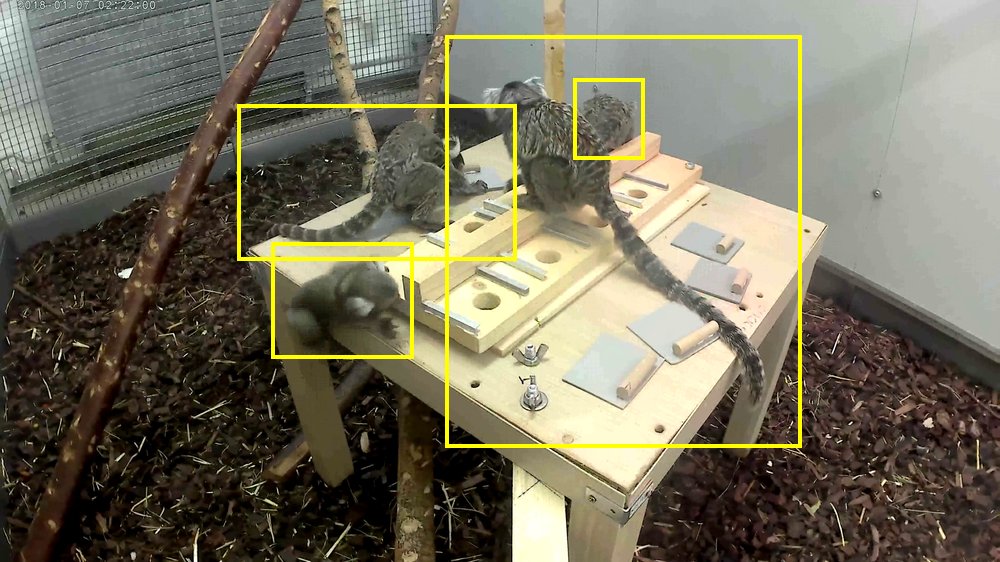} 
        \caption{marmoset}
    \end{subfigure}
    \hfill
    \begin{subfigure}{0.32\textwidth}
        \centering
        \includegraphics[width=\textwidth]{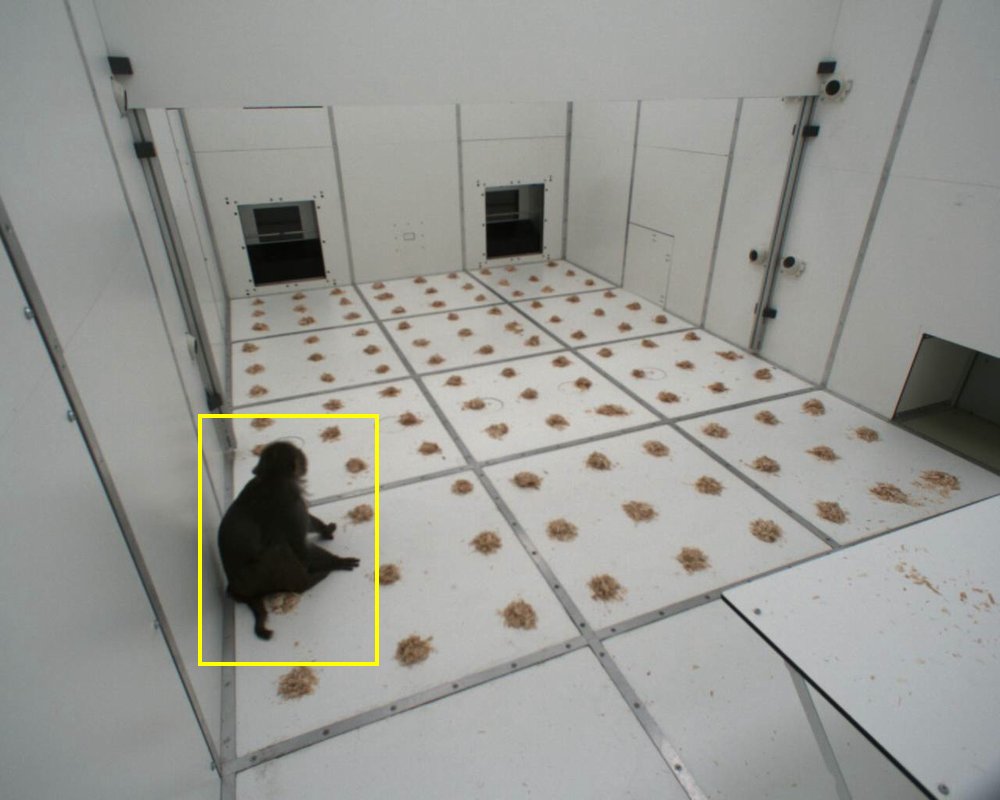}
        \caption{rhesus}
    \end{subfigure}
    \vspace{1em}
    
    \caption{\textbf{Examples of \prmt{} dataset samples.} Examples sorted by sub-dataset. We only show center frames of the video snippets, please see the supplementary material for videos. Bounding boxes of detected primates in yellow.}
    \label{fig:supp_privi_examples}
\end{figure*}

\clearpage

\section{Data Pipeline}
\label{sec:supp_data_pipeline}

\paragraph{Relevance Labeling} Labeling was performed on randomly selected center frames from YouTube snippets after cut detection and discarding videos shorter than 3 seconds. 

Labeling criteria: \emph{
    Label a video as relevant if the video visibly contains a primate (drawings, dolls, reflections, or computer generated content do not count) or if it is a recording of the natural habitat of primates (e.g. forest, savannah) and might contain a primate. In addition: Humans in a video are acceptable, if primate(s) and not the humans are the main focus of the video. If humans are the main focus of the video, label as irrelevant. If it is hard to recognize anything on an image ( blurry, bad lighting), label as irrelevant.  If a video contains text overlaying large parts of the video, label as irrelevant (Logos or text in the corners is fine).}

\paragraph{CLIP embeddings and relevance filter} 
We use \verb|laion/CLIP-ViT-H-14-laion2B-s32B-b79K| from HuggingFace to extract CLIP embeddings \cite{schuhmann_laion-5b_2022}. 
We utilize a 2-layer MLP with input dimension 1024, the CLIP embeddings' dimension, and hidden dimension 256. A dropout layer with probability 0.7 is also applied. We train for 18 epochs using Adam with a learning rate of $10^{-4}$. We use a relevance threshold of 3.5.

\paragraph{Cut Detection} 
We utilize PySceneDetect~\cite{castellano_pyscenedetect_nodate} for detecting cuts in videos. PySceneDetect calculates pixel value differences between frames in HSV colorspace and splits videos at locations with high change. We are using the adaptive detector with a threshold of 3.0.

\paragraph{Data Contamination Screening}
We compare all samples of  the target datasets and YT-Filtered with a CLIP similarity $> 0.9$ and found none of ChimpACT, ChimpBehave or BaboonLand within YT-Filtered. Only 4\,s of video material from the training set of PanAf500 (without any labels) overlap with YT-Filtered. For another train video, we found few YT-Filtered snippets recorded from the same camera trap at different times.

\paragraph{Primate Detection} We are using \verb|IDEA-Research/grounding-dino-base| from HuggingFace as zero-shot object detector \cite{liu_grounding_2024}.   We use the prompt ``monkey.primate.ape.'' and a box confidence threshold of 0.35.

\section{Evaluation Protocol}
\label{sec:supp_evaluation_datasets}

\begin{table}[]
    \nicetable
    \caption{\textbf{Properties of our target datasets for evaluation.} \textbf{CV}: 5-fold cross-validation instead of a  test set. \textbf{ML:} multi-label. \textbf{SL:} single-label. \textbf{Num. Samples:} ChimpACT: Number of annotated (not-interpolated) frames, i.e. every 10th frame. Others: Number of miniclips. }
    \begin{tabular}{lrrrrll}
    \toprule
     &  \multicolumn{3}{c}{\textbf{Num. Samples}} & \multicolumn{2}{c}{\multirow{2}{*}{\textbf{Classes}}}  & \multirow{2}{*}{\thead{ \textbf{Evaluation} \\ \textbf{Protocol}}} \\
     \cmidrule(lr){2-4}
     & train & val & test & && \\
     \midrule
 ChimpACT & 50k & 6.8k & 7.6k &23& ML & frames w/ det. \\
 PanAf500 & 9k & 0.8k & 1.9k&9& SL & miniclip (16f) \\
 BaboonLand & 17.8k&-&5.6k&13&SL & miniclip (90f) \\
 ChimpBehave & 9.2k &-& CV &7&SL& miniclip (20f) \\
 \bottomrule
 \end{tabular}
    \label{tab:supp_available_datasets}
\end{table}

See \cref{tab:supp_available_datasets} for details about our four target datasets for evaluation. 

\para{PanAf500}~\cite{brookes_panaf20k_2024} consists of 125 minutes of camera trap videos from 18 field sites in tropical Africa capturing chimpanzees and gorillas. Following the established protocol~\cite{brookes_panaf20k_2024, sakib_visual_2020}, we train and evaluate on 16-frame miniclips cropped to primates using ground truth bounding boxes. Each miniclip shows one of nine behaviors.

\para{ChimpBehave}~\cite{fuchs_forest_2025} is a dataset of chimpanzees in an indoor enclosure at the Basel zoo. Videos were captured using handheld cameras and behavior recognition is evaluated on 20-frame miniclips cropped to primate bounding boxes and only featuring a single behavior. The dataset features seven behavior classes, which are a subset of the PanAf500 classes. Instead of a dedicated test set, ChimpBehave utilizes five-fold cross-validation.

\para{BaboonLand}~\cite{duporge_baboonland_2025} consists of 18 drone recordings of wild-living olive baboons residing in Mpala, Kenya. From 30\,min of densely annotated 5.3\,k resolution drone footage, \citet{duporge_baboonland_2025} extracted 20\,h of spatio-temporal miniclips, centered on each animal. Each miniclip is annotated into one of twelve behavior classes plus an additional class for occlusions. Annotation is single-label per miniclip with majority vote over per-frame labels.

\para{ChimpACT}~\cite{ma_chimpact_2023} contains 2\,h of video footage of chimpanzees recorded at the Leipzig Zoo. It features frame-wise bounding boxes and multi-label behavior annotations across 23 classes. Two different behavior recognition tasks exist: one with \cite{ma_chimpact_2023} and one without \cite{ma_alphachimp_2024} access to ground truth bounding boxes. We predict the labels for a primate $i$ at frame $j$ by sampling a 64-frame miniclip around frame $j$ and producing a crop centered at $i$'s bounding box with padding to incorporate scene context.

\paragraph{Evaluation Metrics} Prior works used differing naming conventions for evaluation metrics. BaboonLand~\cite{duporge_baboonland_2025} refers to accuracy as Top-1 Micro and to balanced accuracy as Top-1 Macro. ChimpBehave~\cite{fuchs_forest_2025} refers to accuracy as Top1 and to balanced accuracy as MCA.

\paragraph{Training Data Size Ablation} For our evaluation of how well our model and X3D perform with less labeled data (\cref{sec:labeled-data-efficiency}), we create subsets of the dataset with 50\,\%, 25\,\%, and 10\,\% of the training dataset. We randomly sample three different versions of the dataset per desired size. During sampling we ensure that there is at least one sample of each class in each subset.

\section{Implementation Details and Prior Work}
\label{sec:supp_evaluation}

\paragraph{Our method}
We always train five attentive classifiers, evaluate each on the evaluation dataset, and report the mean evaluation score. 

\paragraph{Origin of prior work results (\cref{sec:comparison-sota}, \cref{tab:main_comparison})}
On ChimpACT, results for InternVideo-L and VideoPrism-g are from \cite{zhao_videoprism_2024} and  results for AlphaChimp and X3D are our own reproduction. On PanAf500, results for ChimpVLM are from \cite{brookes_chimpvlm_2024}. All other results are from the respective dataset papers (ChimpACT~\cite{ma_chimpact_2023}, PanAf500~\cite{brookes_panaf20k_2024}, BaboonLand~\cite{duporge_baboonland_2025}, ChimpBehave~\cite{fuchs_forest_2025})

\paragraph{AlphaChimp} There are two different evaluation protocols for ChimpACT: AlphaChimp \cite{ma_alphachimp_2024} and PySlowFast-based methods \cite{ma_chimpact_2023} calculate mAP over 18 classes, excluding five tail classes with very low support (begging, being begged from, taking object, losing object, erection), and perform early stopping on the test set, while results reported in VideoPRISM~\cite{zhao_videoprism_2024} report mAP over 23 classes and do not utilize the test set for model selection. We decided to follow the VideoPRISM protocol and reproduce AlphaChimp results using this protocol for fair comparison. Using the official AlphaChimp~\cite{ma_alphachimp_2024} implementation and leaving hyperparameters unchanged, we are able to match the reported results in \cite{ma_alphachimp_2024} (36.5 mAP in our reproduction; reported results are 34.3 mAP). We then switch to early stopping on validation instead of test and calculating mAP over all 23 classes, leaving everything else constant. As we noticed high variance in the evaluation scores obtained after each epoch, we report the mean over 5 runs. This yields a mAP of 25.35 for AlphaChimp.

\begin{figure*}[t]
    \centering
\includegraphics[width=0.8\linewidth]{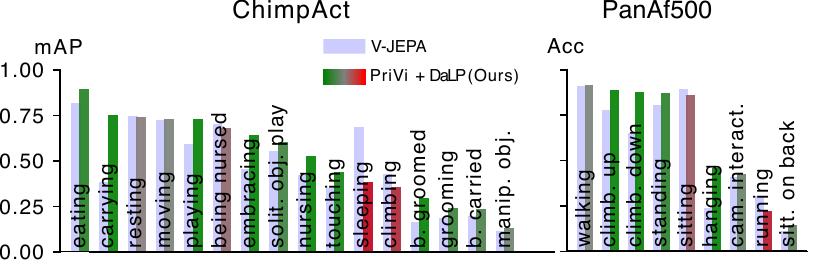}
    \caption{\textbf{Performance of our method (\prmt{}\,+\,\dalp{}) by action class.} Color gradient shows improvement \csquare{mplgreen}, comparable results \csquare{mplgray}, or deterioration \csquare{mplred} compared to V-JEPA \csquare{vjepa_blue}.}
    \label{fig:performance_breakdown}
\end{figure*}

\paragraph{X3D} We are using X3D-L for all experiments, keeping the original input size of $16 \times 312 \times 312$. For scaling X3D with less labels (\cref{fig:head_data_ablations}) on PanAf500, we utilize the PySlowFast codebase, providing the dataset in Kinetics format. For reporting X3D on ChimpACT (\cref{tab:main_comparison}) and for scaling X3D with less labels on ChimpACT (\cref{fig:head_data_ablations}), we predict on miniclips using our ChimpACT dataloader instead of following the AVA protocol, matching the evaluation setting we are using.

\section{Additional Results for Our Method}
\label{sec:supp_performance_breakdown}

\paragraph{Performance Breakdown}
Figure~\ref{fig:performance_breakdown} shows a breakdown of model performance by action class on ChimpACT and PanAf500. For chimpanzees in a zoo setting (ChimpACT), we find good performance for classes with visually distinct appearances or characteristic motion patterns, like eating (89\,\% mAP), carrying (75\,\%), resting (73\,\%), or playing (72\,\%).

For chimpanzees in the wild (PanAf500, camera traps), the majority classes (sitting, walking, standing, 85-91\,\% Acc), as well as rare classes with distinct motion (climbing up and down, 87-89\,\% Acc) perform well. Sitting on back (14\,\% Acc) and hanging (46\,\%) underperform, we speculate this is due to PanAf500's evaluation protocol of cropping to bounding boxes, which loses global information. 

For ChimpBehave, we see a high performance in general, with 90-98\,\% Acc for sitting, standing, walking each. For baboons in the wild (BaboonLand, drone footage), we observe a similar pattern as with ChimpACT, with Walking/Running (95\,\% Acc), Sitting/Standing (87\,\% Acc) and Drinking (83\,\%). Behaviors requiring fine-grained information, like Self-Grooming, Being Groomed, Grooming Somebody, Mutual Grooming perform worse (14-38\,\% Acc).

\begin{table}[]
\centering
    \nicetable
    \renewcommand{\crossdomain}[1]{\textbf{{\color{gray} #1}}}
    \caption{\textbf{Joint domain- and dataset-level pretraining reliably mitigates catastrophic forgetting.} Results are on \emph{val} sets.}
    \begin{tabular}{lrrrr}
    \toprule
    & \multicolumn{2}{c}{\textbf{ChimpACT}}& \multicolumn{2}{c}{\textbf{PanAf500}}\\
    &mAP  &mAP$_w$& Acc & B-Acc \\
    \midrule 
    {\prmt{}}  &{38.75} & {54.32} & {89.65} & {79.95}\\
         + \dalp{}: ChimpACT &  {41.43}  & {57.22} & \crossdomain{84.93} & \crossdomain{69.05} \\ 
         + \dalp{}: PanAf500 & \crossdomain{32.90}	& \crossdomain{48.91}	 & {90.53} & {87.29}  \\
         \midrule
        \textbf{Joint \prmt{} \& ChimpACT}&  40.68& 56.39& \crossdomain{89.91}& \crossdomain{80.89}\\ 
         \textbf{Joint \prmt{} \& PanAf500}& \crossdomain{37.93}& \crossdomain{55.80}& {91.97}& {85.74}\\

 \bottomrule
 \end{tabular}

    \label{tab:supp_joint_pretraining}
\end{table}

\paragraph{Joint pretraining instead of \dalp{}}
To mitigate catastrophic forgetting, we experiment with joint pretraining on \prmt{} and the target dataset instead of first training on \prmt{} and then performing dataset-level pretraining (\dalp{}) on the target dataset. We start with the V-JEPA checkpoint and jointly pretrain for 75\,k steps with a 50:50 mix of \prmt{} and the target dataset. We find that joint pretraining improves performance on the selected target dataset without reducing performance on other target datasets, see \cref{tab:supp_joint_pretraining}. While this requires considerably more training compute than \dalp{}, it might be good option for scenarios where catastrophic forgetting is a concern.

\end{document}